\documentclass{article}

\usepackage{microtype}
\usepackage{graphicx}
\usepackage{subfigure}
\usepackage{booktabs} % for professional tables

\usepackage{hyperref}
\usepackage{url}
\usepackage{algorithm}
\usepackage{algorithmic}

\usepackage{mathtools}
\usepackage{paralist}
\usepackage{subfigure}
\usepackage{url}
\usepackage{booktabs}       % professional-quality tables
\usepackage{amsfonts}
\usepackage{nicefrac}
\usepackage{xcolor}

\newcommand{\PPINN}{P$^2$INNs}

\usepackage{multicol, multirow}

\usepackage{caption}
\usepackage{makecell}

\usepackage[accepted]{icml2024}

\usepackage{amsmath}
\usepackage{amssymb}
\usepackage{amsthm}

\usepackage[capitalize,noabbrev]{cleveref}

\theoremstyle{plain}

\theoremstyle{definition}

\theoremstyle{remark}

\usepackage[textsize=tiny]{todonotes}

\icmltitlerunning{Parameterized Physics-informed Neural Networks for Parameterized PDEs}

\begin{document}

\twocolumn[
% \icmltitle{P$^2$INNs: Parameterized Physics-informed Neural Networks \\ for Solving Parameterized PDEs}
\icmltitle{Parameterized Physics-informed Neural Networks for Parameterized PDEs}




\begin{icmlauthorlist}
\icmlauthor{Woojin Cho}{yonsei,asu}
\icmlauthor{Minju Jo}{lg}
\icmlauthor{Haksoo Lim}{yonsei}
\icmlauthor{Kookjin Lee}{asu}
\icmlauthor{Dongeun Lee}{texas}
\icmlauthor{Sanghyun Hong}{oregon}
\icmlauthor{Noseong Park}{kaist}
\end{icmlauthorlist}

\icmlaffiliation{yonsei}{Yonsei University}
\icmlaffiliation{lg}{LG CNS}
\icmlaffiliation{asu}{Arizona State University}
% \icmlaffiliation{telepix}{Telepix}
\icmlaffiliation{texas}{Texas A\&M University-Commerce}
\icmlaffiliation{oregon}{Oregon State University}
\icmlaffiliation{kaist}{KAIST}

\icmlcorrespondingauthor{Noseong Park}{noseong@kaist.ac.kr}

\icmlkeywords{Machine Learning, ICML}

\vskip 0.3in
]



\printAffiliationsAndNotice{}  % leave blank if no need to mention equal contribution

\begin{abstract}
Complex physical systems are often described by partial differential equations (PDEs) that depend on parameters such as the Reynolds number in fluid mechanics. In applications such as design optimization or uncertainty quantification, solutions of those PDEs need to be evaluated at numerous points in the parameter space. While physics-informed neural networks (PINNs) have emerged as a new strong competitor as a surrogate, their usage in this scenario remains underexplored due to the inherent need for repetitive and time-consuming training. In this paper, we address this problem by proposing a novel extension, parameterized physics-informed neural networks (P$^2$INNs). P$^2$INNs enable modeling the solutions of parameterized PDEs via explicitly encoding a latent representation of PDE parameters. With the extensive empirical evaluation, we demonstrate that P$^2$INNs outperform the baselines both in accuracy and parameter efficiency on benchmark 1D and 2D parameterized PDEs and are also effective in overcoming the known ``failure modes''.
\end{abstract}

\section{Introduction}
Scientific machine learning (SML)~\cite{baker2019workshop} has been growing fast. Unlike traditional tasks in machine learning, e.g., image classification and object detection, SML requires exact satisfaction of important physical characteristics.
Recent work has developed various deep-learning models that encode such physical characteristics, 
that are physically-consistent (e.g., by enforcing conservation laws~\cite{raissi2019physics,lee2021deep} or preserving structures~\cite{Greydanus2019hnn,toth2019hamiltonian,lutter2018deep,cranmer2020lagrangian,lee2021machine}) and symmetrical (e.g., modeling invariance or equivariance by 
design~\cite{battaglia2018relational,satorras2021n}). %, to name a few. 
Among those methods, physics-informed neural networks (PINNs)~\cite{raissi2019physics} %have gained great attention 
are gaining traction in the research community %due to 
because of their sound computational formalism to enforce governing physical laws to learn solutions. PINNs are also %very 
easy to implement by using automatic differentiation \cite{baydin2018automatic} and gradient-based training algorithms that are readily available in any deep-learning %APIs 
frameworks, such as \textsc{PyTorch} \cite{paszke2019pytorch} %and 
or \textsc{TensorFlow} \cite{abadi2016tensorflow}. 
\begin{table}
\centering
\small
\caption{%
\textbf{P$^2$INNs greatly improve the quality of CDR solutions.}
We compare the average absolute (Abs.) and relative (Rel.) errors of PINN and P$^2$INN in six different CDR equations. 
IMP is the improvement ratio of P$^2$INN to PINN. 
(see Section~\ref{sec:Experimental Environments} for details).}

\setlength{\tabcolsep}{2pt}
\renewcommand{\arraystretch}{0.8}
\begin{tabular}{lrrrrrr}
\specialrule{1pt}{2pt}{2pt}
                         \multirow{3}{*}{\textbf{PDE type}}& \multicolumn{2}{c}{\textbf{PINN}} & \multicolumn{2}{c}{\textbf{P$^2$INN}} & \multicolumn{2}{c}{\textbf{IMP. (\%)}} \\ \cmidrule{2-7}
                          & Abs. & Rel. & Abs. & Rel. & Abs. & Rel. \\
\specialrule{1pt}{2pt}{2pt}
 \multirow{1}{*}{\textbf{Convection}} & 0.0496  &0.0871 & 0.0330 & 0.0241  & 33.43 & 72.37 \\ \midrule
                         \multirow{1}{*}{\textbf{Diffusion}} & 0.3611 & 0.6939 & 0.1592 & 0.3190 & 55.91 & 54.03  \\ \midrule
                         \multirow{1}{*}{\textbf{Reaction}}  & 0.5825 & 0.6431 & 0.0041 & 0.0069 & 99.30 &  98.92  \\ \midrule
 \multirow{1}{*}{\textbf{Conv.-Diff.}} & 0.1493 & 0.2793 & 0.0532 & 0.1236 & 64.34 & 55.75 \\ \midrule
                          \multirow{1}{*}{\textbf{Reac.-Diff.}} & 0.4744 & 0.5614 & 0.1319 & 0.2008 & 72.21 & 64.24                                                          \\
\midrule
  \multirow{1}{*}{\textbf{Conv.-Diff.-Reac.}} & 0.4811 & 0.5315 & 0.0391 & 0.0759 & 91.88 & 85.72 \\
\specialrule{1pt}{2pt}{2pt}
\end{tabular}   

\label{tbl:summary_table}
\end{table}

\begin{figure}[t]
\includegraphics[width=0.96\columnwidth]{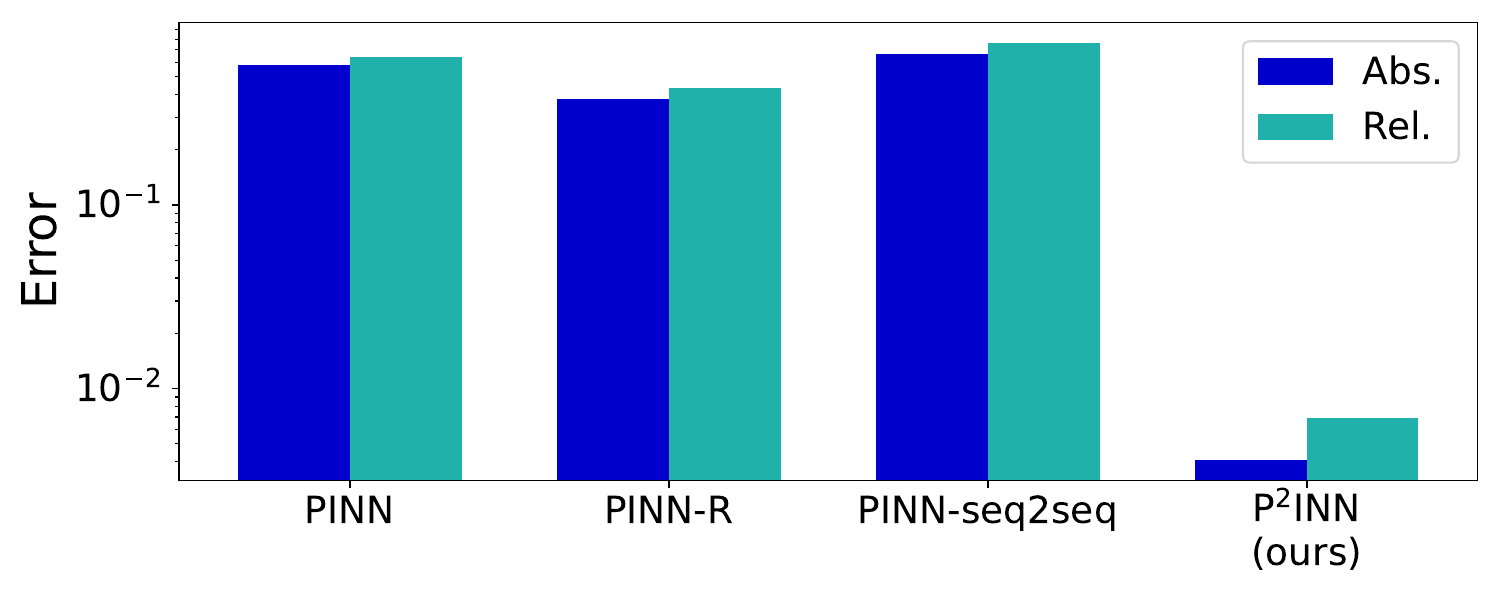}
\caption{%
\textbf{P$^2$INNs outperform the baselines.} P$^2$INNs reduce the average $L_2$ absolute (Abs.) and relative (Rel.) errors by 100$\times$ compared to the baselines. The results are for reaction equations -- the most challenging problems for PINNs.}
\label{fig:comparing_reaction_error}
\end{figure}

PINNs parameterize the solution $u(x,t)$ of partial differential equations (PDEs) 
using a neural network $u_{\Theta}(x,t)$ 
that takes the spatial and temporal coordinates $(x,t)$ as input 
and has $\Theta$ %is 
as the model parameters. 
During training,
the neural network minimizes a \emph{PDE residual loss}  
(cf. Eq.~\eqref{eq:loss_L_f})
denoting the governing equation, at a set of collocation points, and 
a \emph{data matching loss} (cf. Eqs.~\eqref{eq:loss_L_u} and~\eqref{eq:loss_L_b}), which enforces an initial condition and a boundary condition, at another set of collocation points sampled from initial/boundary conditions. This computational formalism %of PINNs 
enables to infuse the physical laws, described by the governing equation $\mathcal{F}(x, t, u)$, into the solution 
model and, thus, is denoted as ``physics informed.''
PINNs have shown to be effective in solving many different PDEs, such as Navier--Stokes equations~\cite{shukla2021parallel,jagtap2020extended,jagtap2020conservative}.
While powerful, PINNs suffer from several obvious weaknesses.
\begin{compactenum} 
    \item[\textbf{W1})] PDE operators are highly nonlinear (making training extremely difficult); 
    \item[\textbf{W2})] Repetitive trainings from scratch are needed when solutions to new PDEs are sought (even for new PDEs arising from new PDE parameters in parameterized PDEs).
\end{compactenum}
There have been various efforts to mitigate each of these issues: 
(For addressing W1) curriculum-learning-type training algorithms that train PINNs from easy PDEs to hard PDEs\footnote{Following the notational conventions in curriculum learning, we use the terms, ``easy'' and ``hard,'' to indicate data that are easy or hard for neural networks to learn.} \cite{krishnapriyan2021characterizing}, 
and (for addressing W2) meta-learning PINNs \cite{liu2022novel}; or directly learning solutions of parameterized PDEs such that $u_{\Theta}(x,t;\pmb{\mu})$, where $\pmb{\mu}$ is a set of PDE parameters, e.g., $\pmb{\mu} = [\beta, \nu, \rho]$ in convection-diffusion-reaction (CDR) equations. However, there has been a less focus on addressing both problems in a unified PINN framework. 

To mitigate the both issues in W1 and W2 simultaneously, we propose a variant of PINNs for solving parameterized PDEs, called \emph{parameterized physics-informed neural networks} (\PPINN{}). \PPINN{} approximate solutions as a neural network of a form $u_{\Theta}(x,t;\pmb{\mu})$ (for resolving W2) and are capable of inferring approximate solutions with accuracy (cf. Table~\ref{tbl:summary_table} and Figure~\ref{fig:comparing_reaction_error}) even for harder PDEs (for resolving W1). A novel modification %we make to the vanilla PINN architecture is that the 
proposed in our model is to explicitly extract a hidden representation of the PDE parameters by employing a separate encoder network, $\pmb{h}_{\text{param}}~=g_{\Theta_p}(\pmb{\mu})$, and uses this hidden representation to parameterize the solution neural network,  $u_{\Theta}(x,t;\pmb{h}_{\text{param}})$. Rather than simply treating $\pmb{\mu}$ as a coordinate in the parameter domain, we infer useful information of PDEs from the PDE parameters $\pmb{\mu}$, constructing the latent manifold on which the hidden representation of each PDE lies. 

To demonstrate the effectiveness of the proposed model, we demonstrate the performance of the proposed model with well-known benchmarks~\cite{krishnapriyan2021characterizing}, i.e., parameterized CDR equations. As studied in~\cite{krishnapriyan2021characterizing}, certain choices of the PDE parameters (e.g., high convective or reaction term) make training PINNs very challenging (i.e., harder PDEs), and our goal is to show that the proposed method is capable of producing approximate solutions with reasonable accuracy for those harder PDEs.

To sum up, our contributions are as follows:
\begin{itemize}
    \item We design a novel neural network architecture for solving parameterized PDEs, \PPINN{}, which significantly improves the performance of PINNs overcoming the well-known weaknesses (W1 and W2).
    \item We empirically demonstrate that explicitly encoding the PDE parameters into a hidden representation plays an important role in improving performance. 
    \item We show that the proposed \PPINN{} are able to learn all the experimented benchmark PDEs via a single training run and greatly outperforms existing PINN-based methods in terms of prediction accuracy.
\end{itemize}

\section{Background and Motivation}
\label{sec:motivation}
{
We start by providing illustrative examples of parameterized PDEs and their solutions to motivate a development of a new efficient variant of PINNs for \textit{multi-query} and \textit{real-time} scenarios. Details on the PDEs can be found in Appendix.}

\subsection{Convection-Diffusion-Reaction Equations}\label{sec:cdr_eq}

As an example, we consider parameterized CDR equations:
\begin{equation}
    {\partial u\over\partial t} \!+\! \beta{\partial u \over\partial x} \!-\! \nu{\partial^2 u\over\partial x^2} \!-\! {\rho u (1 \!-\! u)} \!=\! {0}, \ x \in \Omega, \; t \in [0,T].
    \label{eq:eq_cdr}
\end{equation} 
The equation describes how the state variable $u$ changes over time with the existence of convective (the second term), diffusive (the third term), and reactive (the fourth term) phenomena.
Here, ${\beta}$ is a coefficient about how fast transportable the equation is, ${\nu}$ is a diffusivity for the diffusion phase, and $\rho$ is a scaling parameter about spreading velocity. 
Note that we %specifically 
choose the well-known Fisher's form $\rho{u(1-u)}$, which was used in~\cite{krishnapriyan2021characterizing}, as our reaction term.

{ We note that we choose this class of PDEs due to many advantages: (1) solution characteristics are varying significantly based on PDE parameters, (2) some PDE parameter values introduce challenging situations for PINNs (a.k.a ``failure modes''), and (3) analytical solutions exist. However, we also note that our proposed method is not specifically limited to this PDE class, but is applicable to general PDE classes (e.g., see Section~\ref{sec:helmholtz} for 2D cases).}

\iffalse
\subsection{Helmholtz Equations}
{
We employ the specific Helmholtz equations which were used in \citep{mcclenny2020self} as benchmark PDEs,and the equations are represented in Appendix~\ref{sec:data_detail}. The Helmholtz equations describe the behavior of state variable $u$ in a 2D space, accounting for the effects of wave propagation, and a source term represented by $q(x,y)$. Here $k$ is a parameter related to wave frequency, while $a_1$ and $a_2$ control the spatial variations of the source term. In our experiments, we set the parameters $a_1$ and $a_2$ to a common value $a$.}
\fi

\begin{figure}[t]
    \centering
    \subfigure[Conv. ($\beta=5$)]{\includegraphics[width=0.33\columnwidth]{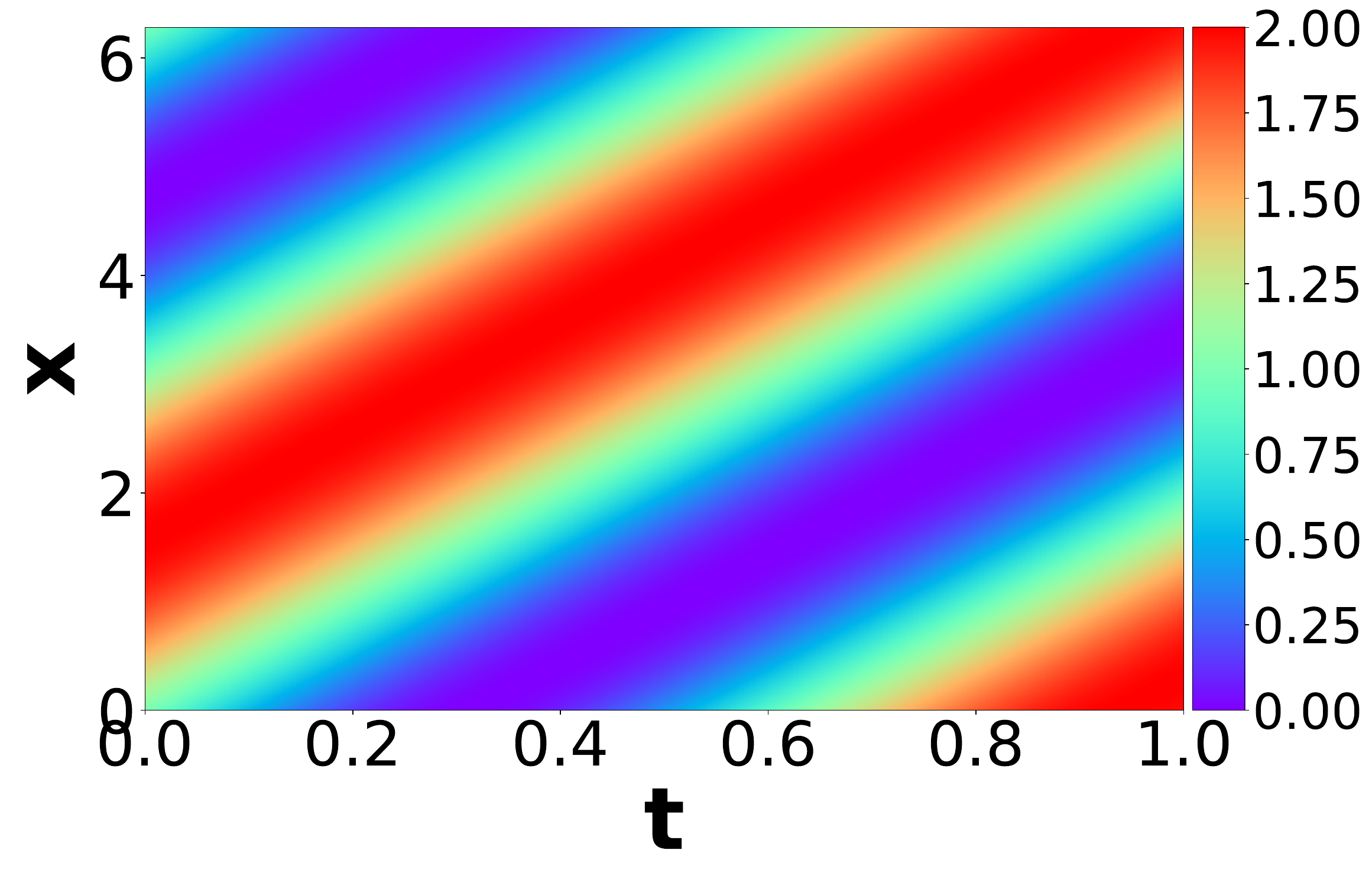}\label{fig:easy_conv}}\hfill
    \subfigure[Conv. ($\beta=10$)]{\includegraphics[width=0.33\columnwidth]{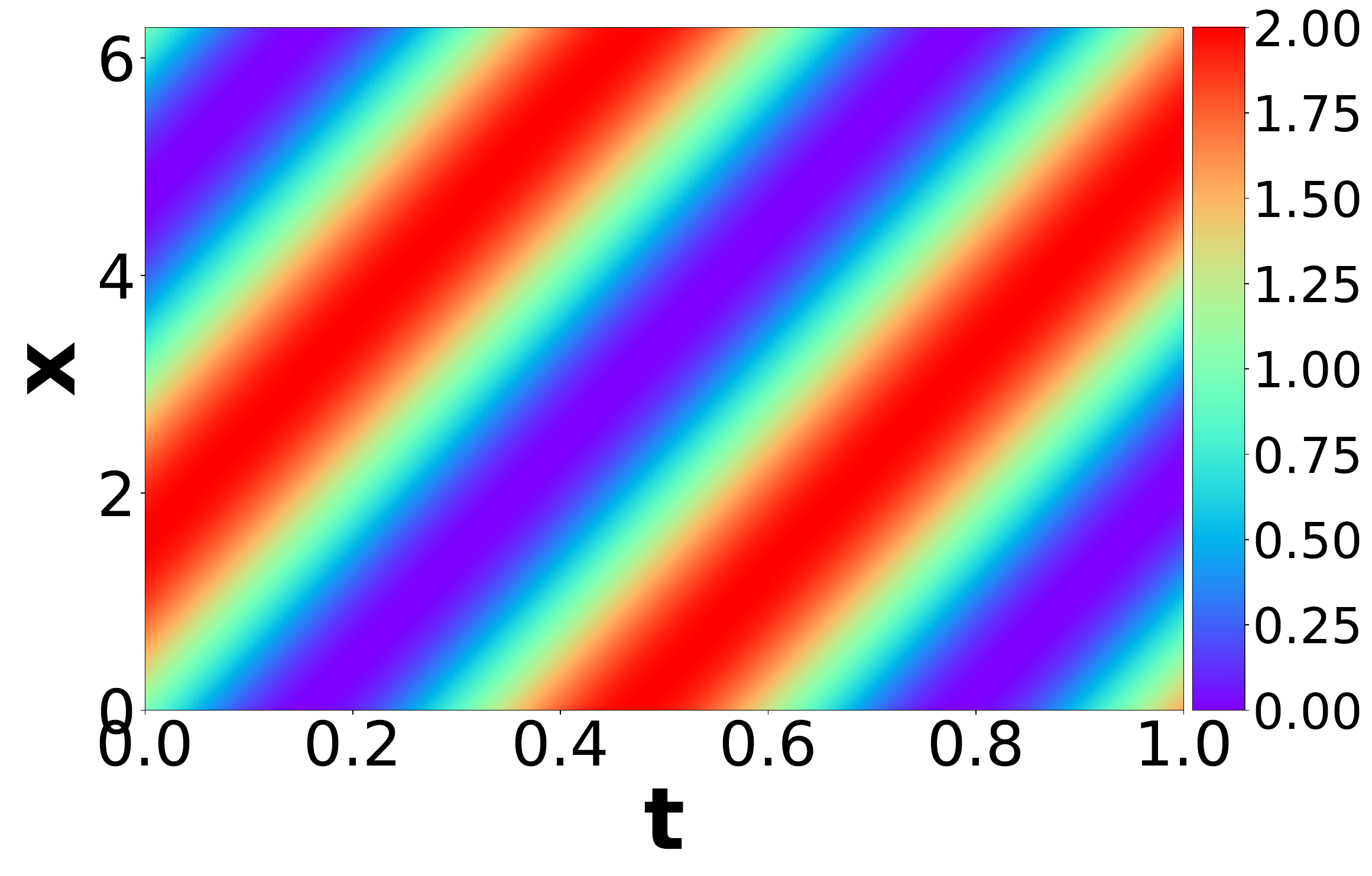}}\hfill
    \subfigure[Conv. ($\beta=15$)]{\includegraphics[width=0.33\columnwidth]{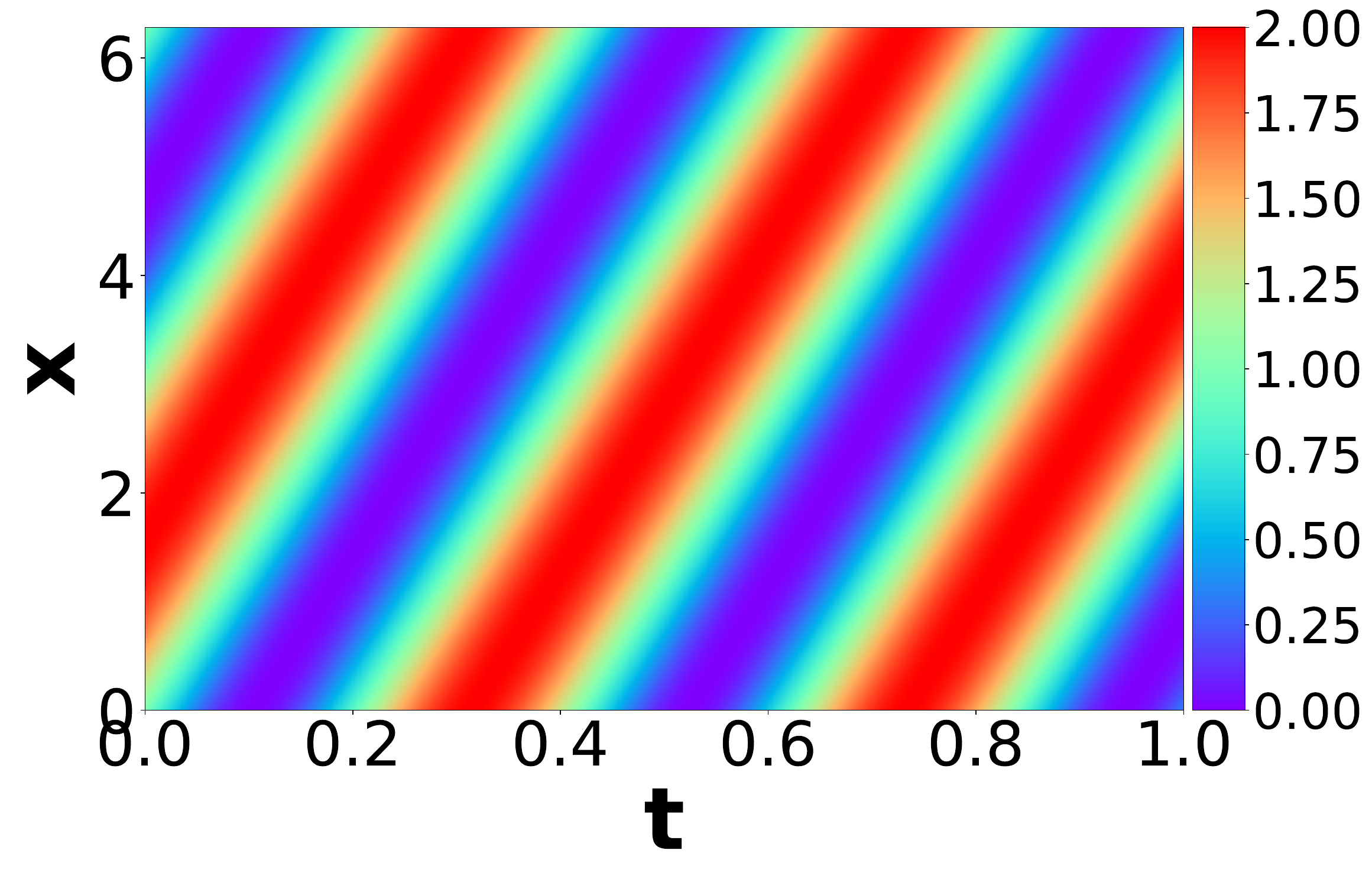}\label{fig:hard_conv}}\\
    
    \subfigure[Reac. ($\rho=1$)]{\includegraphics[width=0.33\columnwidth]{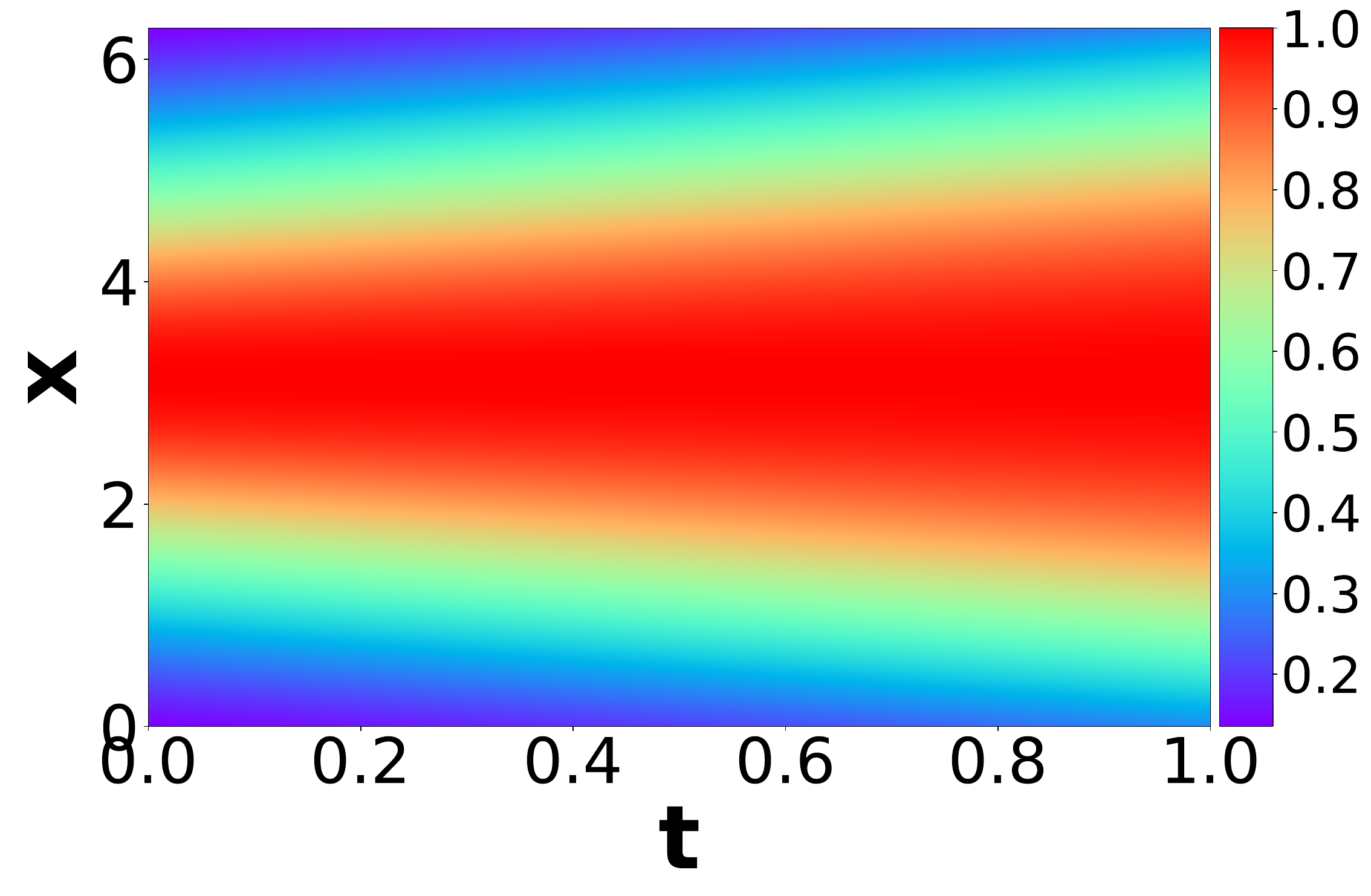}}\hfill
    \subfigure[Reac. ($\rho=4$)]{\includegraphics[width=0.33\columnwidth]{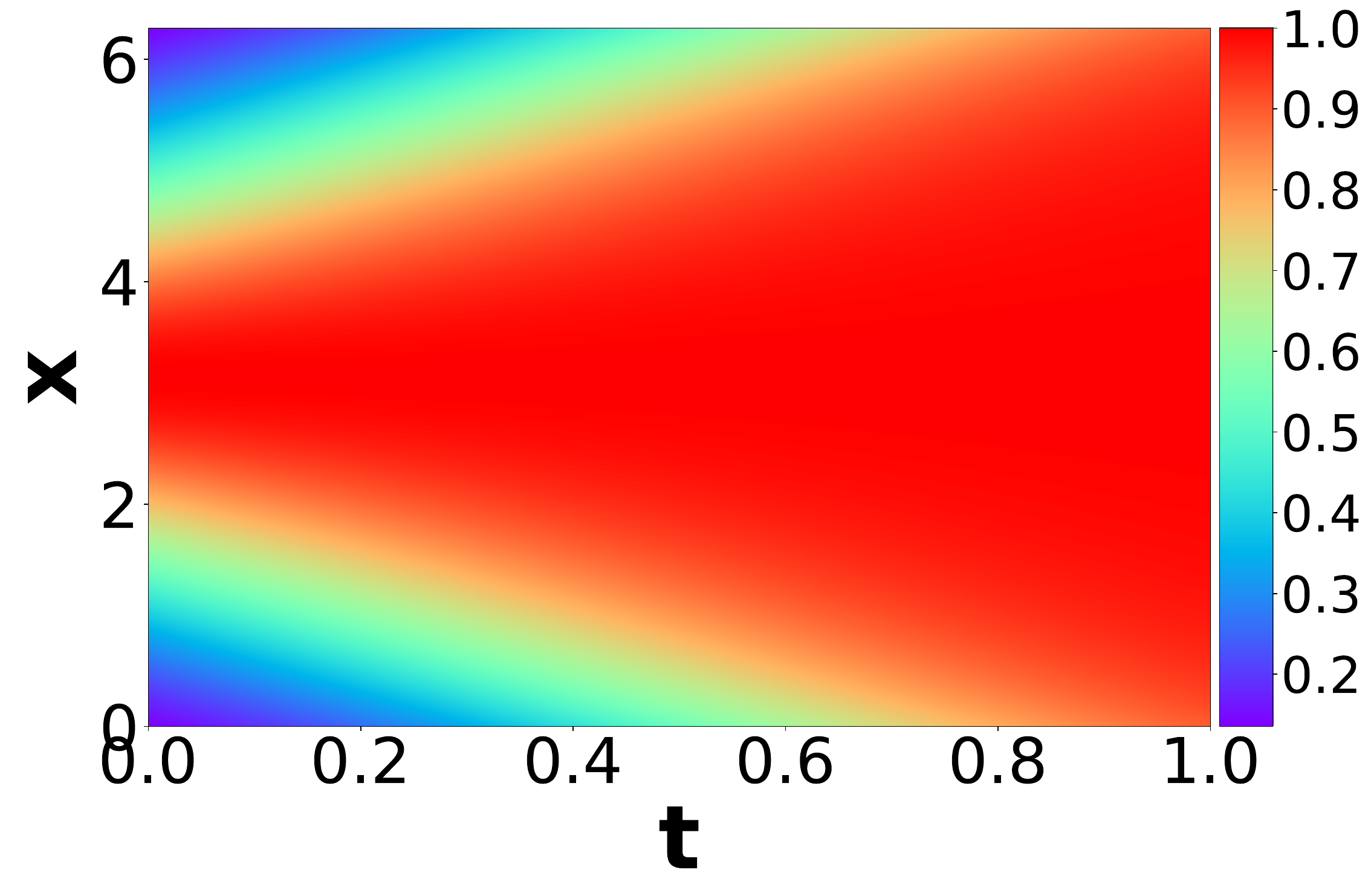}}\hfill
    \subfigure[Reac. ($\rho=7$)]{\includegraphics[width=0.33\columnwidth]{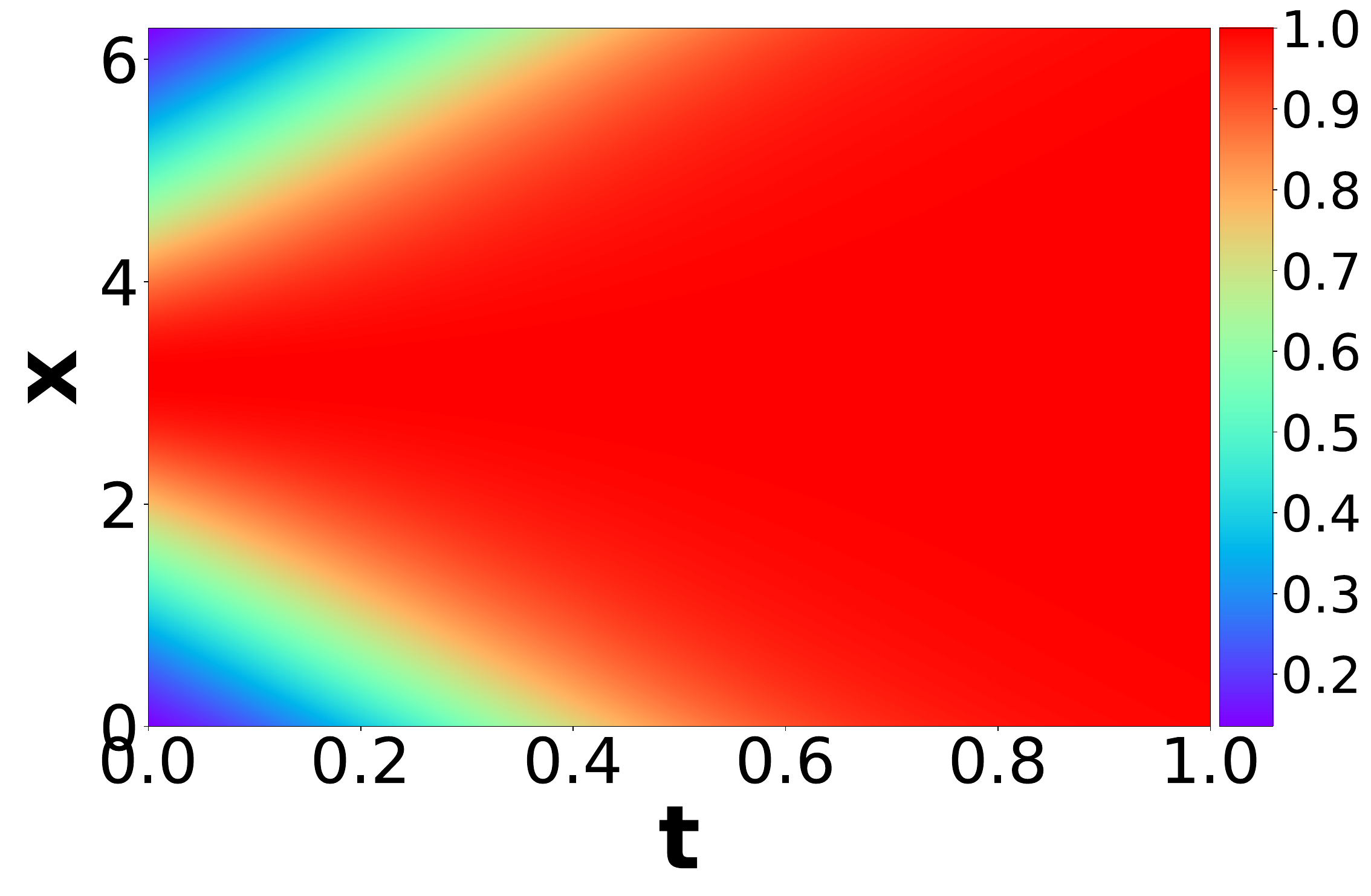}}\\
    
    \caption{The ground-truth solutions of various convection equations with an initial condition of $1 + \sin(x)$ (Figure~\ref{fig:figure1}. (a)-(c)) and reaction equations with an initial condition of
    a Gaussian distribution $N(\pi, ({\pi}/2)^2)$ (Figure~\ref{fig:figure1}. (d)-(f)).
    % $exp(-\frac{(x-\pi)^2}{2(\pi/4)^2})$        
    We note that varied solutions are made (with similar architectures) depending on changes in coefficient.
    }\label{fig:figure1}
% \end{minipage}
\end{figure}

\begin{figure}[t]
    \centering
    \subfigure[Conv.]{\includegraphics[width=0.32\columnwidth]{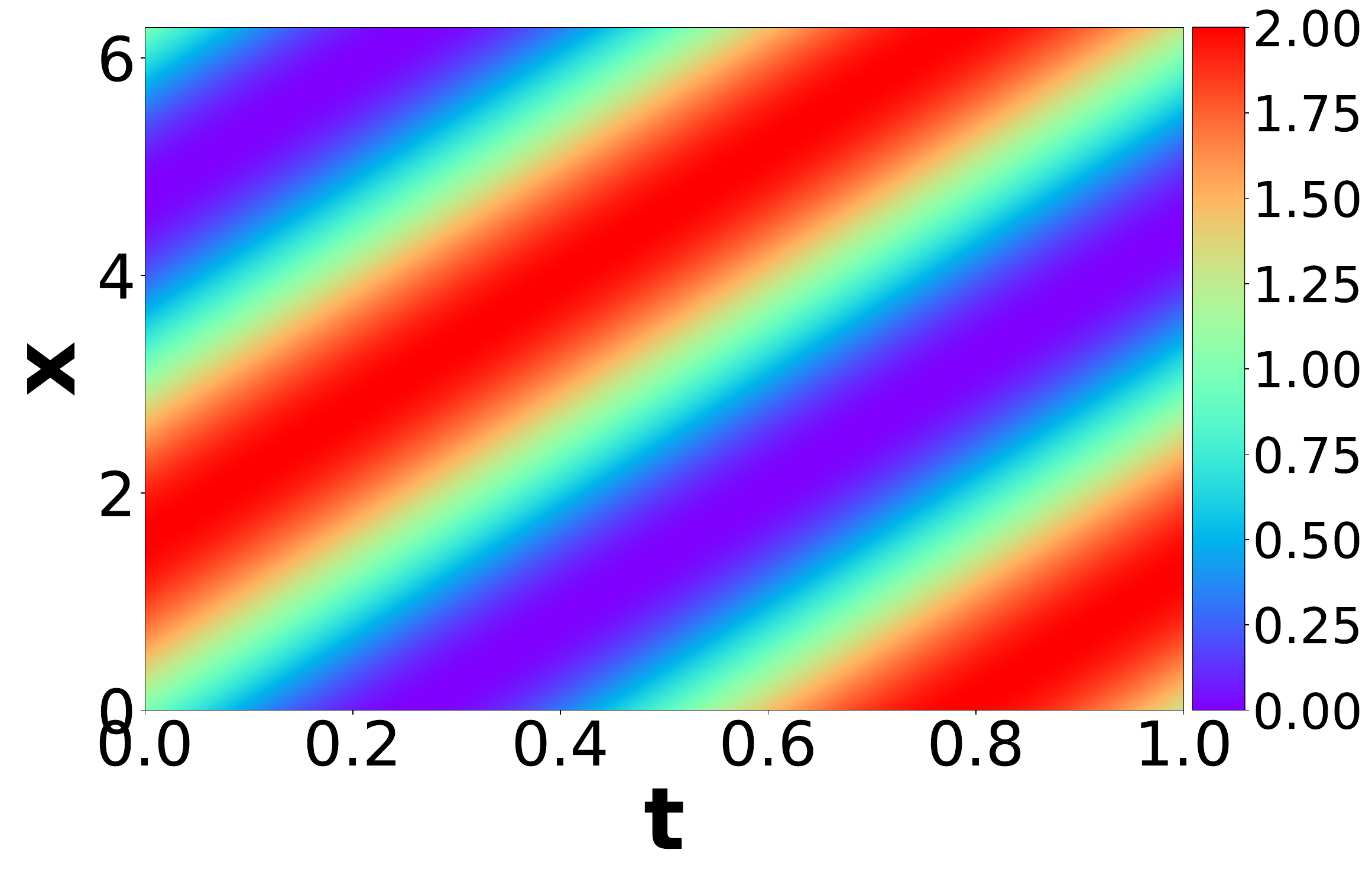}}\hfill
    \subfigure[Diff.]{\includegraphics[width=0.32\columnwidth]{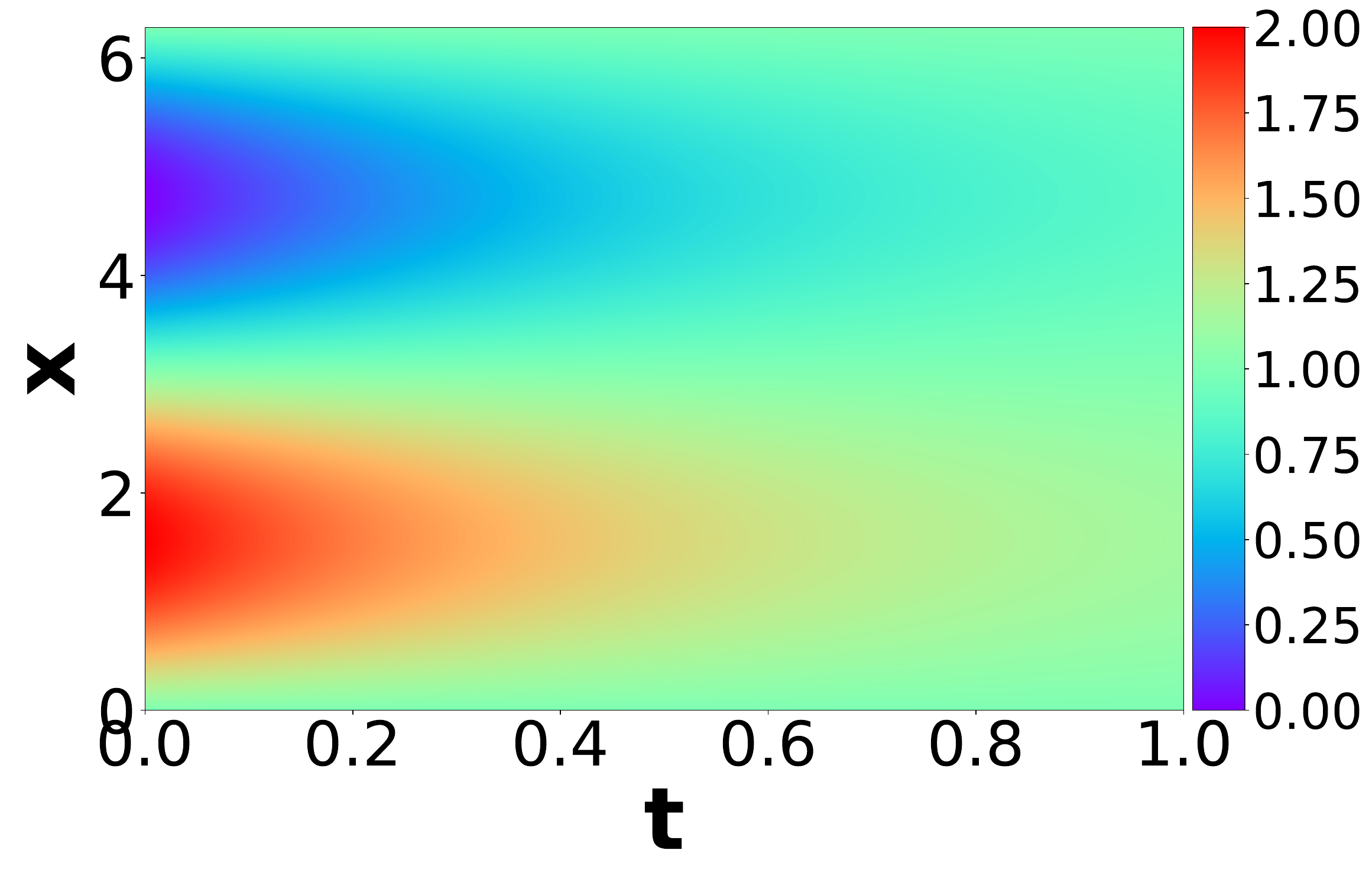}}\hfill
    \subfigure[Conv.-Diff.]{\includegraphics[width=0.32\columnwidth]{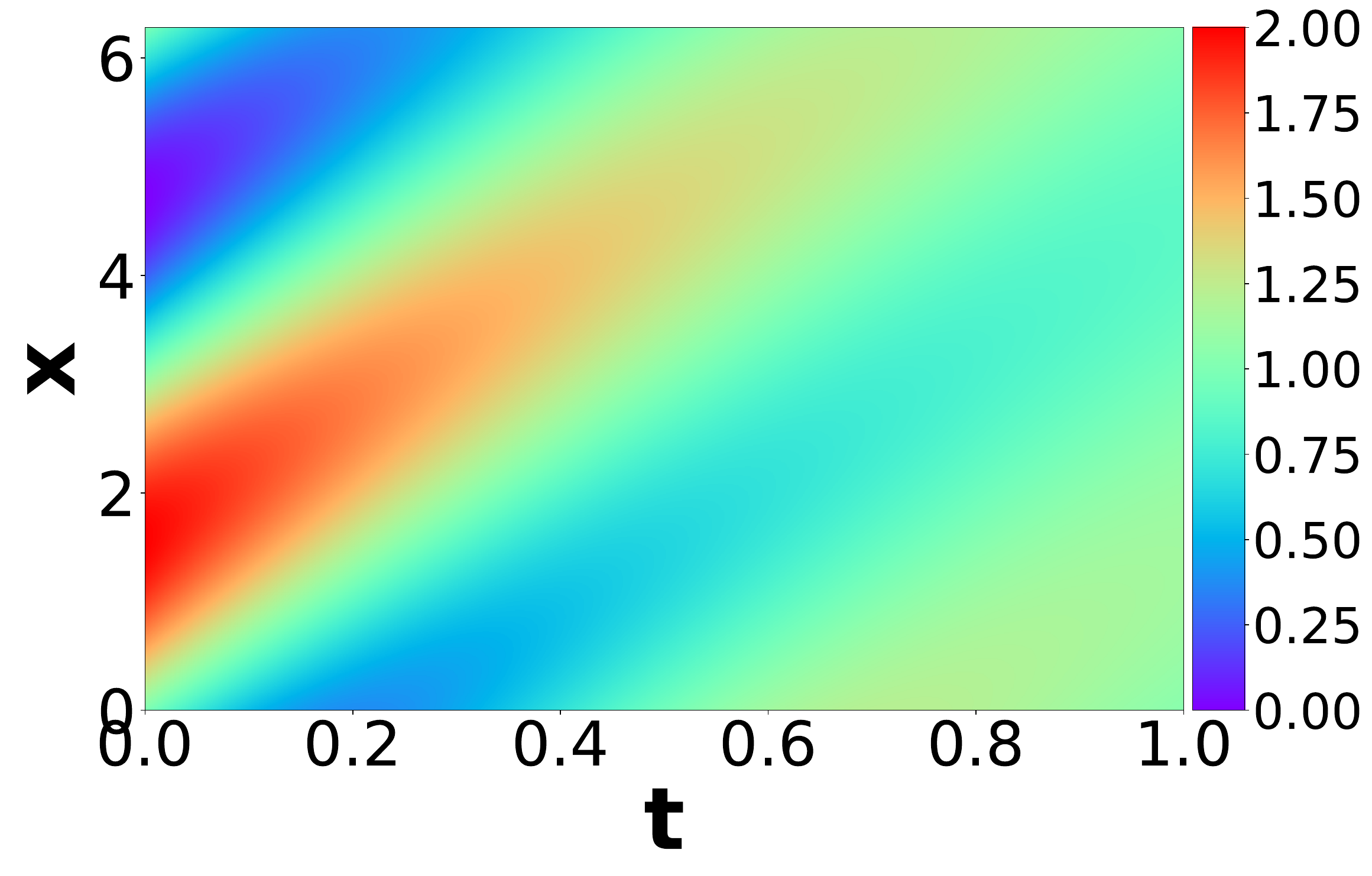}}\\
    
    \subfigure[Reac.]{\includegraphics[width=0.32\columnwidth]{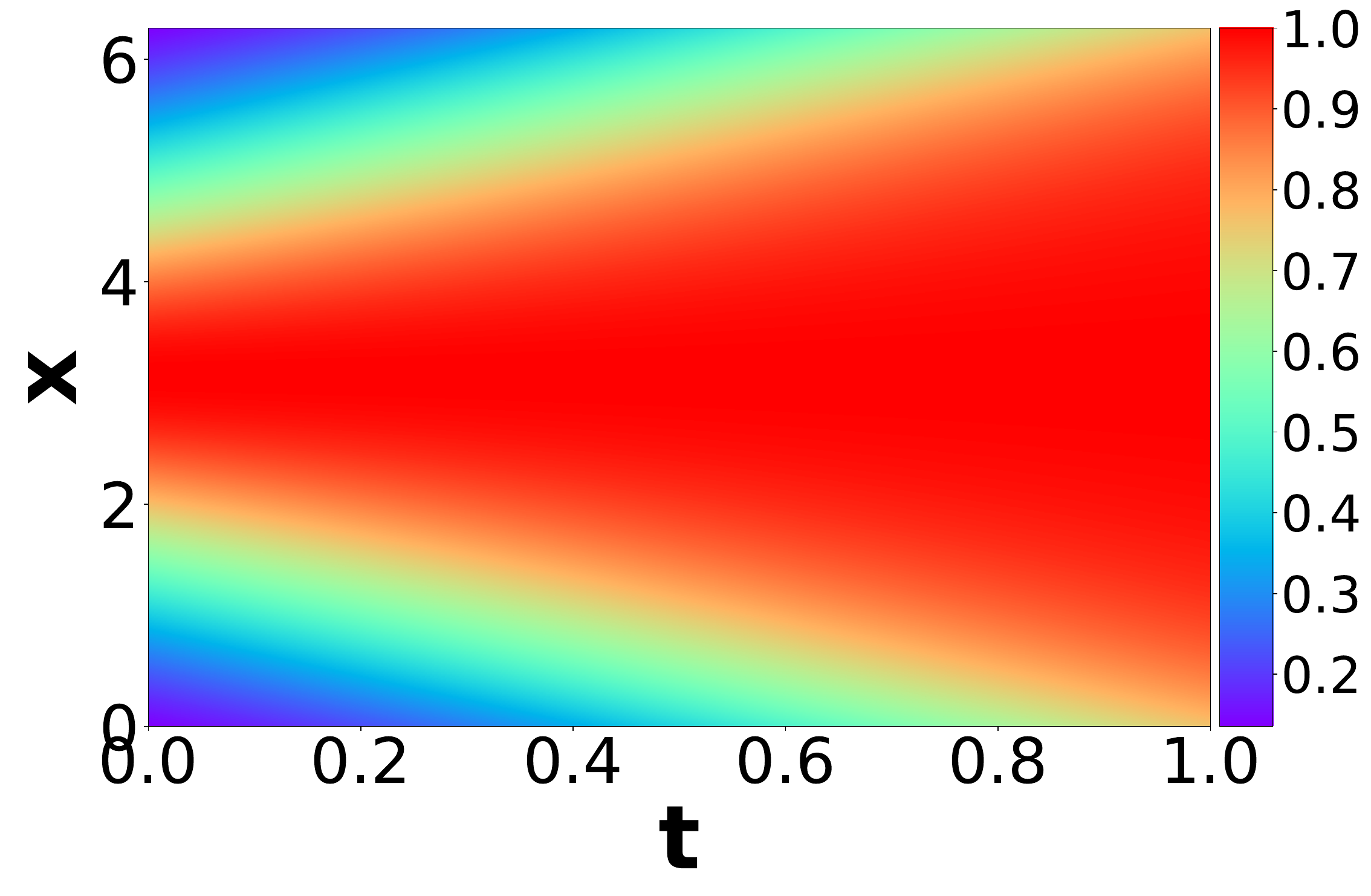}}\hfill
    \subfigure[Diff.]{\includegraphics[width=0.32\columnwidth]{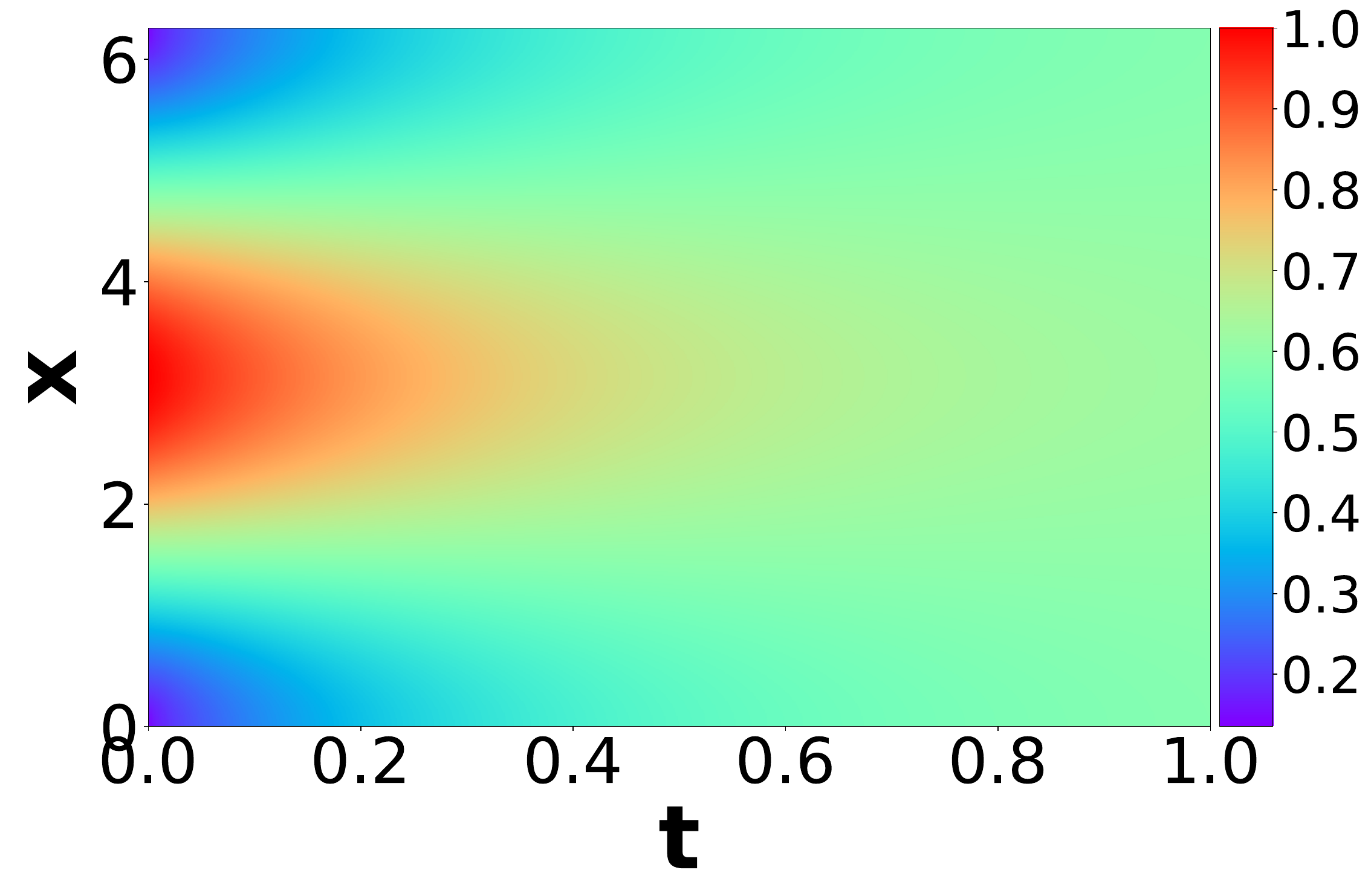}}\hfill
    \subfigure[Reac.-Diff.]{\includegraphics[width=0.32\columnwidth]{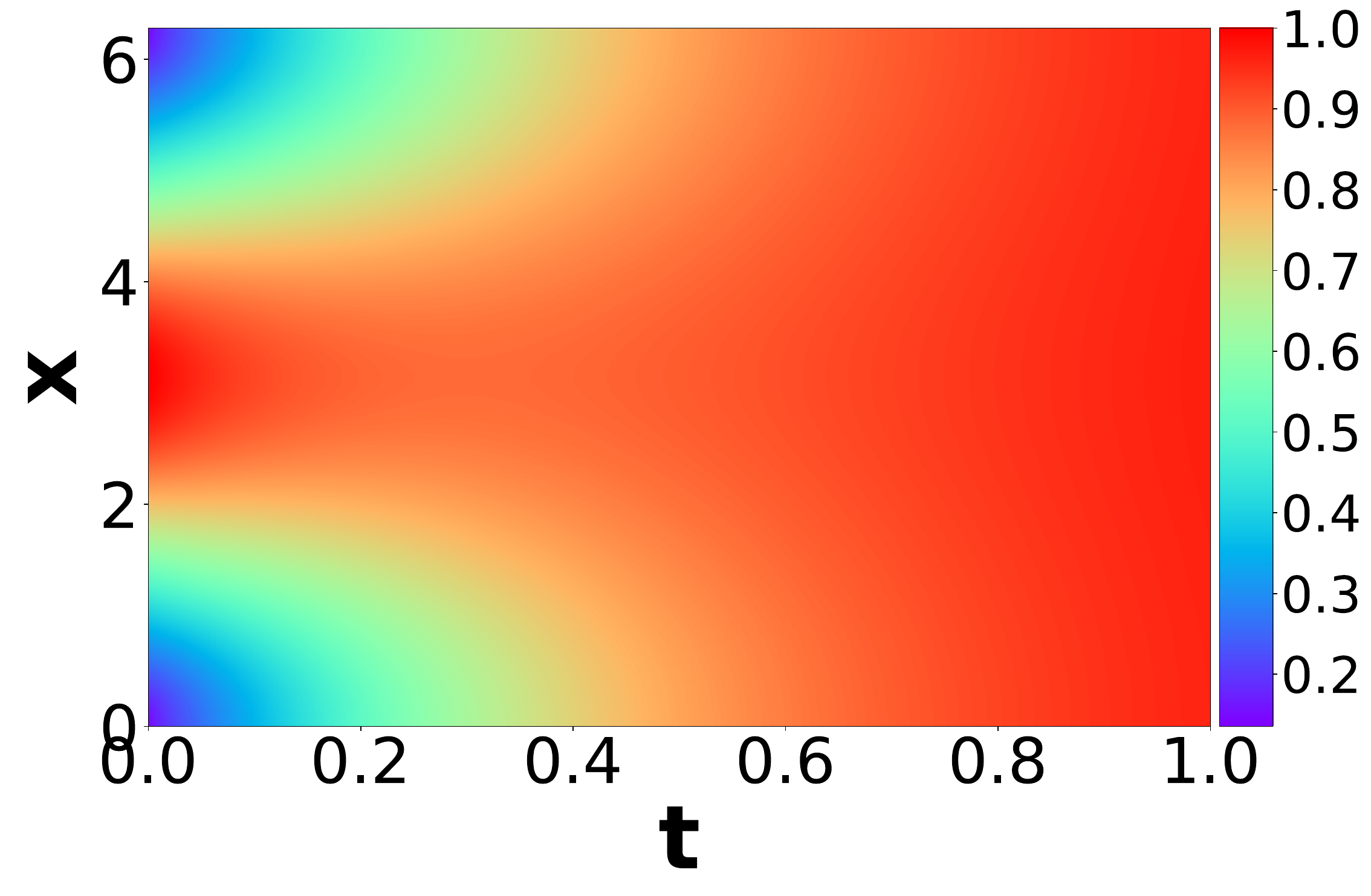}}\\
    
    \caption{The ground-truth solutions of various CDR equations with an initial condition of $1 + \sin(x)$ (Figure~\ref{fig:figure2}. (a)-(c)) or a Gaussian distribution $N(\pi, ({\pi}/2)^2)$ (Figure~\ref{fig:figure2}. (d)-(f)).
    % $exp(-\frac{(x-\pi)^2}{2(\pi/2)^2})$.
    We note that the solution in the last column reflects the first two columns' solutions. Therefore, there also exist similarities across different equation types.}\label{fig:figure2}
\end{figure}

\begin{figure*} [t]
\centering
\includegraphics[width=1.7\columnwidth]{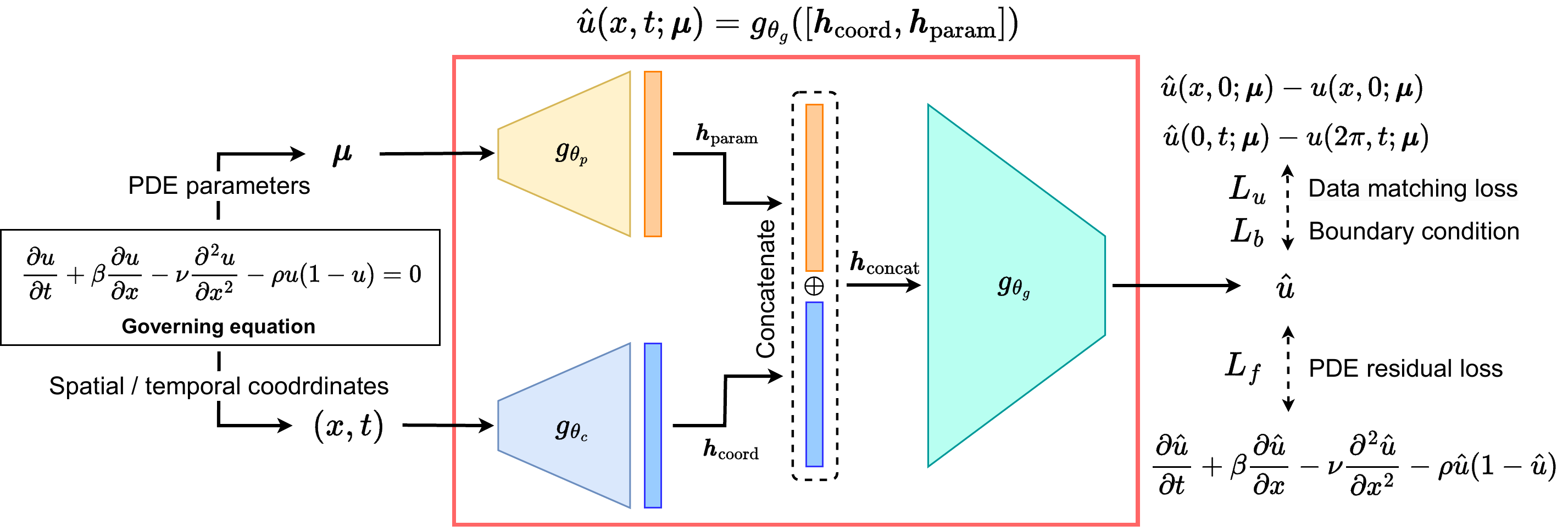}
\caption{\textbf{\PPINN{} architecture.} The two encoders $g_{\theta_p}$ and $g_{\theta_c}$ are added to generate better representations for the PDE parameter and the spatial/temporal coordinate. We also customize the manifold network $g_{\theta_g}$. In this figure, we provide the CDR equation as an example.}\label{fig:archi}
\end{figure*}

\subsection{Motivation}\label{sec:Motivation_sub}
Our goal is to develop a method to solve parameterized PDEs via the computational formalism of PINNs' overcoming W1 and W2. With this in mind, we first attempt to obtain intuitions from the visual inspection of solution snapshots displayed on the $(x,t)$-coordinate space (Figures~\ref{fig:figure1} and \ref{fig:figure2}).

The first set of the examples is shown in Figure~\ref{fig:figure1}: The ground-truth solutions of convection equations (top row) and reaction equations (bottom row) with varying parameters $\beta$ and $\rho$, respectively. As we vary the PDE parameter, e.g., increasing $\beta$, we obtain gradually changing solutions (i.e., becoming more oscillatory, as we go left from Figure~\ref{fig:easy_conv} to Figure~\ref{fig:hard_conv}). This suggests that model parameters of PINNs for varying PDE parameters could have similar values and this can be leveraged in the training of PINNs.

This observation indeed has been investigated in \cite{krishnapriyan2021characterizing} to solve hard PDEs for PINNs. With a higher convective term (large $\beta$), the PDE becomes a hard problem for PINNs to solve due to the spectral bias \cite{rahaman2019spectral} (i.e., the solution is highly oscillatory in time). Thus, \cite{krishnapriyan2021characterizing} proposed a curriculum-learning algorithm which starts to feed an easier PDE and gradually increases $\beta$ until it reaches the target value. This approach, however, drops all the intermediate model parameters obtained in the course of training. Instead, in our approach, we utilize all PDE information to train a single model for the solutions of parameterized PDEs.

The second set of examples (Figure~\ref{fig:figure2}, the solutions of different types of PDEs) provide a similar observation as above: even for different classes of PDEs (e.g., convection, diffusion, and convection-diffusion equations), the solutions gradually change, which can be leveraged in training PINNs.

\paragraph{Motivation \#1: a latent space of parameterized PDEs may exist.}
Since PDEs with similar parameter settings share common characteristics, we conjecture that solutions of parameterized PDEs can be embedded onto a latent space and reconstructed by using a shared decoder network. 

\paragraph{Motivation \#2: it will be more effective to solve similar problems simultaneously.} 
Considering the similarities between solutions parameterized by similar PDE parameters, we conjecture that 
training can be improved by attempting to solve all those similar problems together --- multi-task learning approaches are also based on the same intuition~\cite{kendall2018multi,ruder2017overview}.

Motivated by the observations, we develop a new approach that alleviates the two known weaknesses W1 and W2.

\iffalse
\subsection{PINNs with modulation methods}
The shift modulation proposed in~\cite{functa22} involves adding a shift term to the bias of each layer in the model, to better represent various data with a small number of learnable parameter. However, directly applying shift modulation to PINN does not lead to significant performance improvement. Therefore, we propose a SVD-based modulation method specifically tailored for P$^2$INNs, enhancing its generalizability.
\fi
\section{\PPINN{}: Parameterized PINNs}
Now we introduce our parameterized physics-informed neural networks (\PPINN{}). In essence, our goal is to design a neural network architecture that effectively emulates the action of the parameterized PDE solution function, $u(x,t;\pmb{\mu})$.

\subsection{Model Architecture}

For \PPINN{}, we propose a modularized design of the neural network $u_{\Theta}(x,t; \pmb{\mu})$, which consists of three parts, i.e., two separate encoders $g_{\theta_p}$ and $g_{\theta_c}$, and a manifold network $g_{\theta_g}$ such that  
% \begin{linenomath}
\begin{align}
    u_{\Theta}(x,t; \pmb{\mu}) = g_{\theta_g} ([g_{\theta_c}(x,t); g_{\theta_p}(\pmb{\mu}) ] ),
\end{align}
% \end{linenomath}
where $\Theta = \{ \theta_c, \theta_p, \theta_g\}$ denotes the set of model parameters. The two encoders, $g_{\theta_c}$ and $g_{\theta_p}$, take the spatiotemporal coordinate $(x,t)$ and the PDE parameters $\pmb{\mu}$ as inputs and extract hidden representations such that $\pmb{h}_{\text{coord}} = g_{\theta_c}(x,t)$ and $\pmb{h}_{\text{param}} = g_{\theta_p}(\pmb{\mu})$. The two extracted hidden representations are then concatenated and fed into the manifold network to infer the solution of of the PDE with the parameters $\pmb{\mu}$ at the coordinate $(x,t)$, i.e., $\hat u(x,t;\pmb{\mu}) = g_{\theta_g}([\pmb{h}_{\text{coord}};\pmb{h}_{\text{param}}])$.  
Figure~\ref{fig:archi} summarizes the \PPINN{} architecture. 

The important design choice here is that we explicitly encode the PDE parameters into a hidden representation as opposed to treating the PDE parameters merely as a coordinate in the parameter domain, e.g., $(x,t,\pmb{\mu})$ is combined and directly fed into our ablation model, called PINN-P, for our ablation study in Section~\ref{sec:ablation}. With the abuse of notation, \PPINN{} can be expressed as a function of $(x,t)$, parameterized by the hidden representation: $u_{\Theta}(x,t;\pmb{\mu}) = u_{\{ \theta_c,\theta_g\}}(x,t;\pmb{h}_{\text{param}})$. This expression emphasizes our intention that we explicitly utilize the PDE model parameters to characterize the behavior of the solution neural network.   

\subsubsection{Encoder for Equation Input}\label{sec:enc1}
The equation encoder $g_{\theta_p}$ reads the PDE parameters, and generates a hidden representation of the equation, denoted as $\pmb{h}_{\text{param}}$. We employ the following fully-connected (FC) structure for the encoder:
\begin{align}
    \pmb{h}_{\text{param}} = \sigma(FC_{D_p} \cdots (\sigma(FC_2(\sigma(FC_1(\pmb{\mu})))))),
\end{align} where $\sigma$ denotes a non-linear activation, such as ReLU and tanh, and $FC_i$ denotes the $i$-th FC layer of the encoder. $D_p$ means the number of FC layers. 

We note that $\pmb{h}_{\text{param}}$ has a size larger than that of $\pmb{\mu}$ in our design to encode the space and time-dependent characteristics of the parameterized PDE. Since highly non-linear PDEs show different characteristics at different spatial and temporal coordinates, we intentionally employ relatively high-dimensional encoding.

\subsubsection{Encoder for Spatiotemporal Coordinate}
The spatial and temporal coordinate encoder $g_{\theta_c}$ generates a hidden representation $\pmb{h}_{\text{coord}}$ for $(x, t)$. 
This encoder has the following FC layer structure:
\begin{align}
    \pmb{h}_{\text{coord}} = \sigma(FC_{D_c} \cdots (\sigma(FC_2(\sigma(FC_1(x,t)))))),
    \label{tab:eq_3}
\end{align}
where $FC_i$ and $D_c$ denote the $i$-th FC layer of this encoder and the number of FC layers, respectively. 

\subsubsection{Manifold Network}
The manifold network $g_{\theta_g}$ reads the two hidden representations, $\pmb{h}_{\text{param}}$ and $\pmb{h}_{\text{coord}}$, and infer the input equation's solution at $(x, t)$, denoted as  $\hat u(x,t;\pmb{\mu})$. With the inferred solution $\hat{u}$, we construct two losses, $L_u$ and $L_f$. 
The manifold network can have various forms but we use the following form:
\begin{align}
\hat u(x,t;\pmb{\mu}) = \sigma(FC_{D_g} \cdots \sigma(FC_1(\pmb{h}_{\text{concat}}))),
\label{tab:eq_4}
\end{align}
where $\pmb{h}_{\text{concat}} = \pmb{h}_{\text{coord}} \oplus \pmb{h}_{\text{param}}$, and $\oplus$ is the concatenation of the two vectors; $D_g$ denotes the number of FC layers.

\subsection{Training}\label{sec:train}
Model training is performed by minimizing the regular PINN loss. With the prediction $\hat u$ produced by \PPINN{}, our basic loss function consists of three terms as follows:
\begin{align}
 L({\Theta}) = {w_1}L_u + {w_2}L_f + {w_3}L_b, \label{eq:loss},
\end{align}
where $L_u$, $L_b$, and $L_f$ enforces initial, boundary conditions, and physical laws in PDEs, respectively, and  $w_1, w_2, w_3 \in \mathbb{R}$ are hyperparameters.
In general, the overall training method follows the training procedure of the original PINN \cite{raissi2019physics}. The only exception is that the PDE residual loss associated with multiple PDEs is minimized in a mini-batch whereas in the original PINN, the residual of only one PDE is minimized. To be more specific, in each iteration, we create a mini-batch of 
$\{\pmb{\mu}_i, (x_i,t_i)\}_{i=1}^{B}$, where $B$ is a mini-batch size. We randomly sample the collocation points and, thus, there can be multiple different PDEs, identified by $\pmb{\mu}_i$, in a single mini-batch.

\begin{figure}[t]
\centering
\includegraphics[width=0.95\columnwidth]{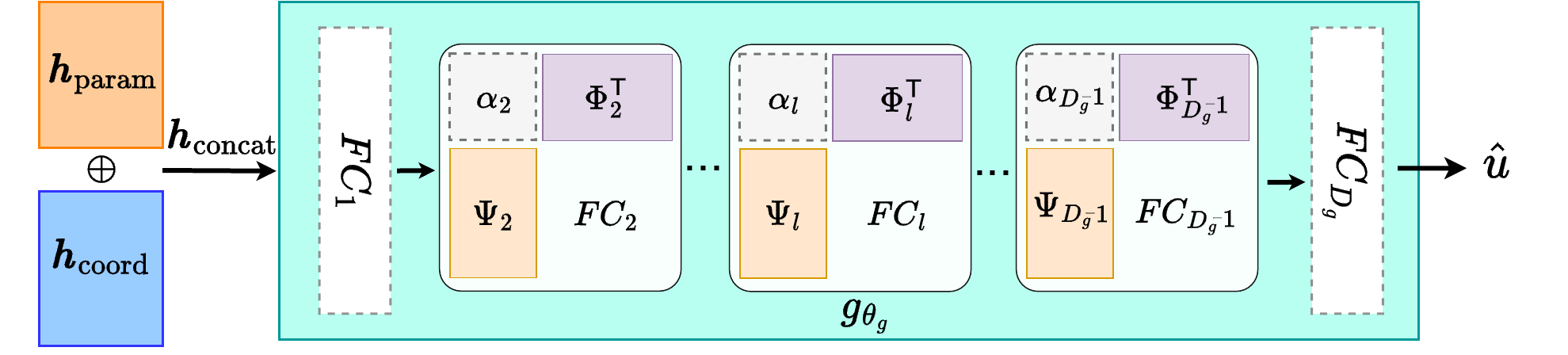}
\caption{\textbf{P$^2$INNs with SVD modulation.} From the pre-trained decoder layer of P$^2$INN, we obtain the bases $\Phi_l, \Psi_l$ for parameterized PDEs through SVD (cf. Eq.~\eqref{eq:eq_svd_mod}). Note that only the diagonal matrices $\alpha_l$ are used for fine-tuning. (The dotted lines represent learnable parameters.)}\label{fig:svd_mod}
\end{figure}

\subsection{Fast Fine-tuning}\label{sec:svd_modulation}
As the ultimate goal in this study is to deploy trained models to a set of specific PDE parameters of our interest, we devise a method to fine-tune the trained models to improve the solution accuracy at those query PDE parameters. To this end, we adopt an idea of SVD-PINNs~\cite{Gao_2022}, which shows that extracting basis through singular value decomposition (SVD) from the weights of a PINN trained on a single PDE equation is effective in transferring information. Extending this insight, we introduce an \textit{SVD modulation} by obtaining bases through applying SVD to the weights of the decoder layers of \PPINN{}. 
Specifically, only the manifold network $g_{\theta_g}$ is transformed to a form that can be modulated (cf. Figure~\ref{fig:svd_mod}); 
each layer, excluding the first and last layers, is decomposed as follows: 
\begin{align}
FC_l=\Psi_l \alpha_l \Phi_l^{\mathsf{T}}, \;\;\; l=2,3, \ldots, {D_g-1}.
\label{eq:eq_svd_mod}
\end{align}
Then, during fine-turning, we set $\{\alpha_l\}_{l=2}^{{D_g-1}}$ to be learnable, while keeping all other parameters in the network fixed. It is an option to fix the parameters of $FC_1$ and $FC_{D_g}$.

In the field of implicit neural representations, where the study on learning coordinate-based continuous neural function is conducted, the \textit{shift modulation}~\cite{functa22} has been one of leading architectural choices. This involves adding a shift term to the bias of each layer in the model, to better represent various data with a small number of learnable parameter. However, we found from empirical experiments that modulating by shift in PINNs does not lead to significant performance improvement. We discuss this further in Section~\ref{sec:CDR_all}. 

\iffalse
We suggest the fine-tuning method, i.e., modulation, for \PPINN{}. SVD-PINNs~\cite{Gao_2022} show that extracting basis through singular value decomposition (SVD) from the weights of a neural network trained on a single PDE equation is effective in transferring information. Extending this insight, we introduce a SVD modulation by obtaining bases through SVD from the decoder of \PPINN{}, a neural network trained on multiple parameterized PDEs. That is, only the manifold network $g_{\theta_g}$ is modulated while learning the equation (cf. Figure~\ref{fig:svd_mod}). 
At first, each layer, excluding the first and last layers, is decomposed as follows: 
\begin{align}
FC_l=\Psi_l \alpha_l \Phi_l^{\mathsf{T}}, \;\;\; l=2,3, \ldots, {D_g-1}.
\label{eq:eq_svd_mod}
\end{align}
Then, during modulation, we set $\{\alpha_l\}_{l=2}^{{D_g-1}}$ to be learnable, while keeping all other parameters in the network fixed. It is an option to fix the parameters of $FC_1$ and $FC_{D_g}$.
\fi

\begin{table*}[t]
\centering
\renewcommand{\arraystretch}{0.3}
\footnotesize
\caption{The relative and absolute $L_2$ errors over all the equations. Our \PPINN{} surpass baselines in all but one cases, even without fine-tuning. IMP. denotes the rate of improvement of our model over the best baseline.}
\begin{tabular}{cccccccccc}
\specialrule{1pt}{2pt}{2pt}
\multirow{2}{*}{} & \multirow{2}{*}{\textbf{PDE type}} & \textbf{Coefficient} & \multirow{2}{*}{\textbf{Metric}} & \multirow{2}{*}{\textbf{PINN}} & \multirow{2}{*}{\textbf{PINN-R}}  & \multirow{2}{*}{\textbf{PINN-seq2seq}} & \multirow{2}{*}{\textbf{P$^2$INN}} & \multirow{2}{*}{\textbf{IMP. (\%)}}\\
 &  & \textbf{range}  &  &  &   & &  & \\
 
\specialrule{1pt}{2pt}{2pt}
\multirow{53}{*}{\textbf{Class 1}} & \multirow{15}{*}{\textbf{Convection}} & \multirow{2}{*}{1$\sim$5} & Abs. err. & 0.0183 & 0.0222  & 0.1281 & \textbf{0.0039} & 78.44 \\
 &  &  & Rel. err. & 0.0327 & 0.0381  & 0.2160 & \textbf{0.0079} & 75.82\\ \cmidrule(lr){3-9}
 &  & \multirow{2}{*}{1$\sim$10} & Abs. err. & 0.0164 & 0.0666  & 0.1924 & \textbf{0.0093} & 43.62\\
 &  &  & Rel. err. & 0.0307 & 0.1195  & 0.3276 & \textbf{0.0179} & 41.78\\ \cmidrule(lr){3-9}
 &  & \multirow{2}{*}{1$\sim$20} & Abs. err. & 0.1140 & 0.1624  & 0.2252 & \textbf{0.0198} & 82.64\\
 &  &  & Rel. err. & 0.1978 & 0.2779  & 0.3819 & \textbf{0.0464} & 76.55\\ \cmidrule(lr){2-9}

 & \multirow{15}{*}{\textbf{Diffusion}} & \multirow{2}{*}{1$\sim$5} & Abs. err. & 0.1335 & 0.1665  & 0.1987 & \textbf{0.1322} & 0.97 \\
 &  &  & Rel. err. & 0.2733 & 0.3462  & 0.4050 & \textbf{0.2710} & 0.84 \\ \cmidrule(lr){3-9}
 &  & \multirow{2}{*}{1$\sim$10} & Abs. err. & 0.2716 & 0.3175  & 0.3149 & \textbf{0.1539}  & 43.34 \\
 &  &  & Rel. err. & 0.5259 & 0.6206  & 0.6174 & \textbf{0.3116} & 40.75 \\ \cmidrule(lr){3-9}
 &  & \multirow{2}{*}{1$\sim$20} & Abs. err. & 0.6782 & 0.7054  & 0.3346 & \textbf{0.1916} & 42.74 \\
 &  &  & Rel. err. & 1.2825 & 1.3401  & 0.6442 & \textbf{0.3745} & 41.87 \\ \cmidrule(lr){2-9}

 & \multirow{15}{*}{\textbf{Reaction}} & \multirow{2}{*}{1$\sim$5} & Abs. err. & 0.3341 & 0.3336  & 0.4714 & \textbf{0.0015} & 99.54 \\ 
 &  &  & Rel. err. & 0.3907 & 0.3907  & 0.5907 & \textbf{0.0027} & 99.31 \\ \cmidrule(lr){3-9}
 &  & \multirow{2}{*}{1$\sim$10} & Abs. err. & 0.6232 & 0.3619  & 0.6924 & \textbf{0.0065} & 98.19 \\
 &  &  & Rel. err. & 0.6926 & 0.4190  & 0.7931 & \textbf{0.0089} & 97.88 \\ \cmidrule(lr){3-9}
 &  & \multirow{2}{*}{1$\sim$20} & Abs. err. & 0.7902 & 0.4320  & 0.8246 & \textbf{0.0042} & 99.02 \\
 &  &  & Rel. err. & 0.8460 & 0.4932  & 0.8960 & \textbf{0.0092} & 98.14 \\
 
\specialrule{1pt}{2pt}{2pt}
\multirow{34}{*}{\textbf{Class 2}} & \multirow{15}{*}{\textbf{Conv.-Diff.}} & \multirow{2}{*}{1$\sim$5} & Abs. err. & 0.0610 & 0.0654  & 0.0979 & \textbf{0.0399} & 34.61 \\
 &  &  & Rel. err. & 0.1175 & 0.1289  & 0.1950 & \textbf{0.0892} & 24.05 \\ \cmidrule(lr){3-9}
 &  & \multirow{2}{*}{1$\sim$10} & Abs. err. & 0.1133 & 0.1313  & 0.0917 & \textbf{0.0576} & 37.25 \\
 &  &  & Rel. err. & 0.2098 & 0.2510  & 0.1959 & \textbf{0.1320} & 32.62 \\ \cmidrule(lr){3-9}
 &  & \multirow{2}{*}{1$\sim$20} & Abs. err. & 0.2735 & 0.2118  & 0.0645 & \textbf{0.0622} & 3.51\\
 &  &  & Rel. err. & 0.5106 & 0.4154  & 0.1504 & \textbf{0.1485} & 1.28\\ \cmidrule(lr){2-9}

 & \multirow{15}{*}{\textbf{Reac.-Diff.}} & \multirow{2}{*}{1$\sim$5} & Abs. err. & 0.1900 & 0.1876  & 0.4201 & \textbf{0.1225} & 34.70 \\
 &  &  & Rel. err. & 0.2702 & 0.2777  & 0.5346 & \textbf{0.1856} & 31.31 \\ \cmidrule(lr){3-9}
 &  & \multirow{2}{*}{1$\sim$10} & Abs. err. & 0.5166 & 0.3809  & 0.6288 & \textbf{0.1833} & 51.88 \\
 &  &  & Rel. err. & 0.6141 & 0.4790  & 0.7274 & \textbf{0.2756} & 42.46 \\ \cmidrule(lr){3-9}
 &  & \multirow{2}{*}{1$\sim$20} & Abs. err. & 0.7167  & 0.7210 & 0.7663 & \textbf{0.0898} & 81.03 \\
 &  &  & Rel. err. & 0.7998  & 0.8105 & 0.8337 & \textbf{0.1411} & 74.68\\
 
\specialrule{1pt}{2pt}{2pt}
\multirow{15}{*}{\textbf{Class 3}} & \multirow{15}{*}{\textbf{Conv.-Diff.-Reac.}} & \multirow{2}{*}{1$\sim$5} & Abs. err. & 0.1663 & 0.0865  & 0.4943 & \textbf{0.0311} & 64.02 \\
 &  &  & Rel. err. & 0.2057 & 0.1415  & 0.6104 & \textbf{0.0525} & 62.88 \\ \cmidrule(lr){3-9}
 &  & \multirow{2}{*}{1$\sim$10} & Abs. err. & 0.5321 & 0.3170  & 0.7051 & \textbf{0.0508} & 83.98 \\
 &  &  & Rel. err. & 0.5928 & 0.3772  & 0.8027 & \textbf{0.0939} & 75.10 \\ \cmidrule(lr){3-9}
 &  & \multirow{2}{*}{1$\sim$20} & Abs. err. & 0.7450 & 0.4080  & 0.7136 & \textbf{0.0353} & 91.94 \\
 &  &  & Rel. err. & 0.7960 & 0.4645  & 0.8100 & \textbf{0.0812} & 82.88 \\
\specialrule{1pt}{2pt}{2pt}
\end{tabular}
\label{tbl:result}
\end{table*}

\section{Evaluation}\label{others}
In this section, we test the performance of \PPINN{} on the benchmark PDE problems: 1D CDR equations and 2D Helmholtz equations, both of which are known to suffer from the failure modes.
We first layout our experimental setup
and show that \PPINN{} outperform the baselines with an extensive evaluation.
We further analyze how \PPINN{} address the failures shown in Section~\ref{sec:motivation}. Due to space reasons, detailed experimental setups and results are in Appendix.

\subsection{Experimental Setup}\label{sec:Experimental Environments}
\paragraph{Datasets.}
For simplicity but without loss of generality, we assume the parameterized 1D CDR equations and 2D Helmholtz equations (cf. Eqs.~\eqref{eq:eq_cdr} and~\eqref{eq:helmholtz_eq}). To generate the ground-truth data, we use either analytic or numerical solutions.  
In case of 1D CDR equations, we analyze the target equations with three types of initial conditions $u(x,0)$: two Gaussian distributions of $N(\pi, ({\pi}/2)^2)$ and $N(\pi, ({\pi}/4)^2)$, and a sinusoidal function of $1+\sin(x)$. 
To solve the equation, we use the Strang splitting method \cite{strang1968construction}. For 2D Helmholtz equations, we obtain the exact solution by calculating it directly.

\paragraph{Baseline and Ablation Methods.} We compare \PPINN{} with three baselines. PINN is the original design based on fully-connected layers with non-linear activations in \cite{raissi2019physics}, and PINN-R is its enhancement by using residual connections, which was used in \cite{kim2021dpm}. PINN-seq2seq \cite{krishnapriyan2021characterizing} is a model that applies the seq2seq learning method to the PINN model, sequentially learning data over time. We divided the entire time into 10 steps. In addition, we define one ablation model for our method, called PINN-P, which has the same structure as original PINN, but the {PDE parameters $\pmb{\mu}$} is treated as a coordinate in the parameter space, i.e., $(x, t,\pmb \mu)$. 

\paragraph{Methodology.}We train PINN and PINN-R for each parameter configuration in each equation type, following the standard PINN training method --- in other words, there are as many models as the number of PDE parameter configurations for an equation type. To train \PPINN{}, however, we train it with all the initial conditions and collocation points of the multiple parameter configurations in each equation type, following the training method in Section~\ref{sec:train}. Therefore, we have only one model in each equation type.

\paragraph{Metrics.}To evaluate the performance of the model, we measure the $L_2$ relative and absolute errors between the solution predicted by the model and the analytic solution. The relative error and the absolute error of the $i$-th equation are defined as the averages of $\left\| \hat{\textit{\textbf{u}}_i}-{\textit{\textbf{u}}_i} \right\|_2 / \left\| {\textit{\textbf{u}}_i} \right\|_2$ and $\left\| \hat{\textit{\textbf{u}}_i}-{\textit{\textbf{u}}_i} \right\|_2$, where $i \in \{1,...,N_e\}$ and $N_e$ is the number of equations used for the task. At this time, the errors are measured for each test points and the average value is used. In addition, we use max error and explained variance score for further analysis (cf. Table~\ref{tbl:gauss_max_exp}). We test with 3 seed numbers and report their mean.

\subsection{1D CDR Equations}
In the experiments, we employ 6 different equation types stemmed from CDR equations (cf. Section~\ref{sec:cdr_eq}) with the varying parameters as listed in Table~\ref{tab:dataset}, and the experimental results are summarized in Table~\ref{tbl:result}. Whereas existing baselines show fluctuating performance, our \PPINN{} show stable performance for all the 6 different equation types. The most notable accuracy differences are made for the diffusion, the reaction, the reaction-diffusion, and the convection-diffusion-reaction equations.

For instance, PINN-R marks an absolute error of 0.4320 whereas P$^2$INN achieves an error of 0.0042 for the reaction equations with the coefficient range of 1 to 20, i.e., 102 time smaller error. The smallest accuracy differences happen for the diffusion equations with the coefficient range of 1 to 5. While PINN and P$^2$INN show similar performance, our method much better predict reference solution for the range of 1 to 20, i.e., an error or 0.6782 by PINN vs. 0.1916 by P$^2$INN. Since large coefficients incur equations difficult to solve, all existing baselines commonly fail in the range. In all cases, our method outperforms PINNs, depending on the equation type, by 33\% to 99\% as reported in Table~\ref{tbl:summary_table}. For the reaction equations, the improvement ratio by our method is significant.

\subsubsection{Inferring Solutions of Unseen PDE Parameters}
%Additionally, 
We further evaluate our \PPINN{} in more challenging situations: testing trained models on PDE parameters that are unseen during training, which can be considered as \textit{real-time} \textit{multi-query} scenarios. 

\begin{figure}[t]
    \centering
    \subfigure[$L_2$ absolute error]
    {\includegraphics[width=0.49\columnwidth]{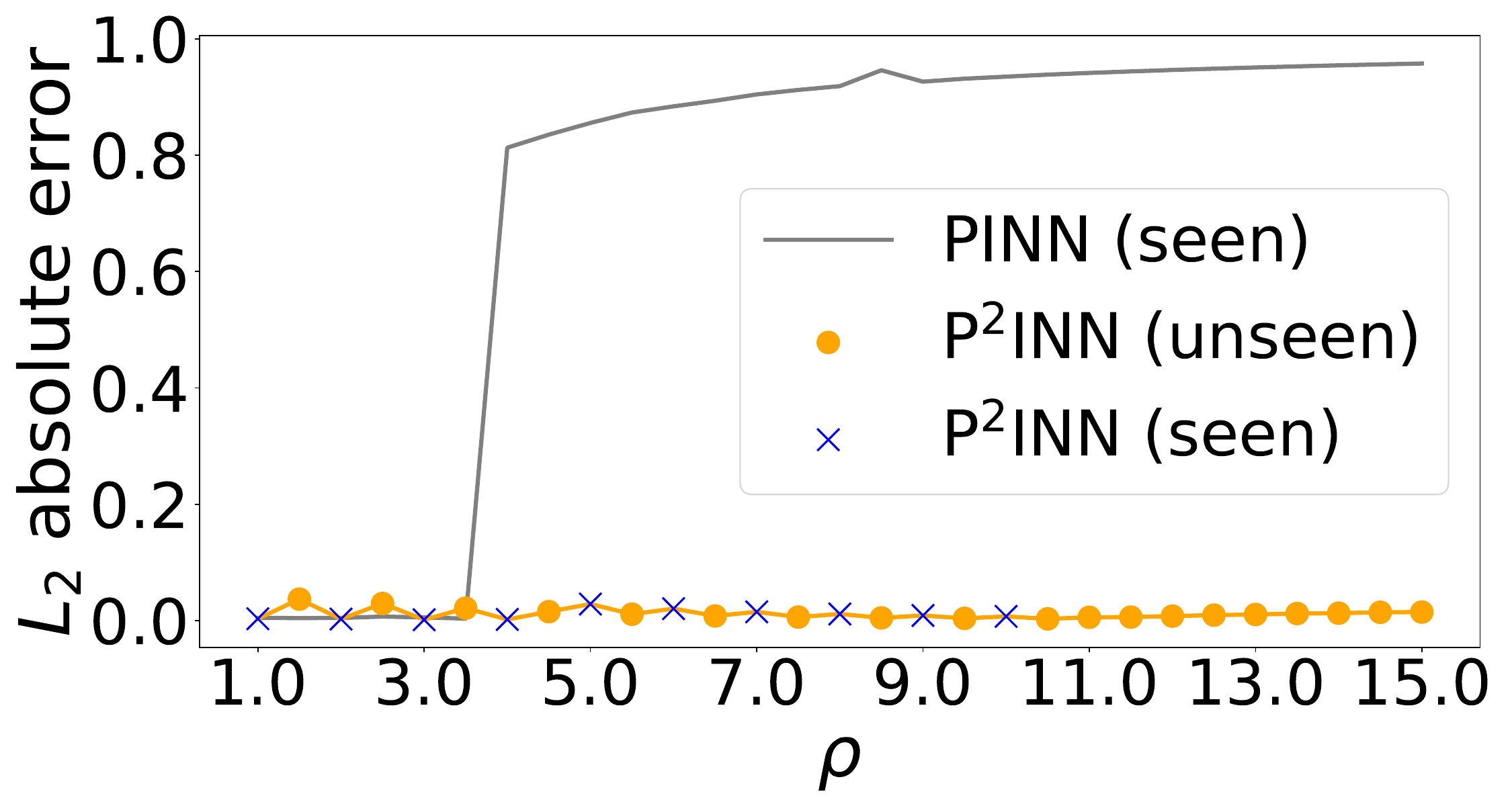}}
    \subfigure[$L_2$ relative error]
    {\includegraphics[width=0.49\columnwidth]{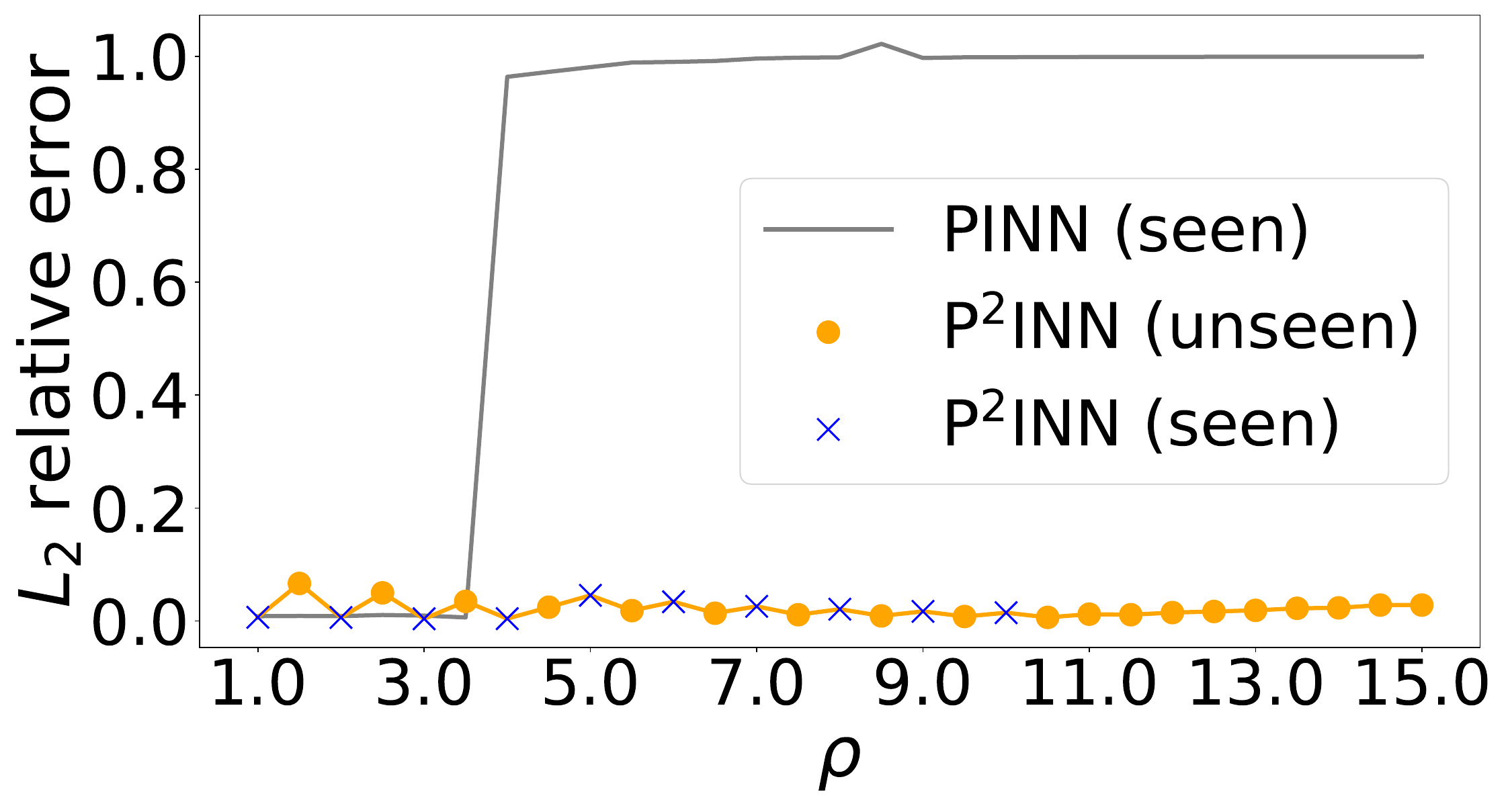}}\\
    \caption{[Reaction equation] Interpolation and extrapolation results for unseen $\rho$.}\label{fig:reac_interp_extrap}
\end{figure}

For reaction equations, we train \PPINN{} on $\rho\in[1, 10]$ with interval 1 and conduct interpolation on $\rho\in[1.5, 9.5]$ with interval 1 and extrapolation on $\rho\in[10.5, 15]$ with interval 0.5. 
As shown in Figure~\ref{fig:reac_interp_extrap}, PINNs' failure for $\rho > 4$ contrasts \PPINN{}' exceptional performance, demonstrating its resilience in extrapolation, not limited good performance only for learned or closely aligned parameters. 

\begin{figure}[t]
    \centering
    \subfigure[Exact]{\includegraphics[width=0.33\columnwidth]{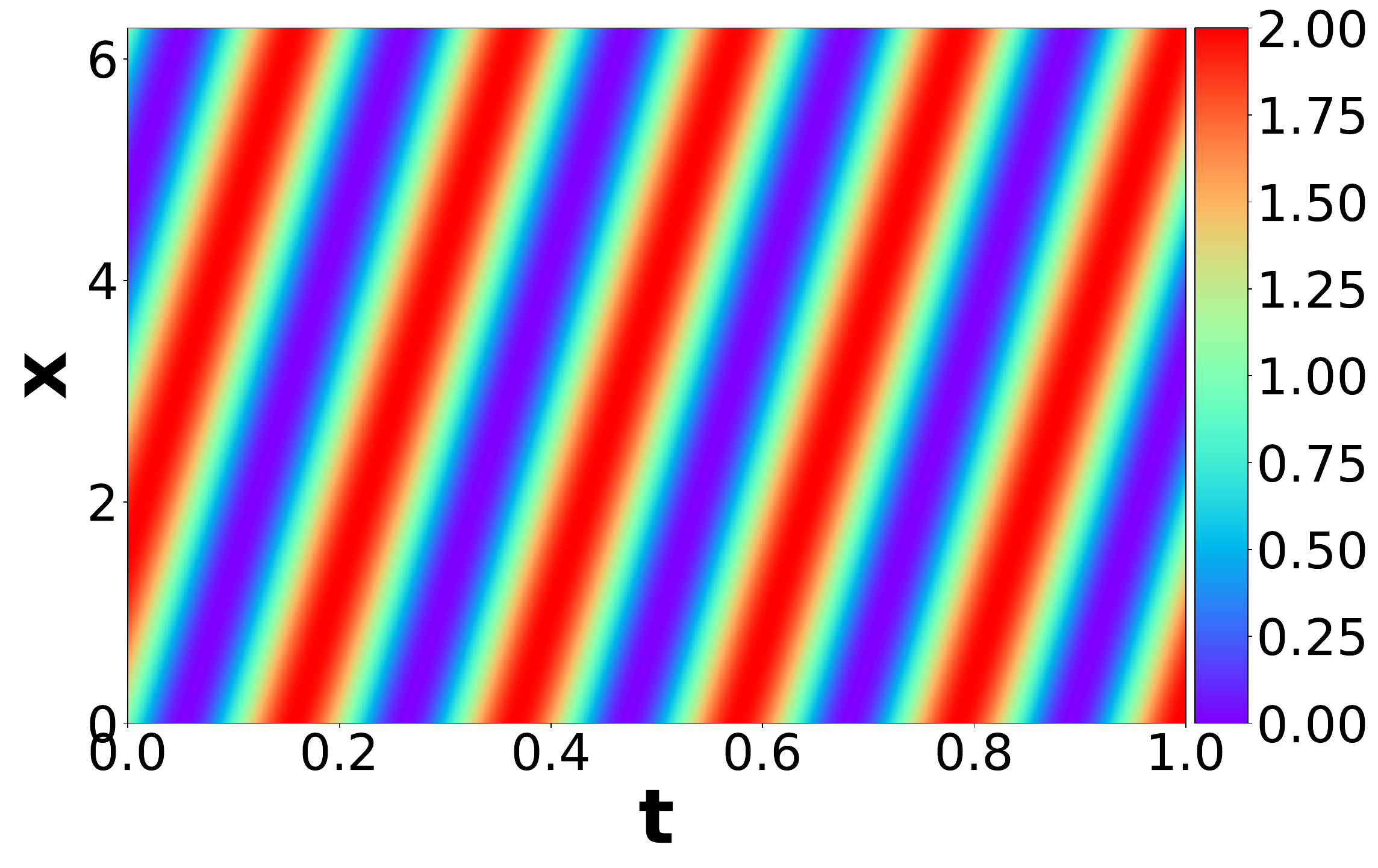}}\hfill
    \subfigure[PINN]{\includegraphics[width=0.33\columnwidth]{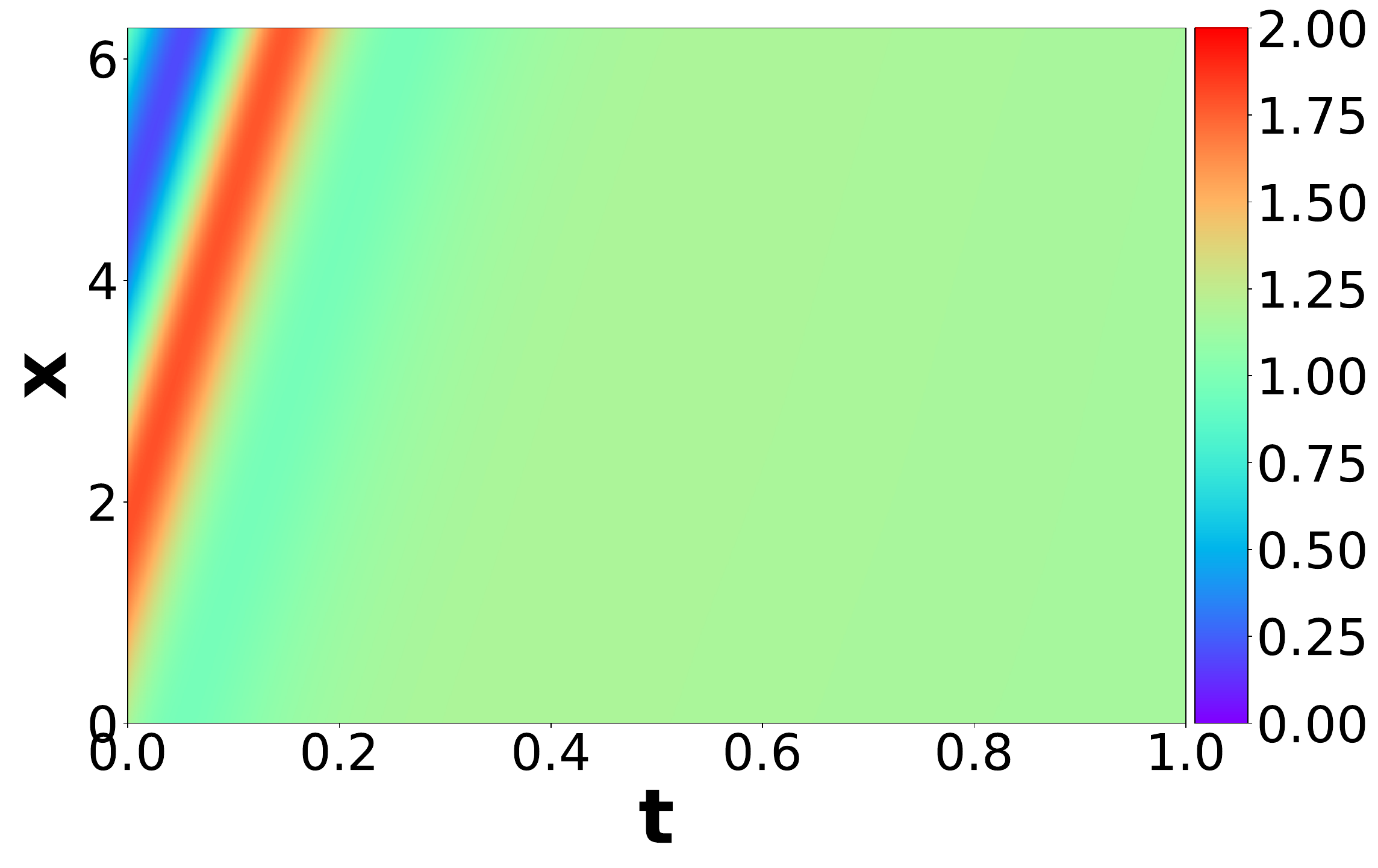}}\hfill
    \subfigure[P$^2$INN]{\includegraphics[width=0.33\columnwidth]{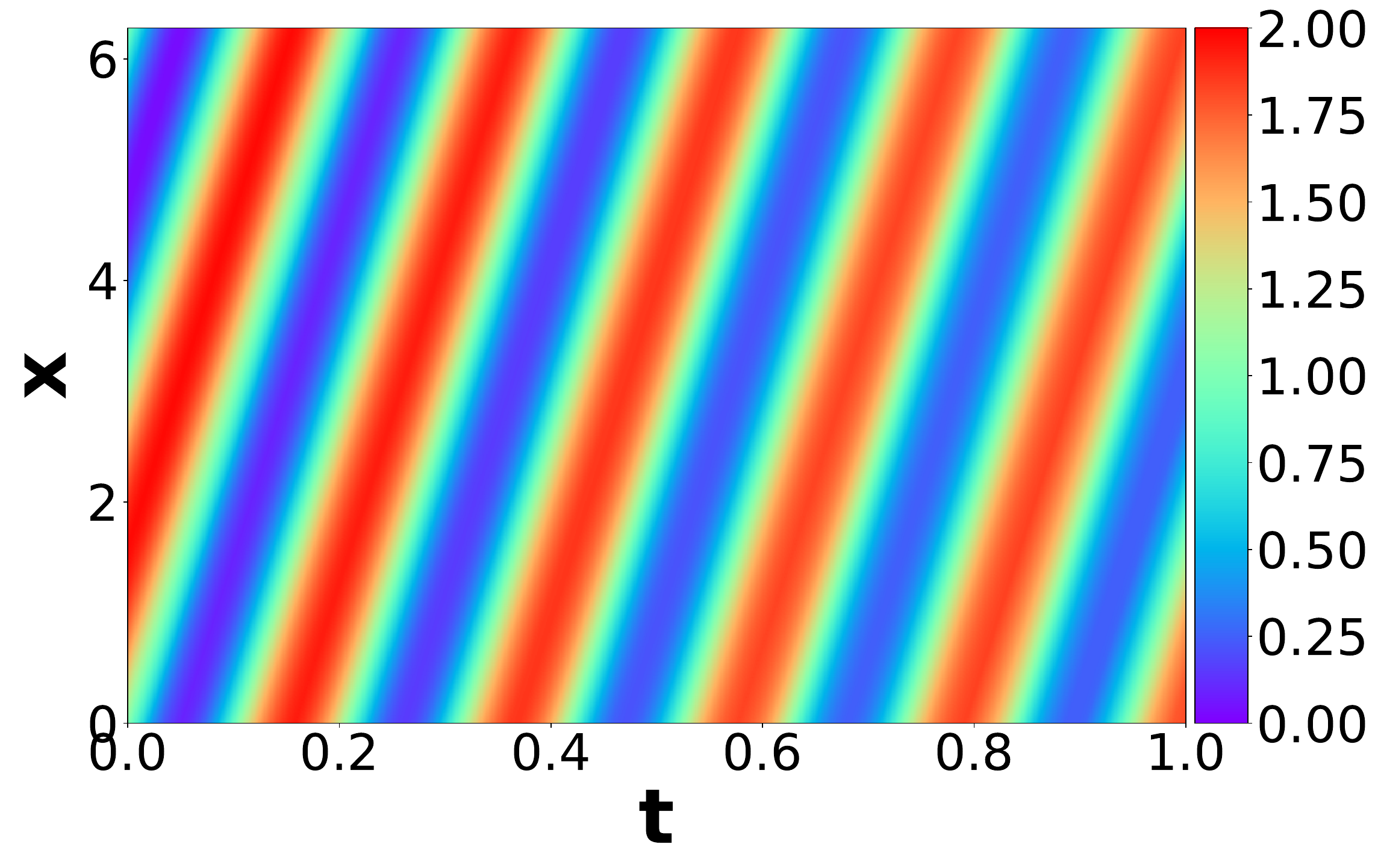}}\\
    
    \subfigure[Exact]{\includegraphics[width=0.33\columnwidth]{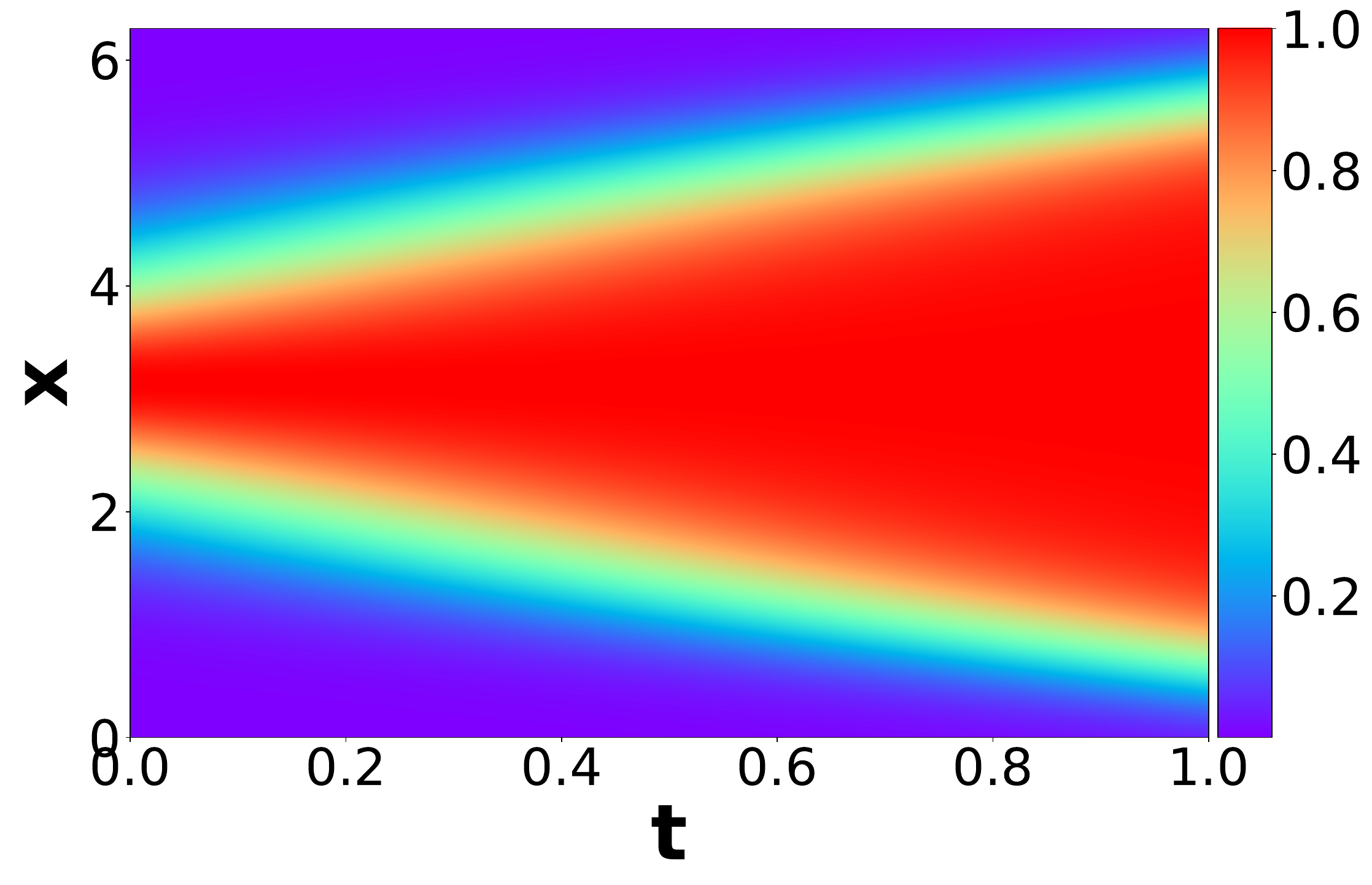}}\hfill
    \subfigure[PINN]{\includegraphics[width=0.33\columnwidth]{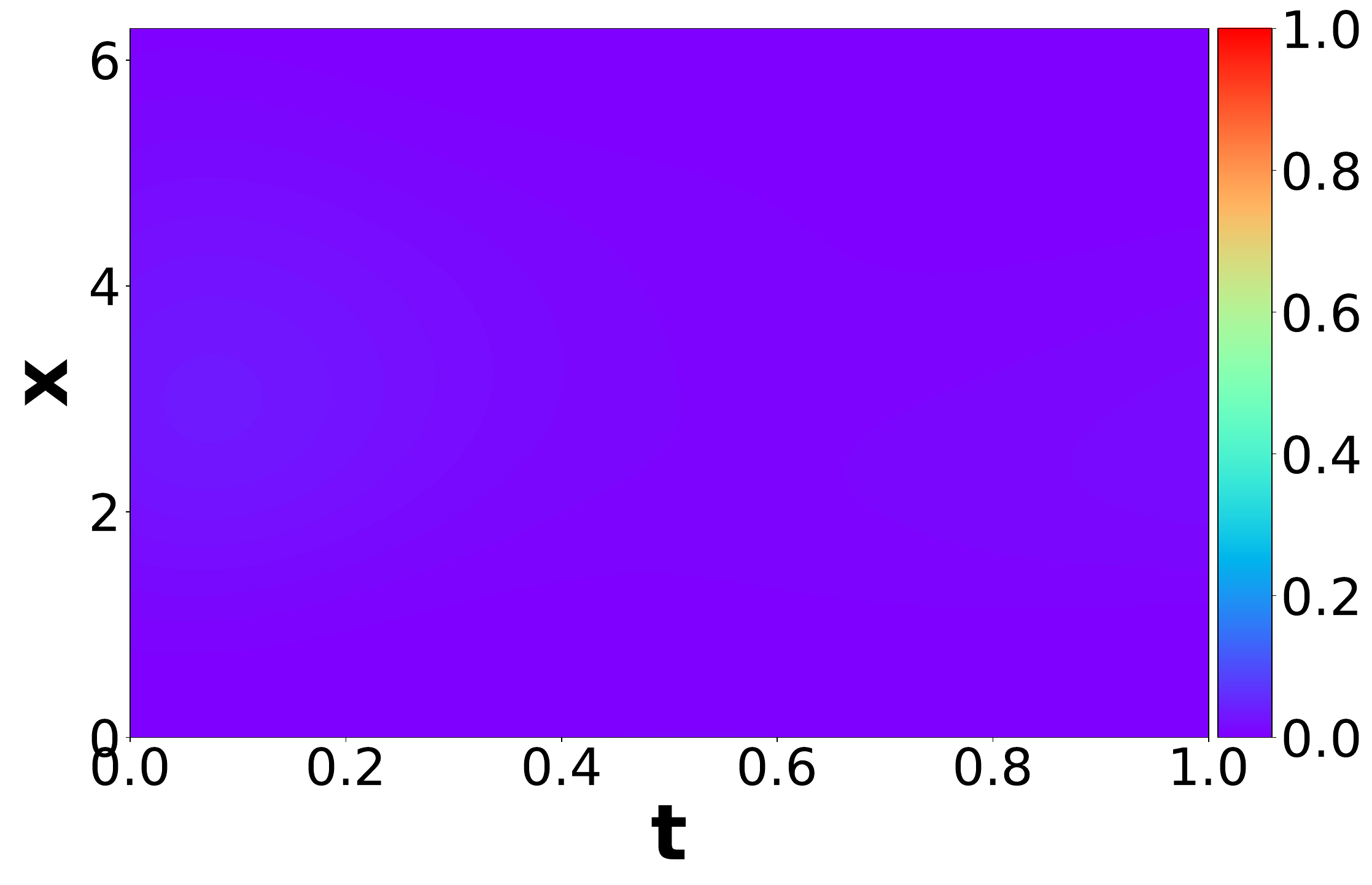}}\hfill
    \subfigure[P$^2$INN]{\includegraphics[width=0.33\columnwidth]{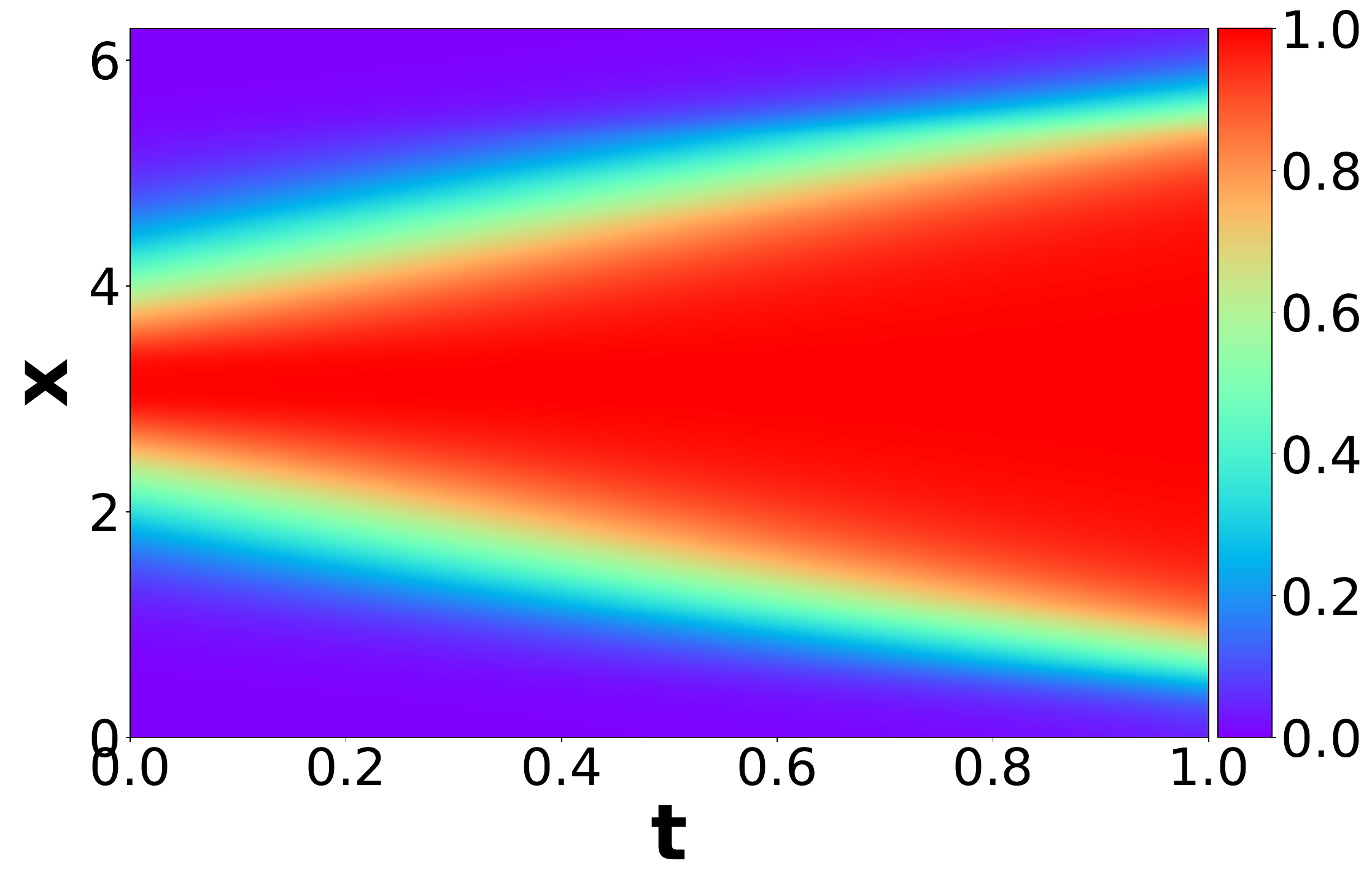}}\\
    
    \captionof{figure}{Failure modes in the convection equation of $\beta = 30$ (a-c), and the reaction equation of $\rho = 5$ (d-f). \PPINN{} much more accurately predict reference solutions.}\label{fig:failure}
\end{figure}

\subsubsection{\PPINN{} in PINN's Failure Modes}\label{sec:failure_mode}
It is well known that PINNs have several failure cases. In particular, CDR equations with large coefficients are notoriously hard to solve with PINNs~\cite{krishnapriyan2021characterizing}. Among the reported failure cases of PINN, the convection equation with $\beta=30$ and the reaction equation with $\rho = 5$ are two representative ones --- in particular, $\beta=30$ corresponds to an extrapolation task after being trained for up to $\beta=20$, which is considered as one of the most challenging task. These two equations generate signals sharply fluctuating over time. As shown in Figure~\ref{fig:failure} and Table~\ref{tbl:failure_finetune}, therefore, PINNs fail to predict reference solutions whereas our method almost exactly reproduce them (cf. Appendix~\ref{a:fine}).

\begin{table}[t]
    \footnotesize
    \caption{Ablation study on the reaction equations.}
    \centering
    \renewcommand{\arraystretch}{0.3}
    \begin{tabular}{lcccc}
    \specialrule{1pt}{2pt}{2pt}
    \multirow{2}{*}{\textbf{Coefficient}} & \multicolumn{2}{c}{\textbf{PINN-P}} & \multicolumn{2}{c}{\textbf{P$^2$INN}}  \\ 
    \cmidrule{2-5}
    \textbf{range} & Abs. err. & Rel. err. & Abs. err. & Rel. err. \\
    \specialrule{1pt}{2pt}{2pt} 
    1$\sim$5  & 0.0083 & 0.0113 & 0.0015 & 0.0027 \\ \midrule
    % 1$\sim$10 & 0.0024 & 0.0049 & 0.0065 & 0.0089 \\\midrule
    1$\sim$20 & 0.8975 & 0.9908 & 0.0042 & 0.0092 \\
    \specialrule{1pt}{2pt}{2pt}
    \end{tabular}\label{tbl:ablation_table}
\end{table}

\begin{table}[t]
\centering
\renewcommand{\arraystretch}{0.8}
\setlength{\tabcolsep}{3.8pt}
\scriptsize
\caption{Experimental results of modulations with an initial condition of
    a Gaussian distribution $N(\pi, ({\pi}/2)^2)$.}
\begin{tabular}{llrrrrr}
\specialrule{1pt}{2pt}{2pt}
\multirow{2}{*}{\textbf{PDE type}}                           & \multirow{2}{*}{\textbf{Metric}}    & \textbf{PINN}  & \multirow{2}{*}{\textbf{P$^2$INN}}
&  \multicolumn{3}{c}{\textbf{Modulation}} \\ 

& & \textbf{(best)} & & \textbf{All} & \textbf{Shift} & \textbf{SVD} \\
\specialrule{1pt}{2pt}{2pt}
\multirow{2}{*}{\textbf{Convection}}        & Abs. err. & 0.0183 & 0.0174 & 0.1959 & 0.0139  & \textbf{0.0138}            \\
                                   & Rel. err. & 0.0327 &    0.0316   & 0.3319 & 0.0248  & \textbf{0.0246}           \\
                                   \midrule

\multirow{2}{*}{\textbf{Reaction}}          & Abs. err. & 0.3336 & 0.0126 & 0.0713 & 0.0095 & \textbf{0.0089}        \\
                                   & Rel. err. & 0.3907 & 0.0229 & 0.1211 & 0.0198 & \textbf{0.0184}           \\ \midrule

\multirow{2}{*}{\textbf{Conv.-Diff.-Reac.}}          & Abs. err. & 0.0865 & 0.0315 & 0.0463 & 0.0321  & \textbf{0.0303}        \\
                                   & Rel. err. & 0.1415 & 0.0508 & 0.0690 & 0.0521  & \textbf{0.0486}        \\
                                   
\specialrule{1pt}{2pt}{2pt}
\end{tabular}\label{tbl:modulation_results}
\end{table}

\subsubsection{Ablation Studies}\label{sec:ablation}
As an ablation study, we do not separately encode $(x,t,\pmb{\mu})$ but directly feed it into our ablation model, PINN-P (i.e., employing a single encoder network $g(x,t,\pmb \mu)$ without explicitly having an encoder for PDE parameters, $g_{\theta_p}(\mu)$). We test on reaction equations, which are hard for PINN baselines to solve, and summarize the results in Table~\ref{tbl:ablation_table}. As shown in Table~\ref{tbl:ablation_table}, especially in a wide coefficient range, \PPINN{} outperforms the ablation model, justifying our model design to separately encode the PDE parameters and the spatial/temporal coordinate. More details of the experiments and other ablation studies are in Appendix.

\subsubsection{P$^2$INN Learned Various Equation Types}\label{sec:CDR_all}
Now, we test the proposed model on a more challenging case, i.e., learning a single solution network for all six different types of CDR equations (cf. Section~\ref{sec:cdr_eq}) at the same time.
%In this section, We train P$^2$INN with a total of six different types of CDR equations (cf. Section~\ref{sec:cdr_eq}) at the same time. Then, 
We compare the performance of our proposed PINN-specific modulation method (cf. Section~\ref{sec:svd_modulation}) against updating all the parameters of the network (denoted as ``All'') and shift modulation~\cite{functa22} (denoted as ``shift''). That is, using P$^2$INN trained with a range of 1 to 5 as a pretrained model, we test how the each method fine-tunes the pretrained model. For the experiment, we fine-tune only for 15 epochs on each type of CDR equations and summarize the results of convection, reaction, and convection-diffusion-reaction equations in Table~\ref{tbl:modulation_results}. The full experimental results are in Appendix~\ref{a:full_mod}.

\begin{figure}[t]
    \centering
    \subfigure[$\beta=3$ (seen)]
    {\includegraphics[width=0.49\columnwidth]{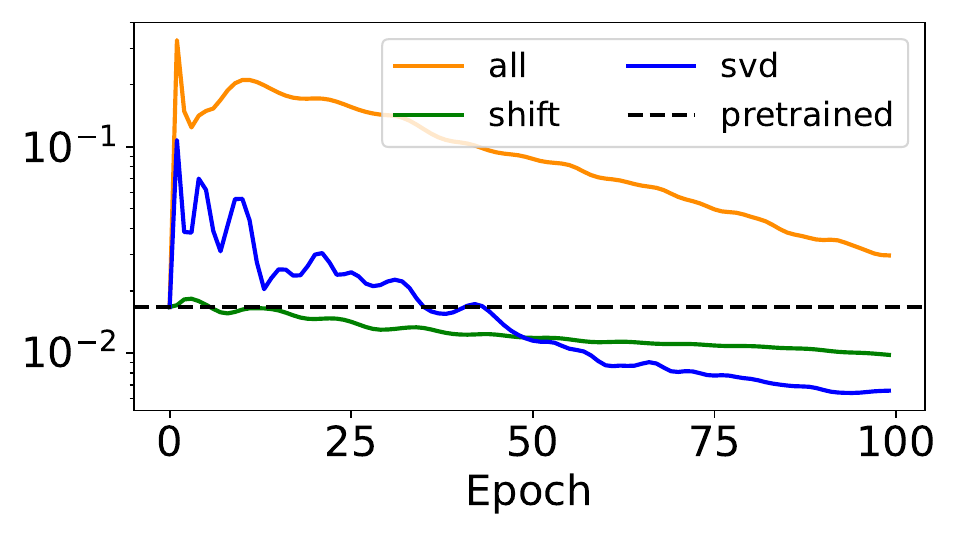}}
    \subfigure[$\beta=8$ (unseen)]
    {\includegraphics[width=0.49\columnwidth]{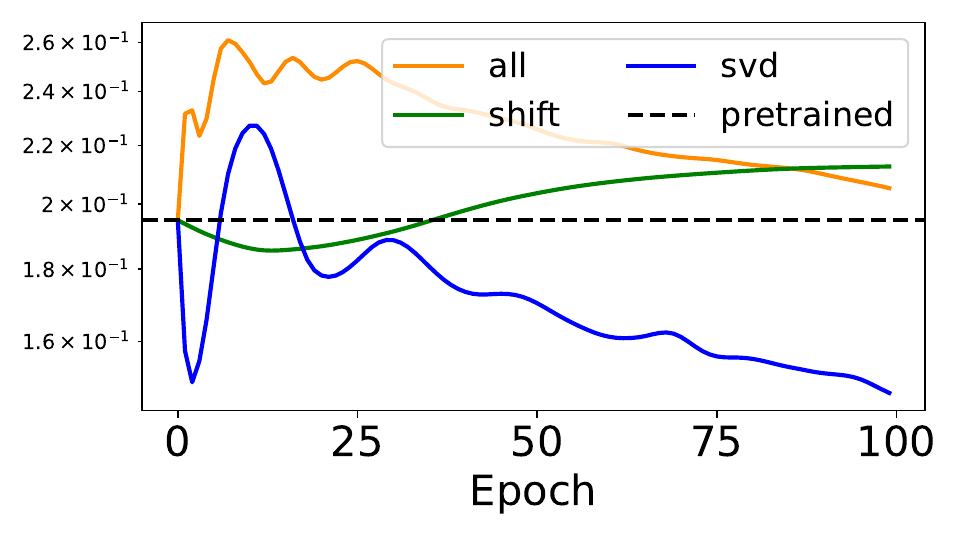}}
\caption{[Convection equation] Comparison of $L_2$ absolute errors based on modulation methods with an initial condition of a Gaussian distribution $N(\pi, ({\pi}/2)^2)$}\label{fig:mod_conv}
\end{figure}

As shown in Table~\ref{tbl:modulation_results}, P$^2$INN accurately approximates and distinguishes the solution for each PDE type even in these challenging scenarios. Moreover, our SVD-based modulation outperforms all other baselines, including shift modulation and the pretrained model. Additionally,  Figure~\ref{fig:mod_conv}, shows how each model infers solutions of seen and unseen PDE parameters at first 100 epochs. For both seen and unseen PDE parameters, SVD-based methods show the lowest $L_2$ absolute errors, proving its generalizability and robustness. In other words, we demonstrate that through our SVD modulation, P$^2$INN can be adapted to various PDEs with small number of trainable parameters and only a few epochs.

\subsection{2D Helmholtz Equations}\label{sec:helmholtz}
\begin{figure}[t]
    \subfigure[Exact]
    {\includegraphics[width=0.25\columnwidth]{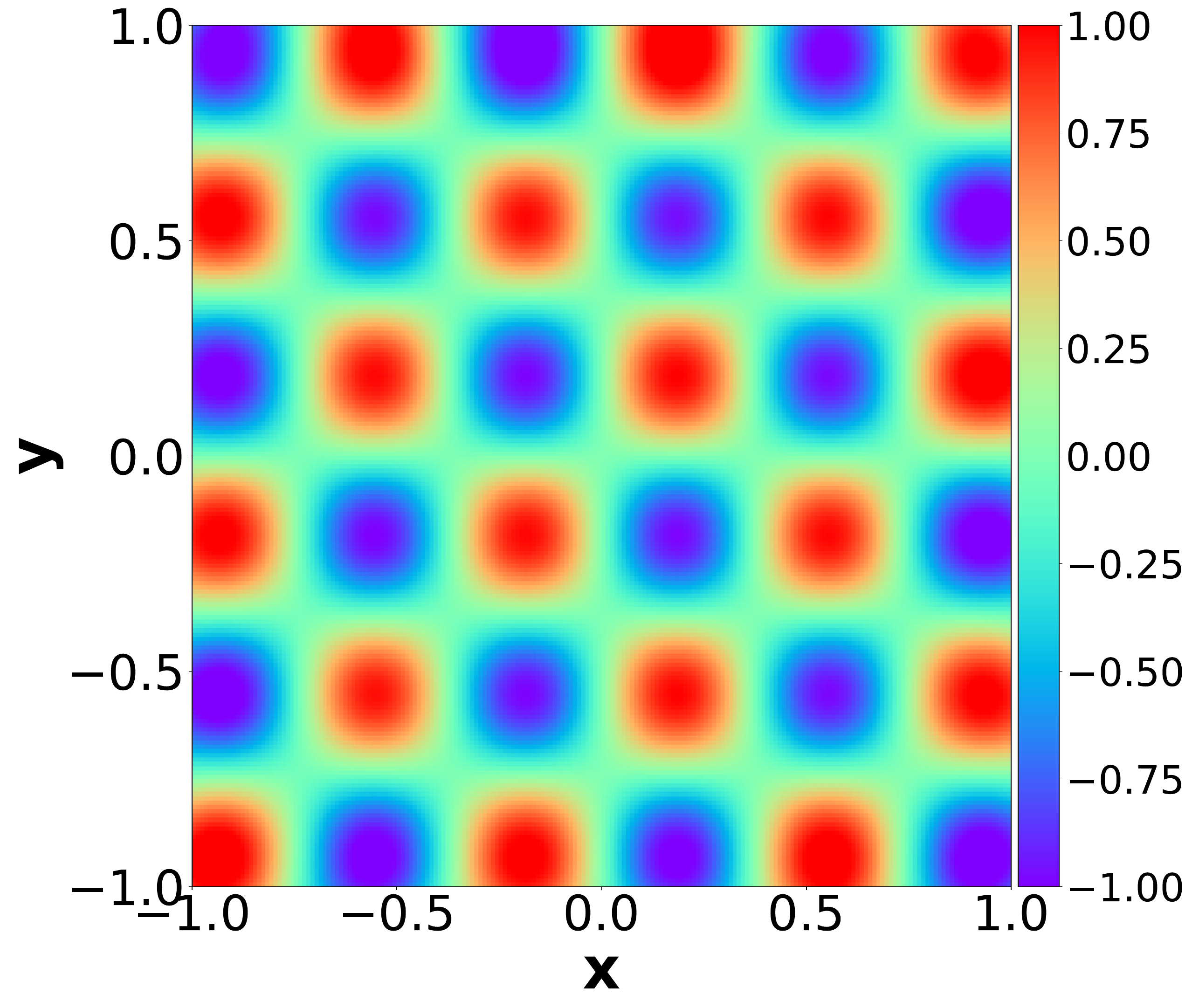}}\hfill
    \subfigure[PINN]
    {\includegraphics[width=0.25\columnwidth]{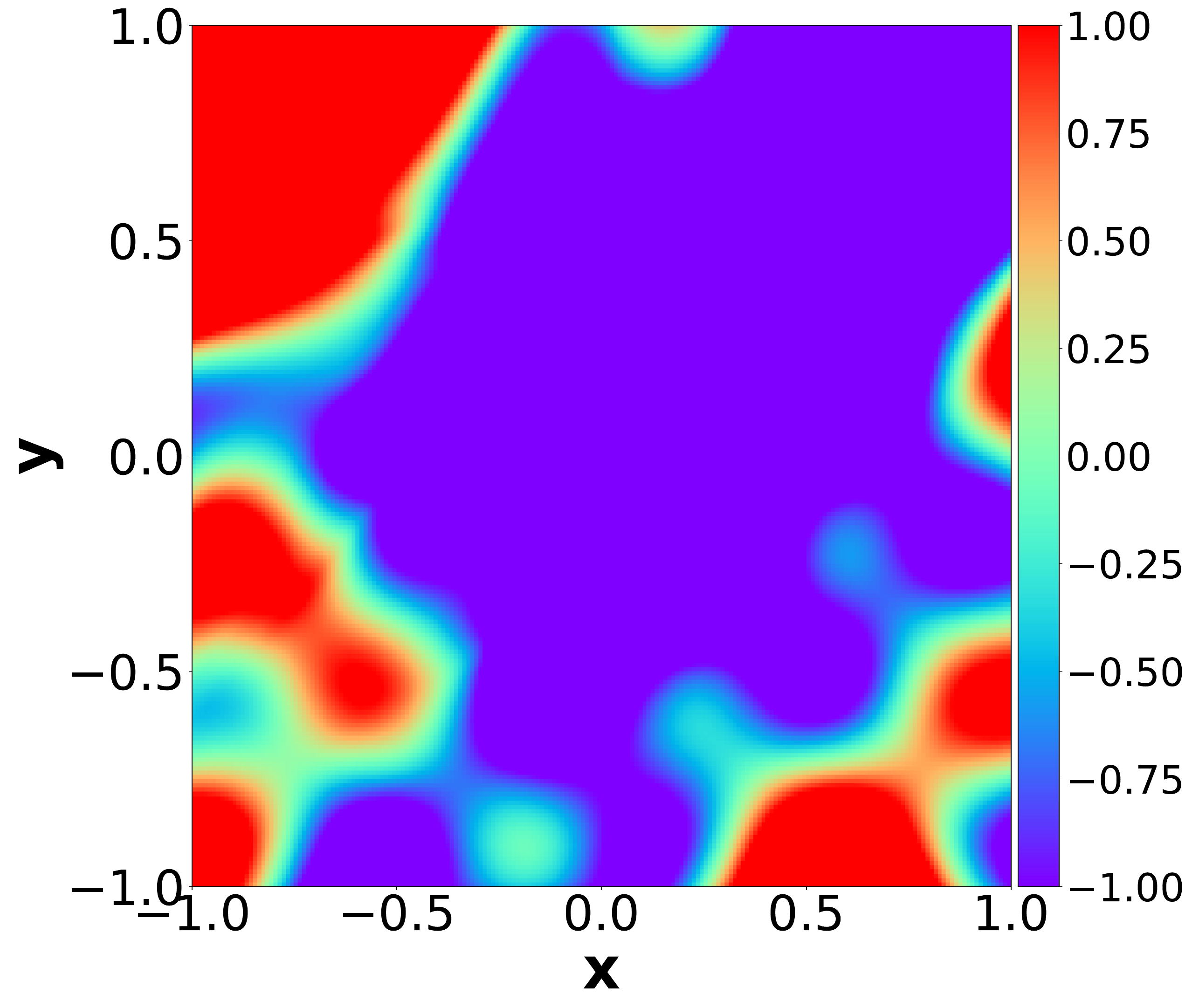}}\hfill
    \subfigure[PINN-R]
    {\includegraphics[width=0.25\columnwidth]{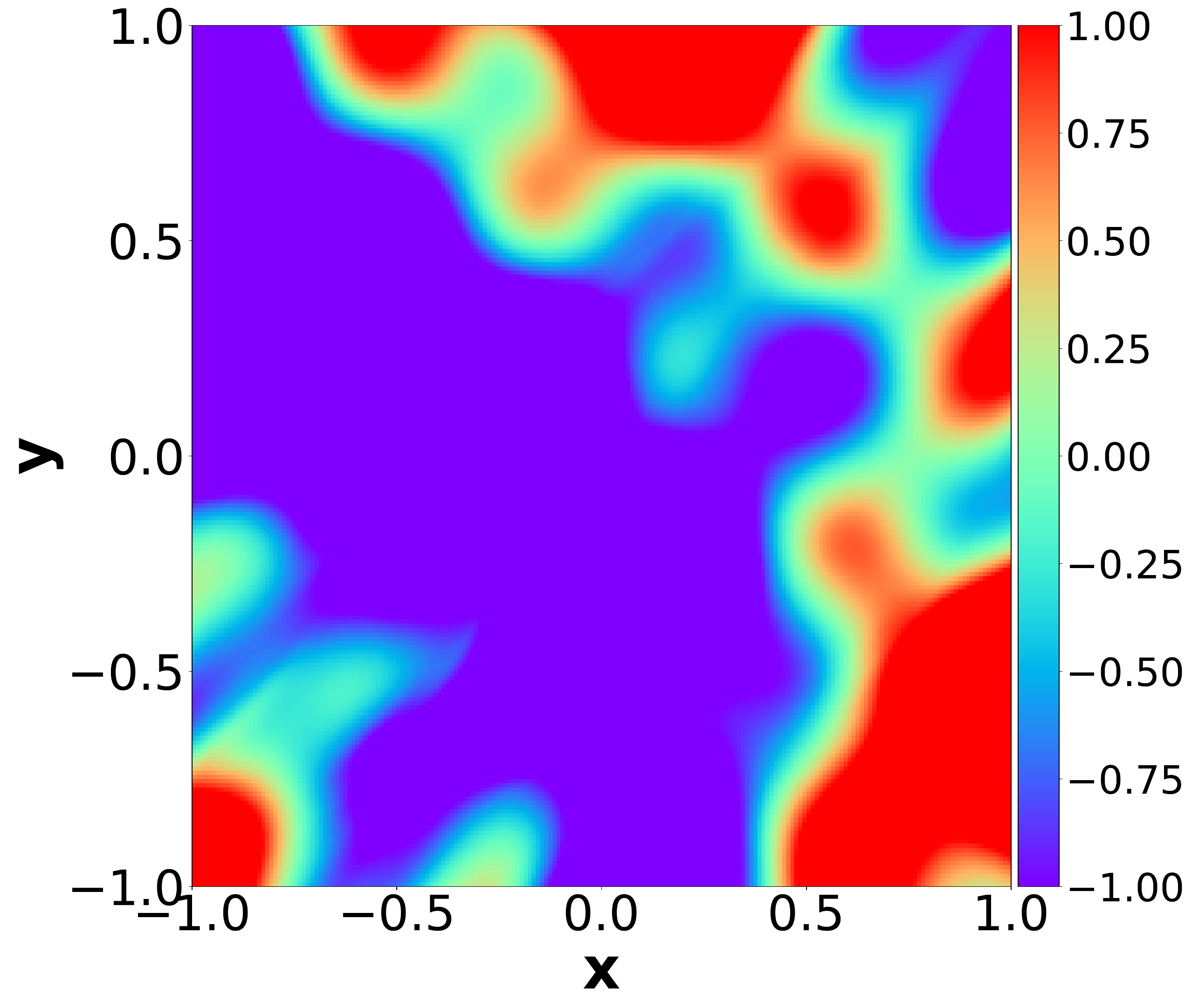}}\hfill
    \subfigure[P$^2$INN]
    {\includegraphics[width=0.25\columnwidth]
    {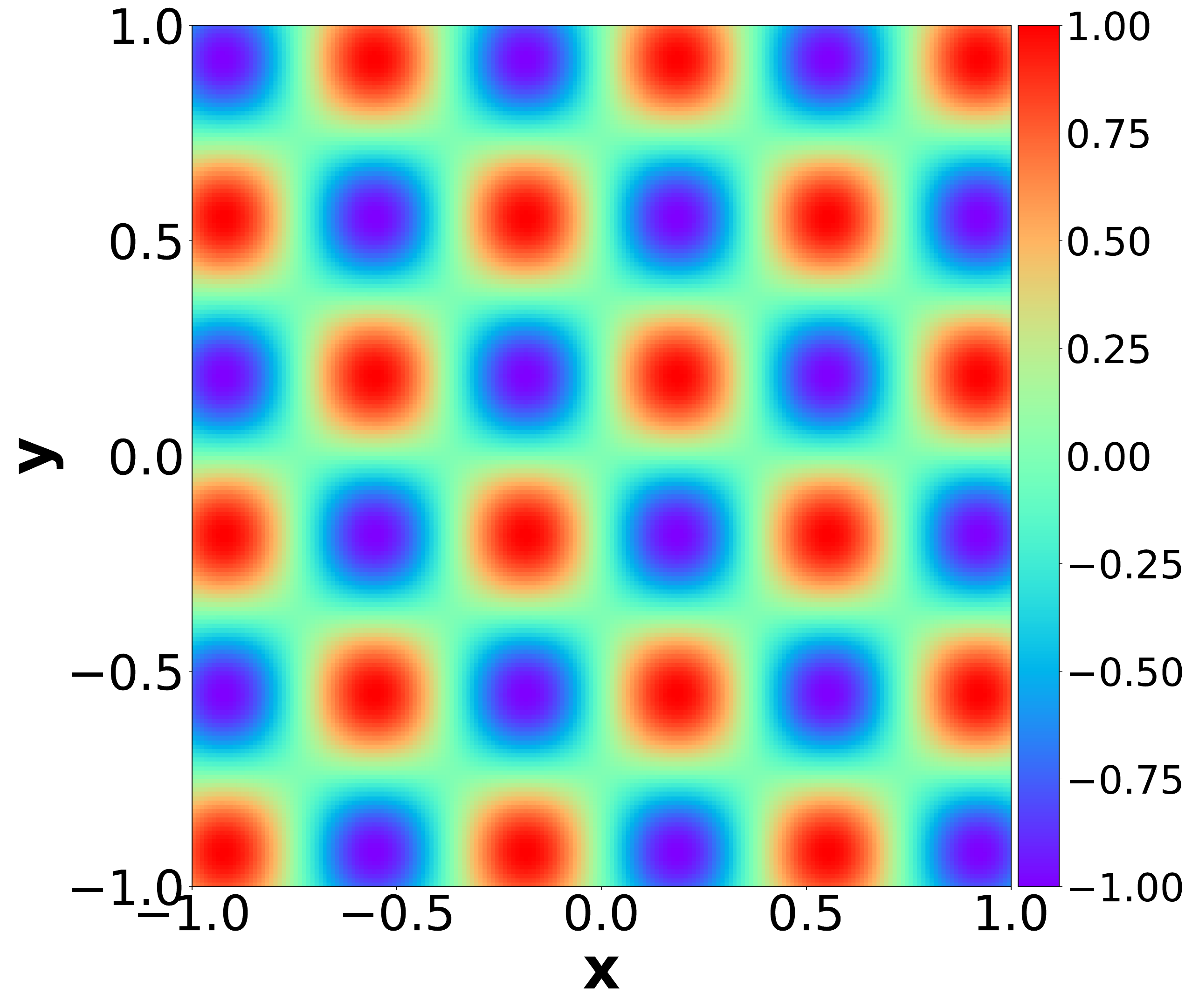}}\\

    \subfigure[Exact]
    {\includegraphics[width=0.25\columnwidth]{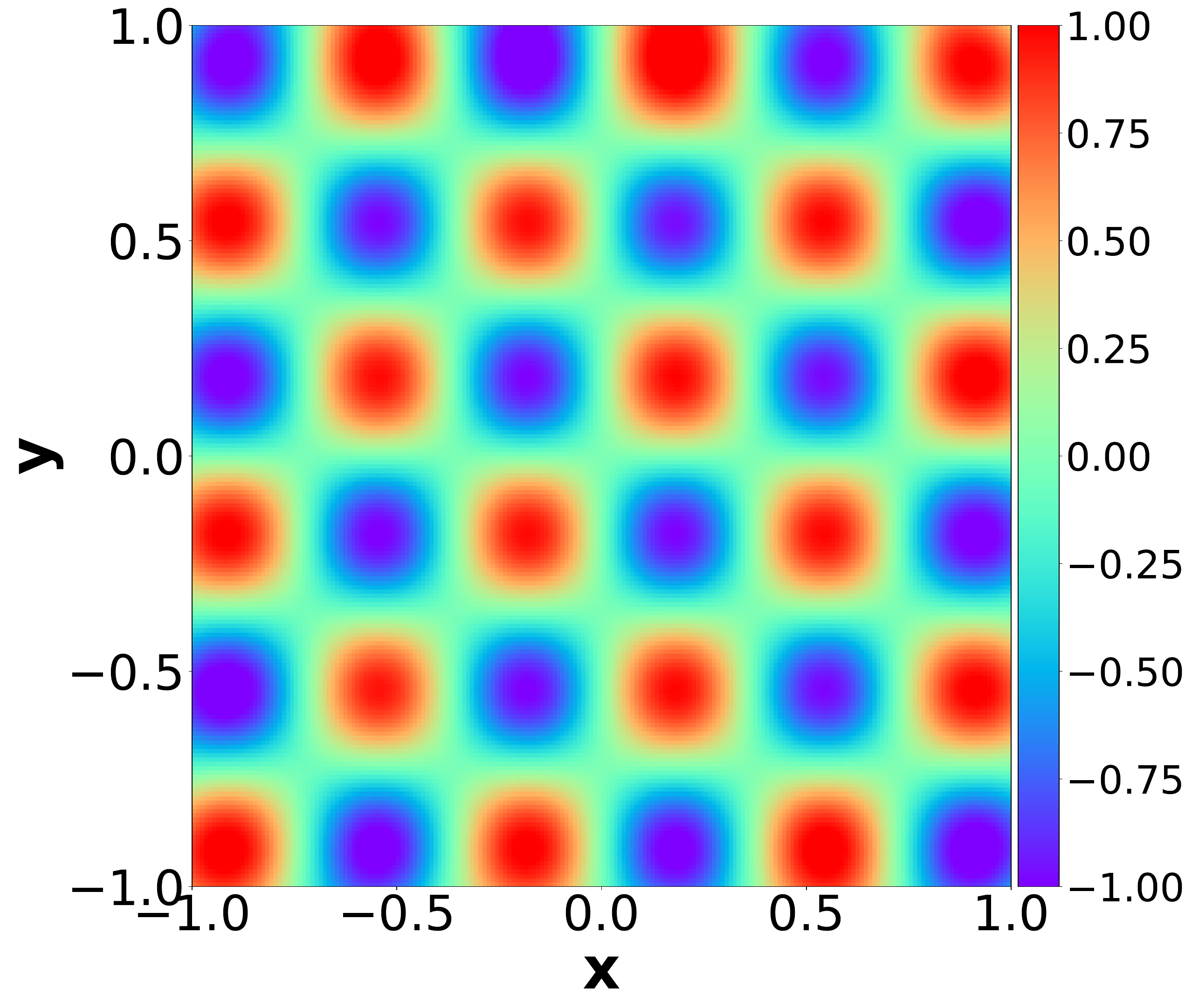}}\hfill
    \subfigure[PINN]
    {\includegraphics[width=0.25\columnwidth]{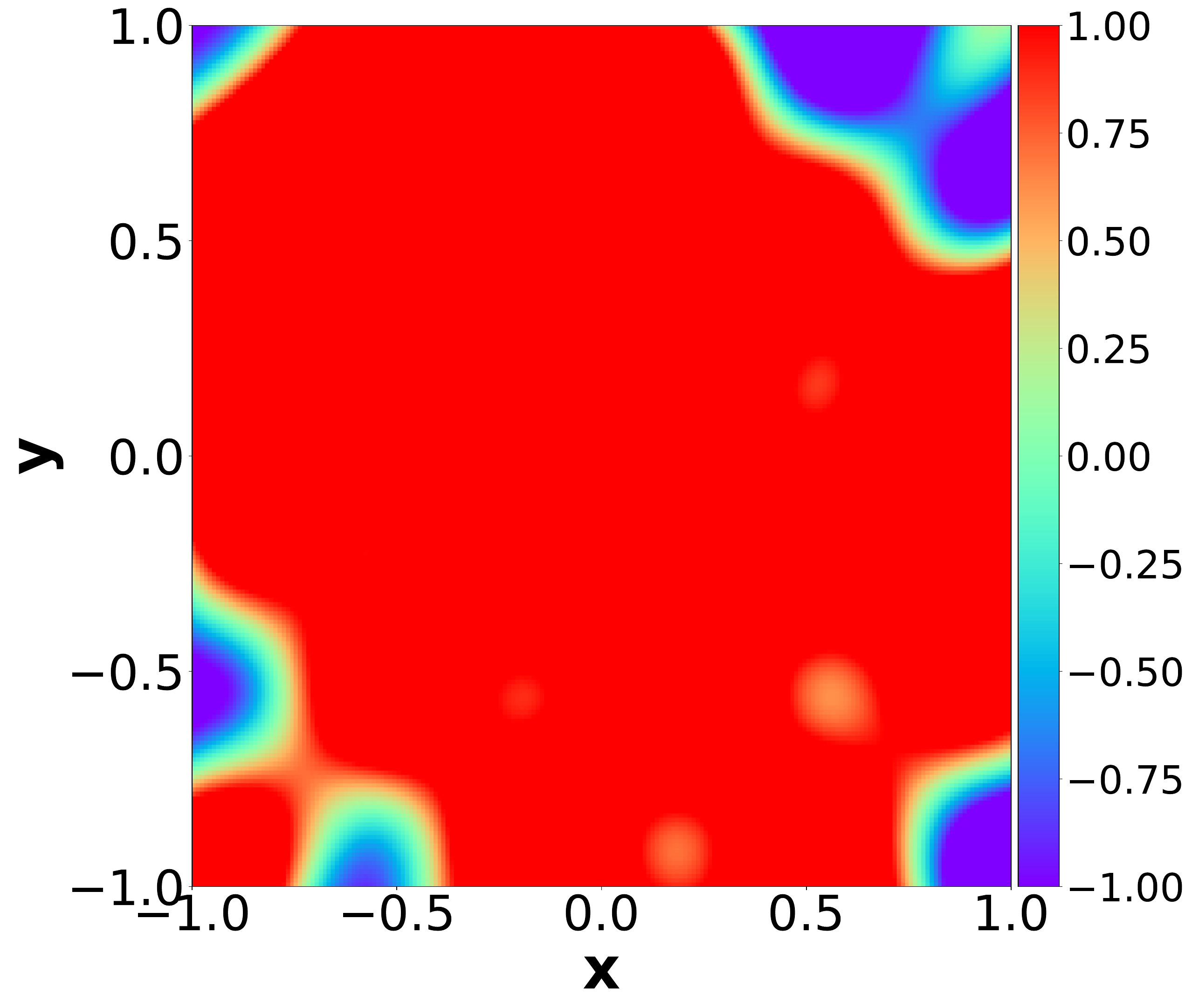}}\hfill
    \subfigure[PINN-R]
    {\includegraphics[width=0.25\columnwidth]{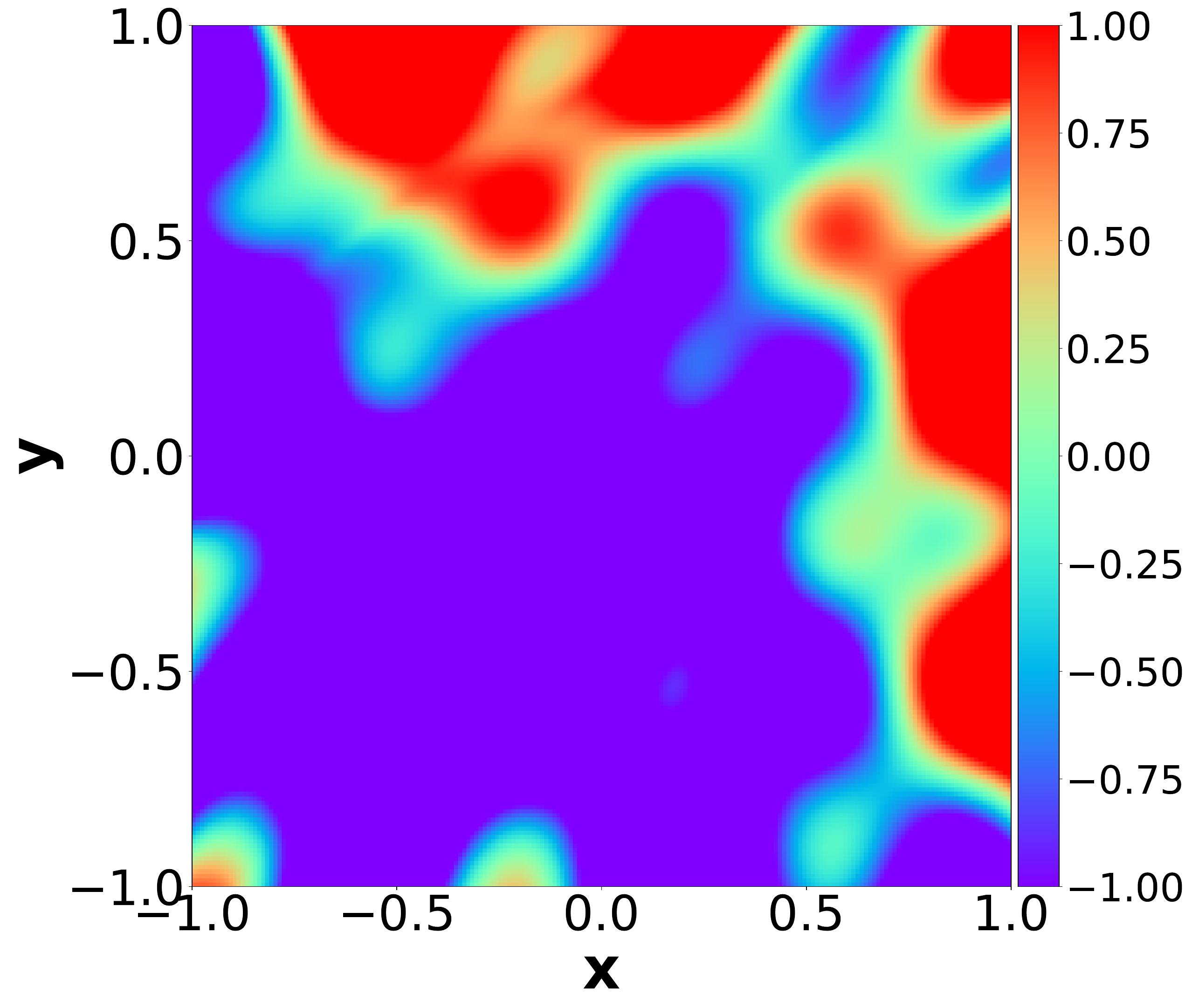}}\hfill
    \subfigure[P$^2$INN]
    {\includegraphics[width=0.25\columnwidth]
    {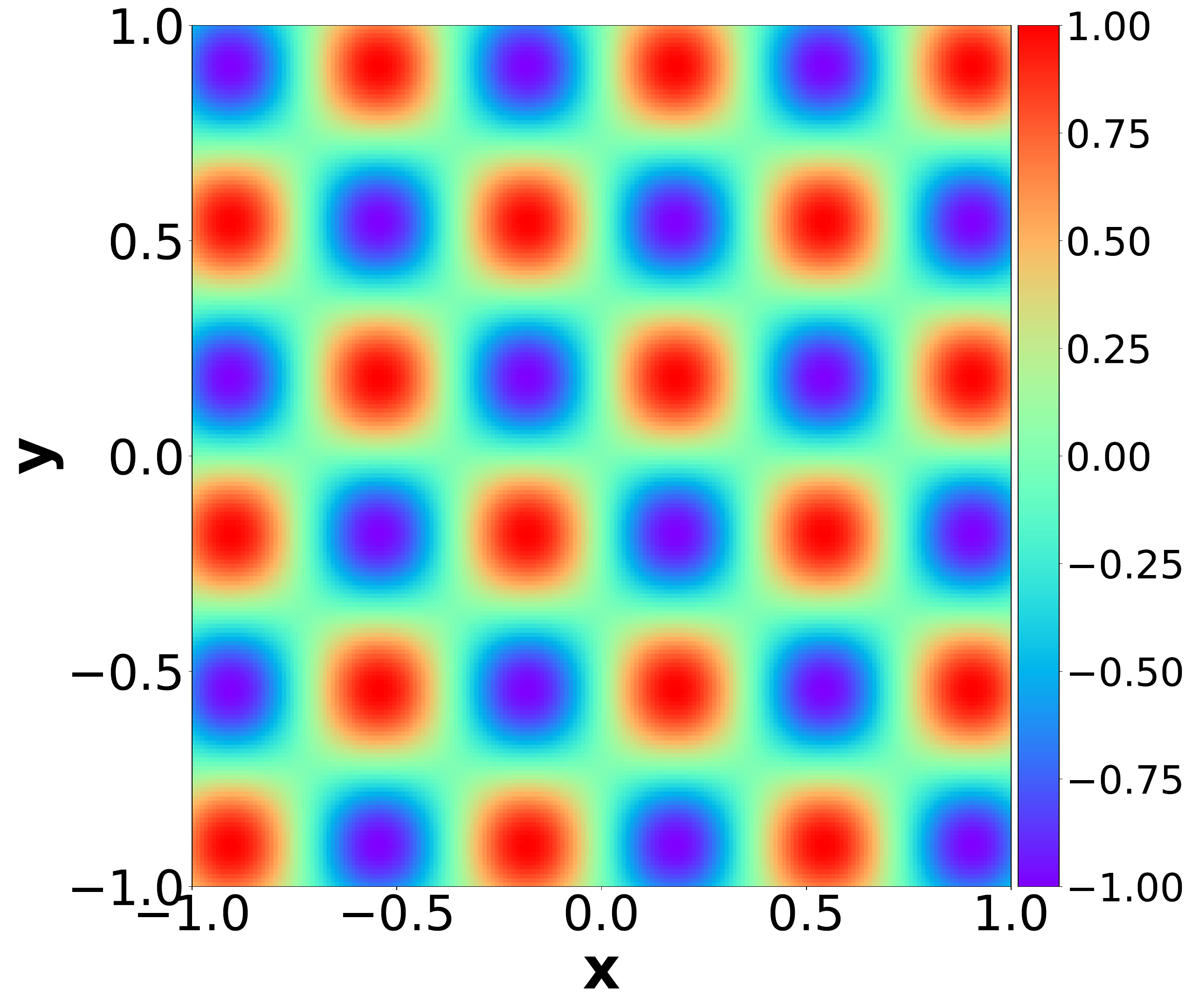}}\\
\caption{[2D-Helmholtz equation] Exact solutions and results of baselines and P$^2$INN for $a=2.7$ (seen) (a-d) and $a=2.75$ (unseen) (e-h)}\label{fig:helmholtz_example}
\end{figure}

For 2D Helmholtz equations, we train models with $a\in[2.5, 3.0]$ with interval 0.1, and then test them with interval 0.05. Notably, as shown in Figure~\ref{fig:helmholtz_example}, \PPINN{} consistently shows good performance in both cases where $a$ is a seen parameter($a=2.7$) and an unseen($a=2.75$) parameter. However, PINN and PINN-R struggle, despite the fact that both of these values are within the seen(trained) parameter range for two models. 

These outcomes underscore our method's capacity to deliver robust solutions, a characteristic that extends to unexplored parameter spaces. Therefore, when solving equations in some coefficient range, \PPINN{} offer exceptional efficiency as they only require testing within the learned latent space, eliminating the need for additional training. On top of that, the experiment on 2D PDEs reaffirms the robustness and efficacy of our proposed P$^2$INN approach in higher-dimensional settings, where there are non-trivial boundary conditions. More results are listed in Appendix~\ref{a:helmholtz}.

\section{Related Work}\label{gen_inst}

% There is plenty of literature dealing with PDEs. We briefly introduce related work mostly used in this work.

\paragraph{Machine Learning Methods for Solving Partial Differential Equations.}
Traditional numerical methods such as finite element methods and finite difference methods have clear pros and cons \cite{patidar2016nonstandard,li1997adaptive,srirekha2010infinite}. The more accurate the results, the more expensive the calculation of numerically approximated formulas. It means that to earn more accurate solutions, it needs to use finer grids, which implies more cost. To alleviate those cons, researchers were attracted to machine learning approaches \cite{karniadakis2021physics,https://doi.org/10.48550/arxiv.2201.05624}. 
After various trials like using the Galerkin or Ritz method \cite{rudd2015constrained}, PINNs proposed a transformative way of using deep learning for solving general governing PDEs in a physically sound and easy-to-formulate computational formalism \cite{raissi2019physics}. As elaborated above, PINNs, however, possess weaknesses which must be addressed \cite{krishnapriyan2021characterizing}: (1) there are classes of PDEs that it is difficult for PINNs to learn (e.g., PDEs exhibiting high oscillation or sharp transitions in spatial and/or temporal domains) and (2) gradient-based training often converges to a local optimum of models. Another line of research for solving PDEs is to analyze operator learning for differential equation or deep Ritz methods \cite{yu2018deep,li2020fourier,gupta2021multiwavelet} but PINNs still have its potential for mainly focusing on governing equations which describe physical phenomena.

\paragraph{Physics as Inductive Biases.} 
There have been various strategies to impose physical constraints on neural networks \cite{LagrangianNN,rudd2015constrained,lee2021machine}. Most of them focus on imposing constraints for outputs or injecting specific physical conditions into neural networks. As a simple but effective solution, PINNs directly impose physical conditions into neural networks by using a governing equation itself as a loss~\cite{raissi2019physics}. This loss function is called $L_f$. In this way, PINNs can learn the residual error of the governing equation. If initial conditions are given, we can add an initial error loss term $L_u$. Furthermore, if there are specific boundary conditions, we can specify boundary conditions in $L_b$.

\paragraph{Recent Developments in PINNs.}
In the recent literature, PINNs have evolved in many different ways to resolve issues inherent with the vanilla PINNs. Some architectural enhancements have been made in \cite{cho2024hypernetwork} (a low-rank extension PINNs for model efficiency and a hypernetwork for effective training) and  in \cite{cho2024separable} (a separable design of model parameters for efficient training). A systematic assessment for PINNs and a new sample strategie have been investigated in  PINNACLE \cite{lau2023pinnacle}. There have been some effort to combine PINNs with symbolic regression in universal PINNs \cite{podina2023universal} and to devise a preconditioner for PINNs from an PDE operator preconditioning perspective \cite{de2023operator}. Lastly, novel optimizers for effective training of PINNs have been proposed in  \cite{yao2023multiadam} (MultiAdam) and \cite{muller2023achieving} (based on natural gradient descent).

\section{Conclusions}
PINN is a highly applicable and promising technology for many engineering and scientific domains. In particular, it has the strength that training is possible only with a PDE to be solved, without any additional data. However, due to the highly nonlinear characteristic of PDEs, PINNs show very poor performance in certain parameterized PDE problems. In addition, there is a weakness that the model must be re-trained from scratch to analyze a new PDE. To solve these chronic issues, we propose parameterized physics-informed neural networks (\PPINN{}), which can learn similar parameterized PDEs simultaneously. Through this approach, it is possible to overcome the failure situation of PINNs that could not be solved in previous studies. To ensure that it is effective in general cases, we use more than thousands of CDR equations and show that \PPINN{} outperform baselines in almost all cases of the benchmark PDEs.

\section*{Impact Statement}
Our method can significantly reduce the energy consumption in training PINNs for many equations. When solving for an one trivial equation only, however, existing PINN methods show better efficiency.

\section*{Acknowledgements}
This work was partly supported by Samsung Electronics Co., Ltd. (No. G01240136, KAIST Semiconductor Research Fund (2nd)), the Korea Advanced Institute of Science and Technology (KAIST) grant funded by the Korea government (MSIT) (No. G04240001, Physics-inspired Deep Learning), and Institute for Information \& Communications Technology Planning \& Evaluation (IITP) grants funded by the Korea government (MSIT) (No. RS-2020-II201361, Artificial Intelligence Graduate School Program (Yonsei University)). K. Lee acknowledges support from the U.S. National Science Foundation under grant IIS 2338909. 

\bibliography{icml2024}
\bibliographystyle{icml2024}

%%%%%%%%%%%%%%%%%%%%%%%%%%%%%%%%%%%%%%%%%%%%%%%%%%%%%%%%%%%%%%%%%%%%%%%%%%%%%%%
%%%%%%%%%%%%%%%%%%%%%%%%%%%%%%%%%%%%%%%%%%%%%%%%%%%%%%%%%%%%%%%%%%%%%%%%%%%%%%%
% APPENDIX
%%%%%%%%%%%%%%%%%%%%%%%%%%%%%%%%%%%%%%%%%%%%%%%%%%%%%%%%%%%%%%%%%%%%%%%%%%%%%%%
%%%%%%%%%%%%%%%%%%%%%%%%%%%%%%%%%%%%%%%%%%%%%%%%%%%%%%%%%%%%%%%%%%%%%%%%%%%%%%%

\appendix
\onecolumn

\section{Datasets}\label{sec:data_detail}
\subsection{1D Convection-Diffusion-Reaction Equations}
Each of these individual PDEs has their own importance and has been studied extensively: 

\begin{compactenum}
    \item Convection-diffusion equations are used in fluid dynamics, particle chemistry, computational finance, and so on,
    \item Reaction-diffusion equations are popular in the domain of biophysics and mathematical biology,
    \item Convection equations, diffusion equations, and reaction equations are for describing transport, diffusive, and reactive phenomena, respectively in simplified settings. 
\end{compactenum}

In total, there are six classes of Convection-Diffusion-Reaction equations, each of which has its own importance in science. For each dataset, we select 1,000 collocation points, 256 initial points, 100 boundary points, and 1,000 test points.
\subsection{2D Helmholtz Equations}

\begin{equation}
    % \begin{align}
        \begin{split}
        \frac{\partial^2 u(x,y)}{\partial x^2} + \frac{\partial^2 u(x,y)}{\partial y^2} + k^2 u(x,y) -q(x,y) = 0 \\
        q(x,y)= (-(a_1 \pi)^2-(a_2 \pi)^2+k^2)\sin(a_1 \pi x) \sin(a_2 \pi y)
        \end{split}
    \label{eq:helmholtz_eq}
    % \end{align} 
\end{equation}

\begin{align}
    u(x,y) = k^2 \sin(a_1 \pi x) \sin(a_2 \pi y).
\label{eq:helmholtz_eq_sol}
\end{align}

We employ the specific Helmholtz equations which were used in \citep{mcclenny2020self} as benchmark PDEs (cf. Eq.~\eqref{eq:helmholtz_eq}), and it can be directly solved as Eq.~\eqref{eq:helmholtz_eq_sol}.
The Helmholtz equations describe the behavior of state variable $u(x, y)$ in a 2D space, accounting for the effects of wave propagation, and a source term represented by $q(x,y)$. Here $k$ is a parameter related to wave frequency, while $a_1$ and $a_2$ control the spatial variations of the source term. In our experiments, we set $k$ to $1$ and the parameters $a_1$ and $a_2$ to a common value $a$. For each dataset, we select 1,000 collocation points, 400 boundary points, and 100 test points.

\section{More Details on Experimental Setup}\label{sec:experiment_setup}

\subsection{Loss}\label{app:loss}
With the prediction $\hat u$ produced by \PPINN{}, our basic loss function consists of three terms as follows:
\begin{align}
 L({\Theta}) = {w_1}L_u + {w_2}L_f + {w_3}L_b, \label{eq:loss_appnedix}
\end{align}

and $L_u$, $L_f$ and $L_b$ are defined as follows:
\begin{align}
 L_u &= {1 \over N_{u}}\sum_{N_{u}}\Big(\hat{u}(x,0;\pmb{\mu})-u(x,0;\pmb{\mu})\Big)^2, \label{eq:loss_L_u}\\
 L_f &= {1 \over N_{f}}\sum_{N_{f}}\Big(\mathcal{F}(x, t, \hat{u};\pmb{\mu}) \Big)^2, \label{eq:loss_L_f}\\ 
 L_b &= {1 \over N_{b}}\sum_{N_{b}}\Big(\hat{u}(0,t;\pmb{\mu})-\hat{u}(2\pi,t;\pmb{\mu})\Big)^2, \label{eq:loss_L_b}
\end{align}

where ${N_{u}}$, ${N_{f}}$, and $N_{b}$ are the cardinalities of the sets of initial conditions, collocation points, and boundary conditions; $w_1, w_2, w_3 \in \mathbb{R}$ are hyperparameters. The first and the second terms denote the data matching loss $L_u$ and the PDE residual loss $L_f$, respectively. In addition, we separately add the boundary condition term $L_b$, forcing their values equal at both top and bottom parts (see Figures~\ref{fig:figure1} and~\ref{fig:figure2}).

\subsection{Baseline}\label{sec:baseline_appendix}
Each baseline and ablation model is trained in the following way:
\begin{compactenum}
\item PINN, PINN-R, and PINN-seq2seq do not read PDE parameters, such as $\beta, \nu, \rho$ and $a$, but are trained separately for each of the coefficient settings.
\item PINN-P, an ablation model of \PPINN{}, is able to process PDE parameters and is trained for all coefficient settings in each equation type.
\item Therefore, PINN, PINN-R, and PINN-seq2seq require many trained models for solving parameterized PDEs whereas PINN-P and our method require a single trained model to solve them.
\end{compactenum}

\subsection{Implementation}

% { [From main paper.. should reorganize later] 
We implement \PPINN{} with \textsc{Python} 3.7.11 and \textsc{Pytorch} 1.10.2 that supports \textsc{CUDA} 11.4.
We run our evaluation on a machine equipped with Intel Core-i9 CPUs and \textsc{NVIDIA RTX A6000} and \textsc{RTX 2080 Ti} GPUs.

\section{Model Configuration and Efficiency}\label{a:hyperparam}

\subsection{Dataset Statistics}

\begin{table}[h]
\centering
\footnotesize
\caption{Dataset statistics. For each equation type, we test three different coefficient ranges. In Conv.-Diff.-Reac., $\beta, \nu, \rho$ are non-zeros.}
\renewcommand{\arraystretch}{0.8}
\begin{tabular}{lrrrrrrrr}
\specialrule{1pt}{2pt}{2pt}
 % \multirow{2}{*}{\textbf{Coefficient}} & \multicolumn{6}{c}{\textbf{Single type}} & & \textbf{All types}
% \\ \cmidrule{2-7} \cmidrule{9-9}
 \textbf{Coefficient range}& \textbf{Convection} & \textbf{Diffusion} & \textbf{Reaction} & \textbf{Conv.-Diff.} & \textbf{Reac.-Diff.} & \textbf{Conv.-Diff.-Reac.} \\
\specialrule{1pt}{2pt}{2pt}
\textbf{1$\sim$ 5} & 5 & 5 & 5 & 25 & 25 & 125 \\ 
\midrule
\textbf{1$\sim$10} & 10 & 10 & 10 & 100 & 100 & 1,000 \\ \midrule
\textbf{1$\sim$20} & 20 & 20 & 20 & 400 & 400 & 8,000 \\ 
\specialrule{1pt}{2pt}{2pt}
\end{tabular}

\label{tab:dataset}
\end{table}

Table~\ref{tab:dataset} represents dataset statistics, and our dataset generation source code is mainly based on~\citep{krishnapriyan2021characterizing}.

\subsection{Model Efficiency and Hyperparameters}
Our baselines, PINN, PINN-R, and PINN-seq2seq, are designed with 6 layers, and the dimension of hidden vector is 50.
For training, we employ Adam optimizers with learning rate of $1e-3$.
For our method, we set $D_p, D_c,$ and $D_g$ to 4, 3, and 5 respectively. In the loss function in Eq.~\eqref{eq:loss}, we set $w_1, w_2$, and $w_3$ to 1. We use a hidden vector dimension of 50 for $g_{\theta_c}$ and $g_{\theta_g}$, and 150 for $g_{\theta_p}$. For $g_{\theta_g}$. Considering that our method is able to solve multiple equations with one model, the total model size for our method is much smaller than other baselines (see Appendix~\ref{sec:abl_p2inn}).

\section{Sensitivity Analyses}\label{a:sensitivity}

\begin{table}[h]
\centering
% \small
\caption{Experimental results of \PPINN{} by varying the dimension of $g_{{\theta}_p}$}
\setlength{\tabcolsep}{6pt}
\renewcommand{\arraystretch}{1.0}
\begin{tabular}{ccccc}
\specialrule{1pt}{2pt}{2pt}
\multirow{2}{*}{\textbf{Dim.}} & \multicolumn{2}{c}{\textbf{Convection}}  & \multicolumn{2}{c}{\textbf{Reaction}} \\ 
\cmidrule{2-5}

  & Abs. err. & Rel. err.  & Abs. err. & Rel. err. \\
\specialrule{1pt}{2pt}{2pt} 
\textbf{80}  & 0.0030 & 0.0060  & 0.0036 & 0.0054 \\ \midrule
\textbf{160} & 0.0029 & 0.0054  & 0.0016 & 0.0029\\ \midrule
\textbf{320} & 0.0023 & 0.0045  & 0.0061 & 0.0082\\
\specialrule{1pt}{2pt}{2pt}
\end{tabular}

\label{tbl:sensitivity}

\end{table}

We test \PPINN{} by varying the hidden vector dimension of $g_{{\theta}_p}$ in \{80, 160, 320\}. Convection and reaction equations with the coefficient range of 1 to 5 are used for testing and we summarize the result in Table~\ref{tbl:sensitivity}. As shown in Table~\ref{tbl:sensitivity}, \PPINN{} attain small errors in every hyperparameter setting compared to other baselines, which proves the robustness of our model.

\section{Additional Experiments}
\subsection{Large Range}

\begin{table}[h]
\centering
\caption{Experimental results on reaction equation with $\rho \in [1, 50]$}
\begin{tabular}{ccccc}
\specialrule{1pt}{2pt}{2pt}

 \textbf{Metric} & \textbf{PINN} & \textbf{PINN-R}  & \textbf{PINN-seq2eq} & \textbf{P$^2$INN} \\
\specialrule{1pt}{2pt}{2pt} 
\textbf{Abs. err.}  & 0.9053 & 0.4732 & 0.9190 & \underline{0.0486} \\
\textbf{Rel. err.} & 0.9383 & 0.5211 & 0.9582 & \underline{0.1322}\\ 
\specialrule{1pt}{2pt}{2pt}
\end{tabular}

\label{tbl:large_range}
\end{table}

We conduct additional experiments on the reaction equation with an initial condition of Gaussian distribution$(N(\pi,(\pi/2)^2)$). In these experiments, we test on equations with $\rho \in [1,50]$, which is an extremely wide range, to compare how models work in highly challenging scenarios in terms of range. As summarized in Table~\ref{tbl:large_range}, \PPINN{} surpass others, showing that \PPINN{} even works properly in the extremely large coefficient ranges.

% \clearpage
\subsection{Comparison with Meta-learning Algorithms}

\begin{table}[ht!]
\centering
\setlength{\tabcolsep}{6pt}
\renewcommand{\arraystretch}{0.8}
\caption{Comparison of our model with meta-learning based PINNs} 
\begin{tabular}{ccccc}
\specialrule{1pt}{2pt}{2pt}

 \textbf{PDE type} & \textbf{Metric} & \textbf{MAML} & \textbf{Reptile}  &  \textbf{P$^2$INN} \\
\specialrule{1pt}{2pt}{2pt} 
\multirow{2}{*}{\textbf{Convection}} & \textbf{Abs. err.}  & 0.0579 & 0.0173 & \underline{0.0039}  \\ 

& \textbf{Rel. err.} & 0.1036 & 0.0347 & \underline{0.0079} \\ \midrule

\multirow{2}{*}{\textbf{Reaction}} & \textbf{Abs. err.}  & 0.0029 & 0.0033 & \underline{0.0015}  \\ 

& \textbf{Rel. err.} & 0.0057 & 0.0064 & \underline{0.0027} \\ 

\specialrule{1pt}{2pt}{2pt}
\end{tabular}

\label{tbl:meta}

\end{table}

We compare \PPINN{} with two other meta-learning methods (i.e., MAML, Reptile). We first train three models using convection and reaction equations with coefficient range of 1$\sim$5 and then fine-tune MAML and Reptile. As shown in Table ~\ref{tbl:meta}, our model shows best performance among three models in every case, even without fine-tuning steps.

\section{Fine-tuning \PPINN{}}\label{a:fine}

In general, our \PPINN{} outperform other baselines in most of the tested equations. We can fine-tune the pre-trained model to further increase the accuracy and in this section, we show the efficacy of the fine-tuning step with intuitive visualizations.

\subsection{Experiments with Gaussian Distribution as an Initial Condition}
\begin{figure}[h]
\subfigure[Exact solution]{\includegraphics[width=0.32\columnwidth]{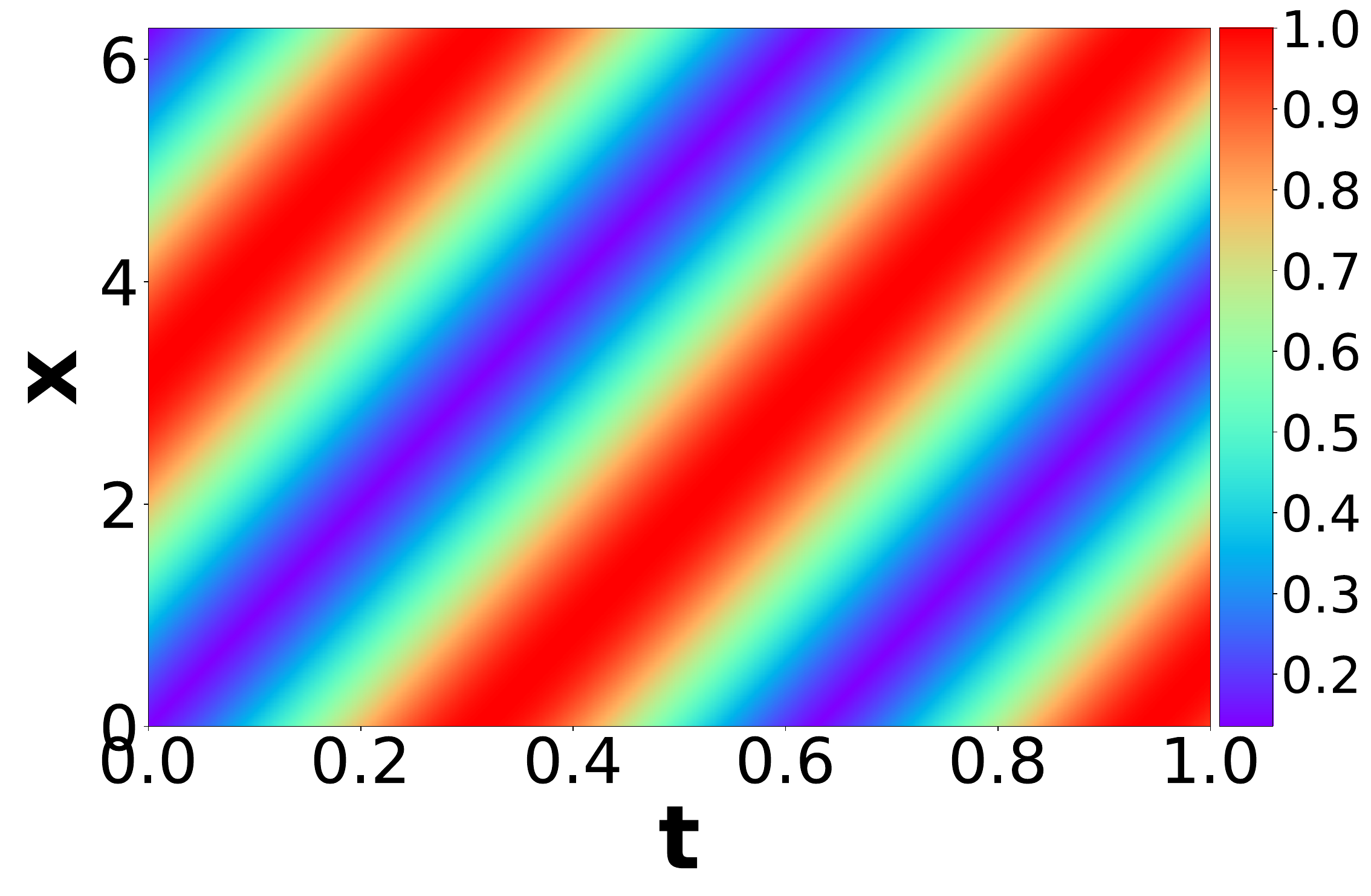}}\hfill
\subfigure[Result of PINN]{\includegraphics[width=0.32\columnwidth]{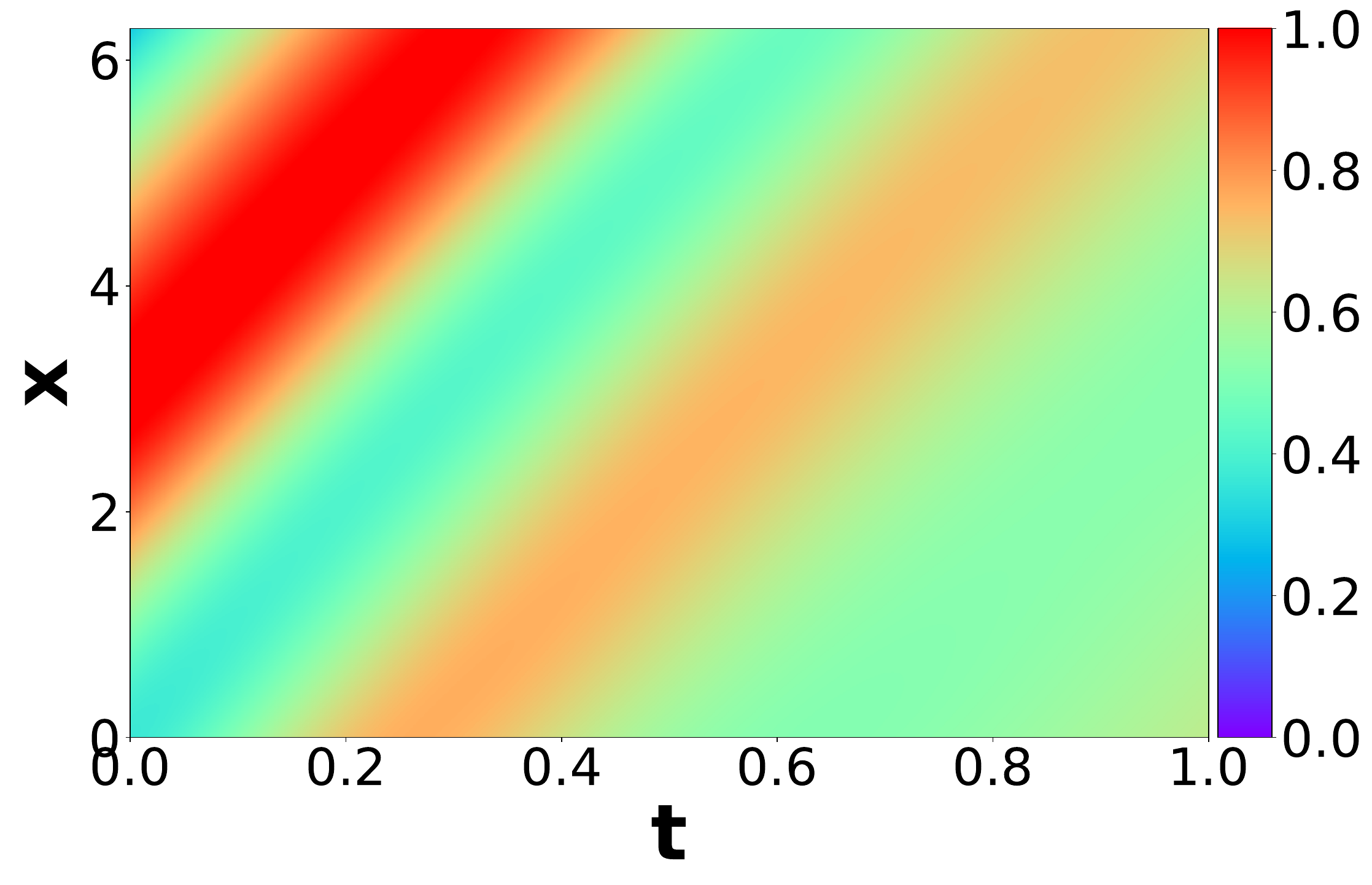}}\hfill
\subfigure[Result of \PPINN{}]{\includegraphics[width=0.32\columnwidth]{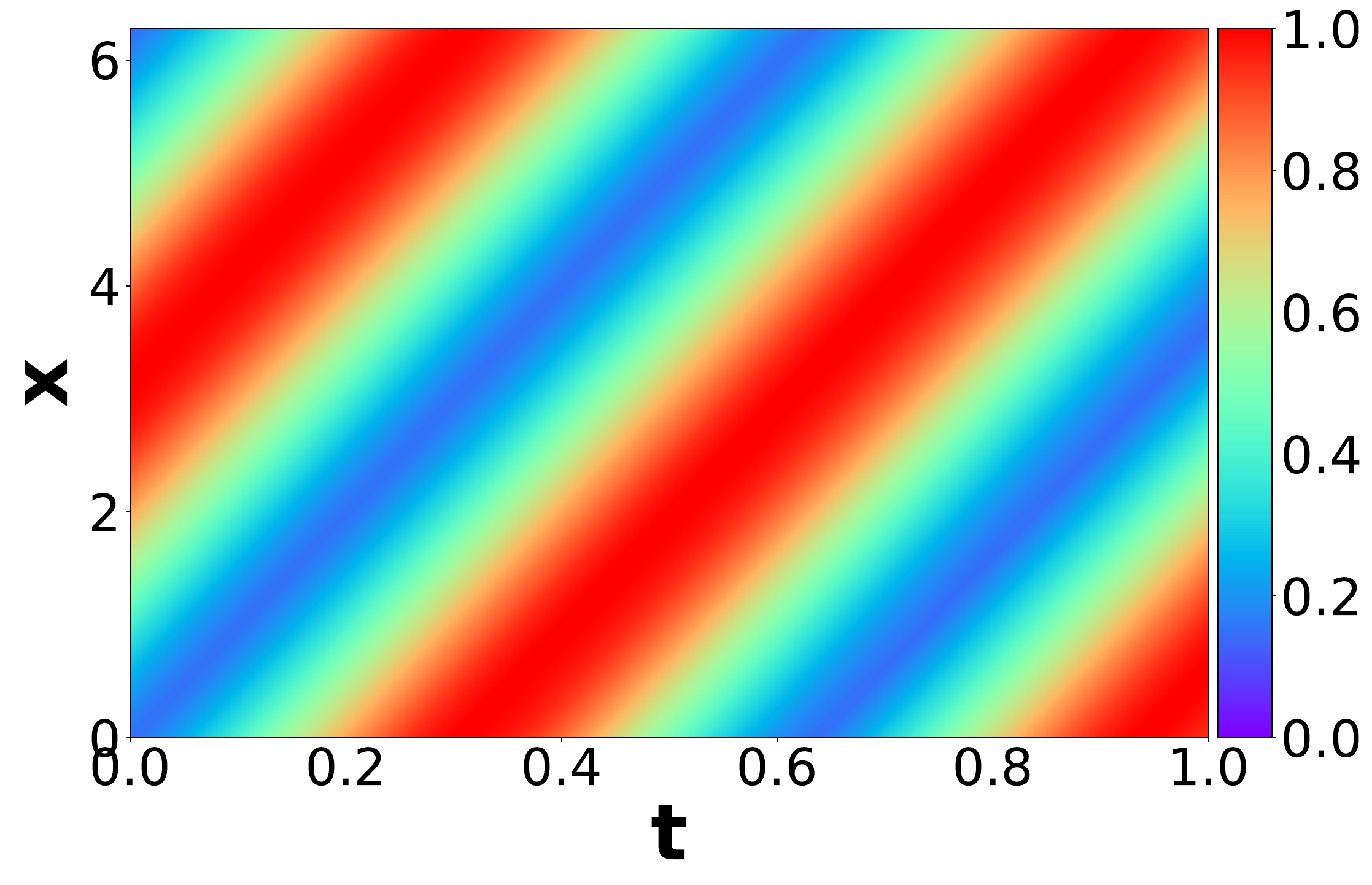}}\\

\subfigure[Exact solution]
{\includegraphics[width=0.32\columnwidth]{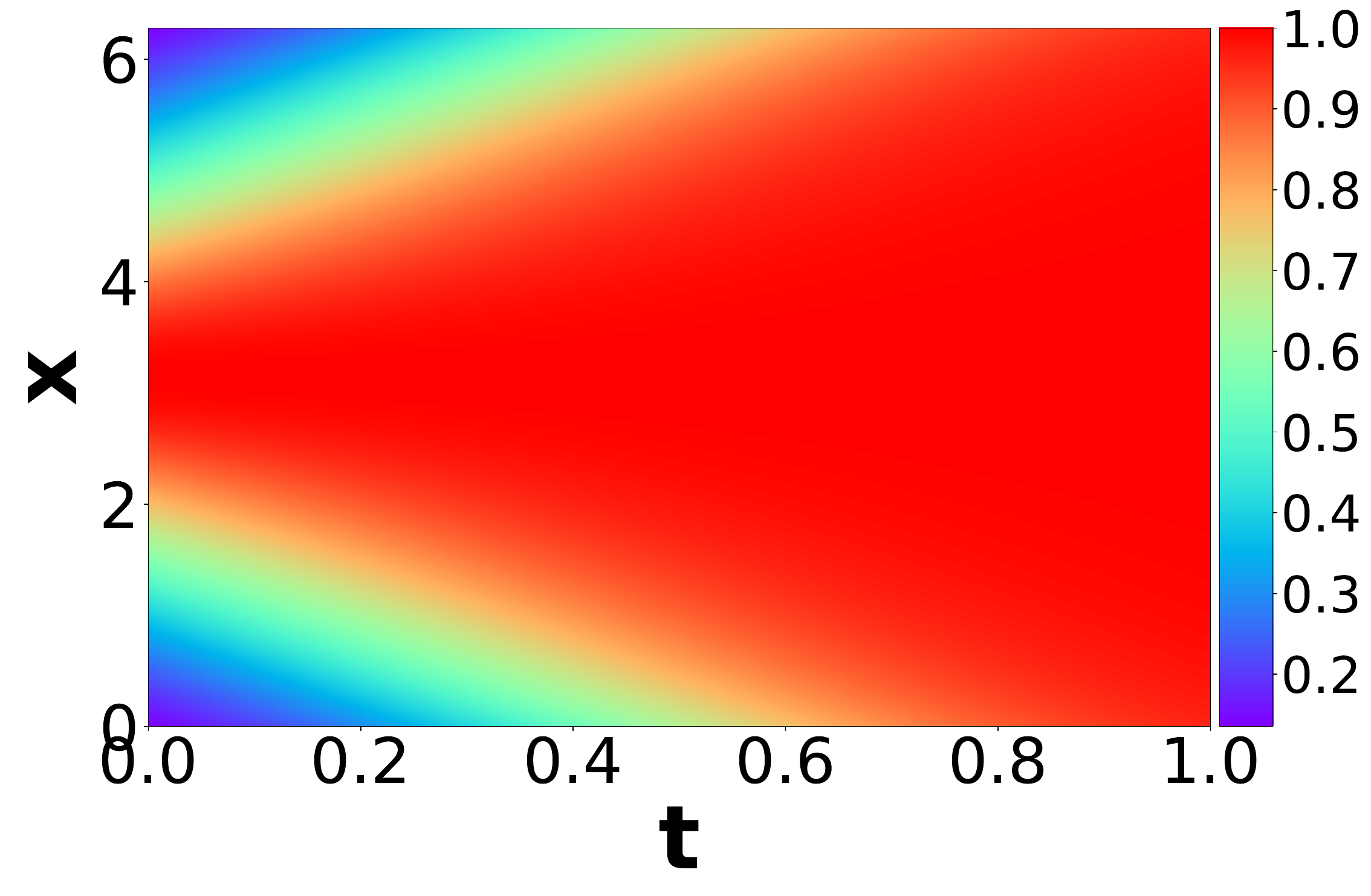}}\hfill
\subfigure[Result of PINN]{\includegraphics[width=0.32\columnwidth]{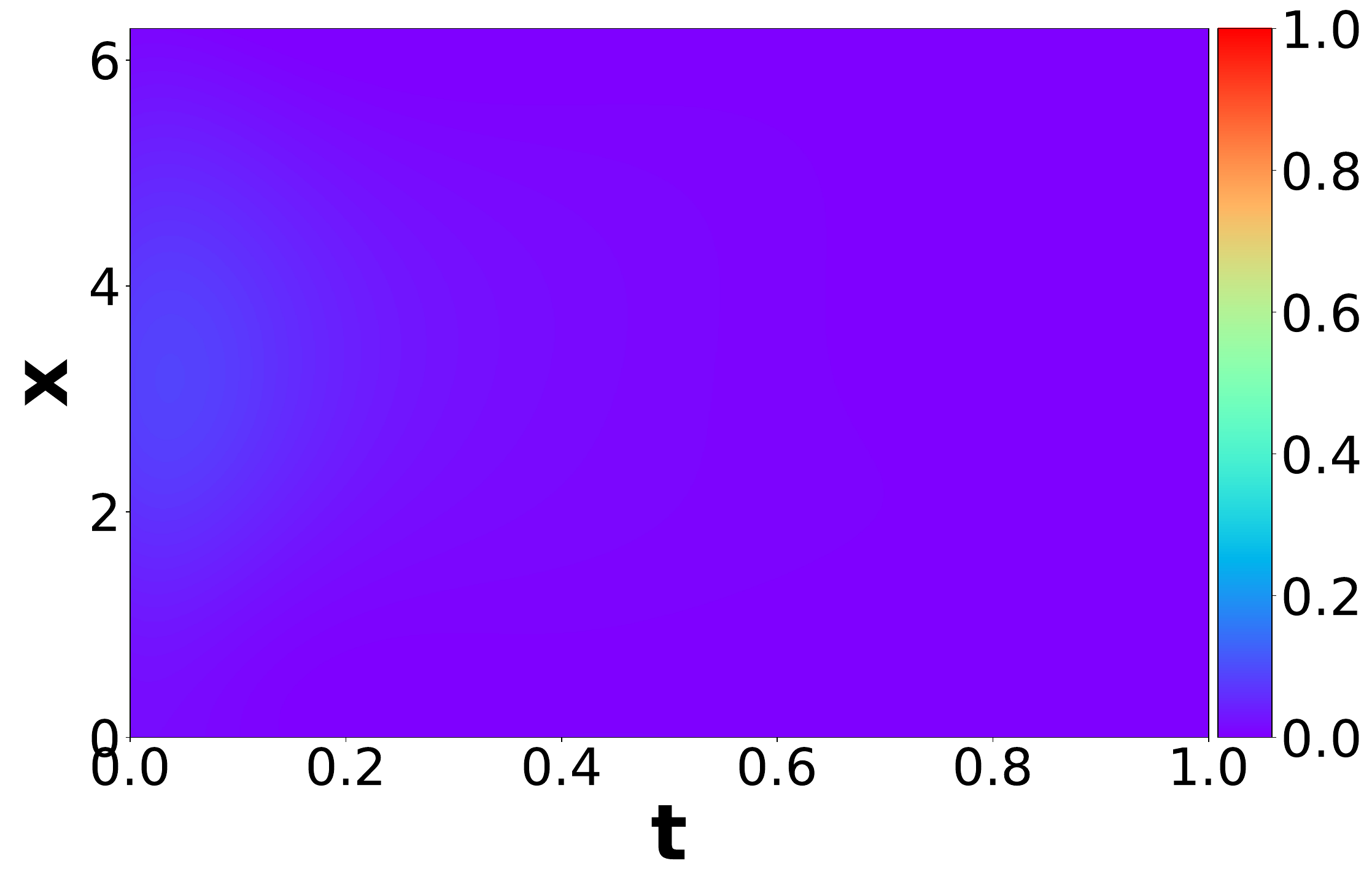}}\hfill
\subfigure[Result of \PPINN{}]{\includegraphics[width=0.32\columnwidth]{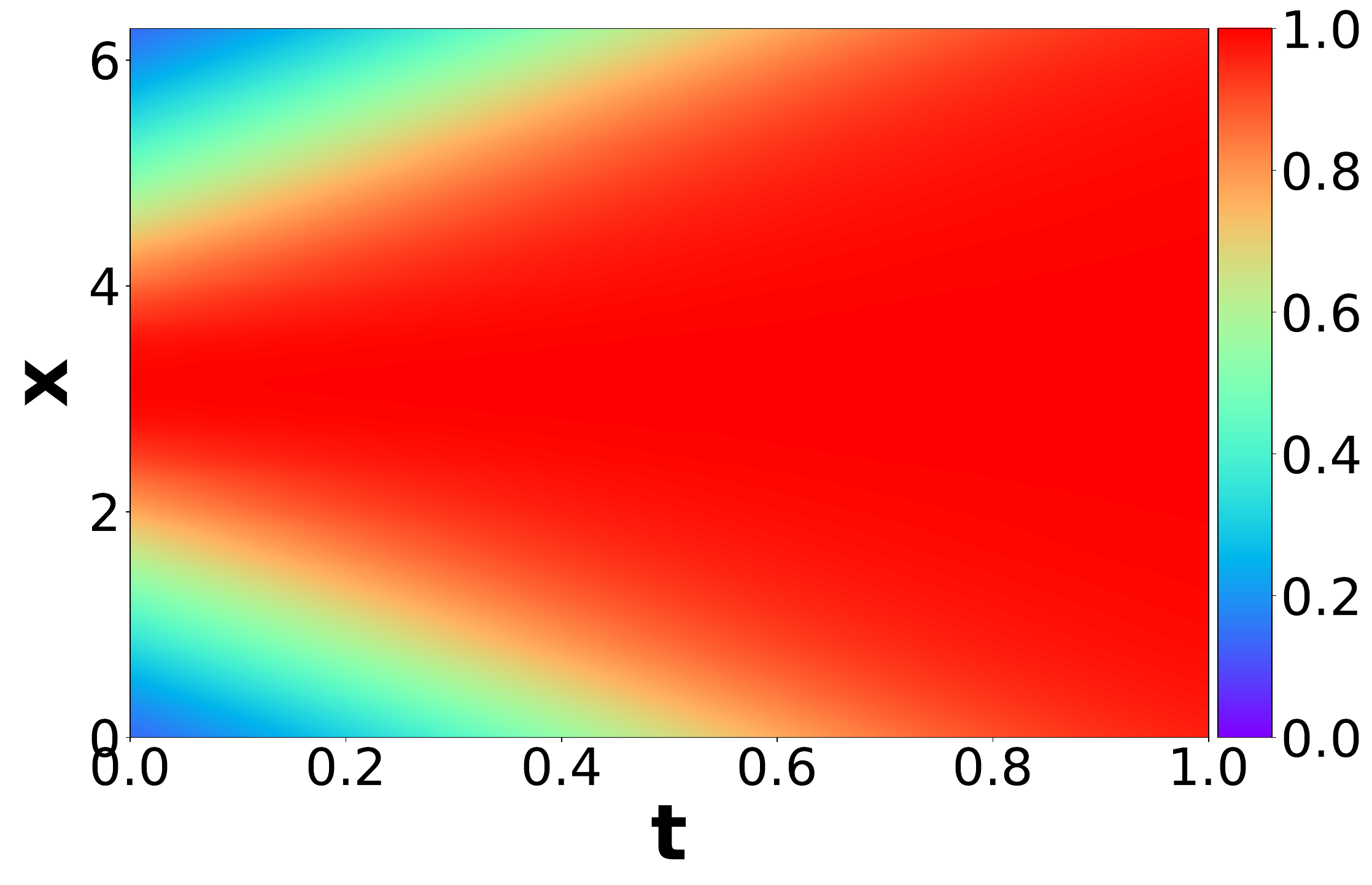}}\\

\caption{Experimental results of fine-tuning P$^2$INN. Convection equation of $\beta=30$ (Figure~\ref{fig:main_result_figure}. (a)-(c)). Reaction equation of $\rho=5$ (Figure~\ref{fig:main_result_figure}. (d)-(f)). Figures~\ref{fig:main_result_figure} (c) and (f) are the results after fine-tuning, and the results before fine-tuning can be checked in Figure~\ref{fig:main_coeff}. }\label{fig:main_result_figure}
\end{figure}

Experiments summarized in Table~\ref{tbl:result} use the initial condition of the Gaussian distribution $N(\pi, (\pi/2)^2)$. We fine-tune P$^2$INN from Table~\ref{tbl:result} on two PDEs: a convection equation with $\beta=10$, and a reaction equation with $\rho=5$. For the coefficient range used in pre-training, we select $\beta \in [1, 20]$ and $\rho \in [1, 10]$, respectively. We compare our fine-tuned model with vanilla PINN and results are summarized in Figure~\ref{fig:main_result_figure}.

\begin{figure}[ht!]
\subfigure[$\beta = 10$]{\includegraphics[width=0.32\columnwidth]{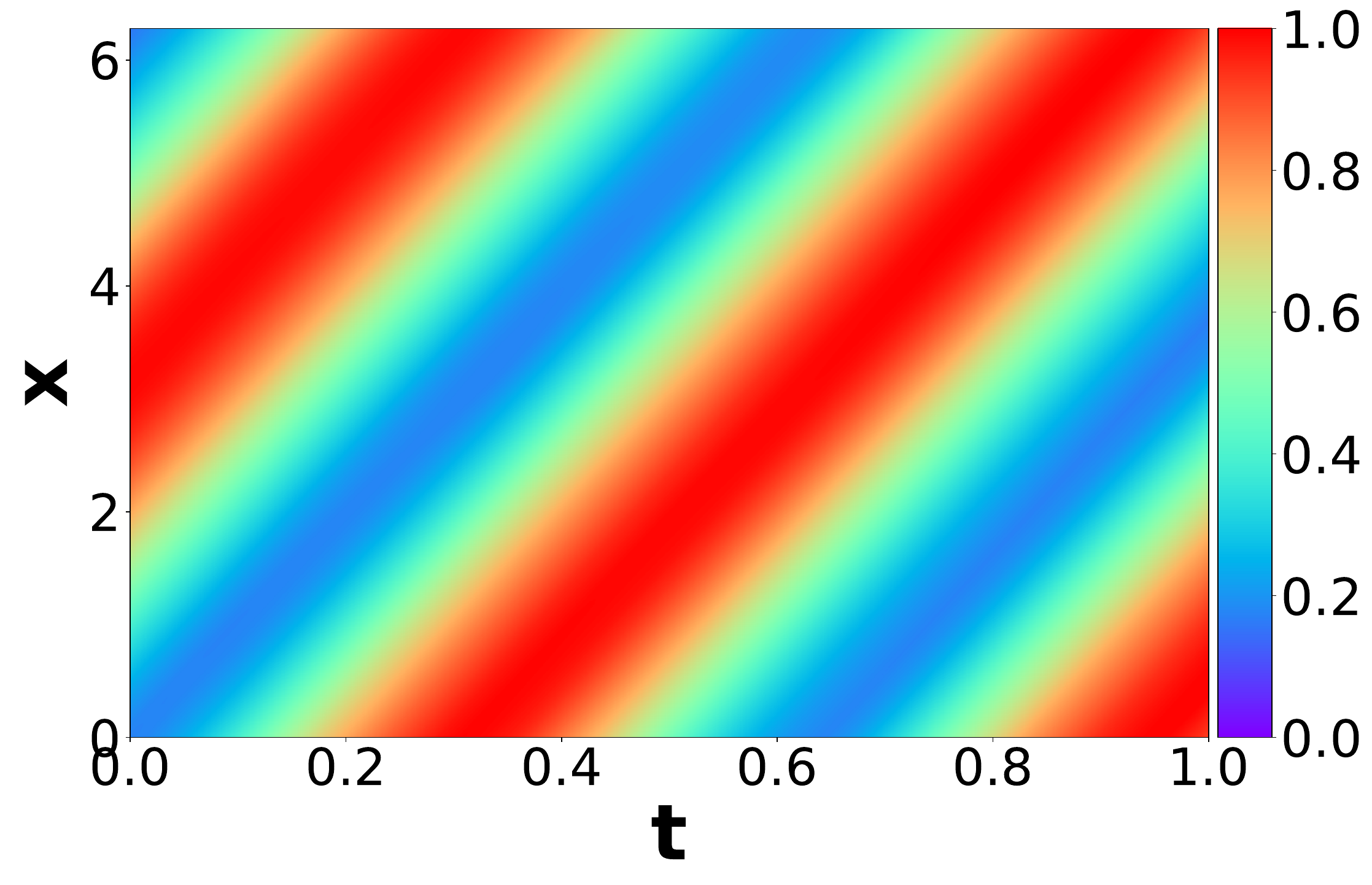}}\hfill
\subfigure[$\beta = 15$]{\includegraphics[width=0.32\columnwidth]{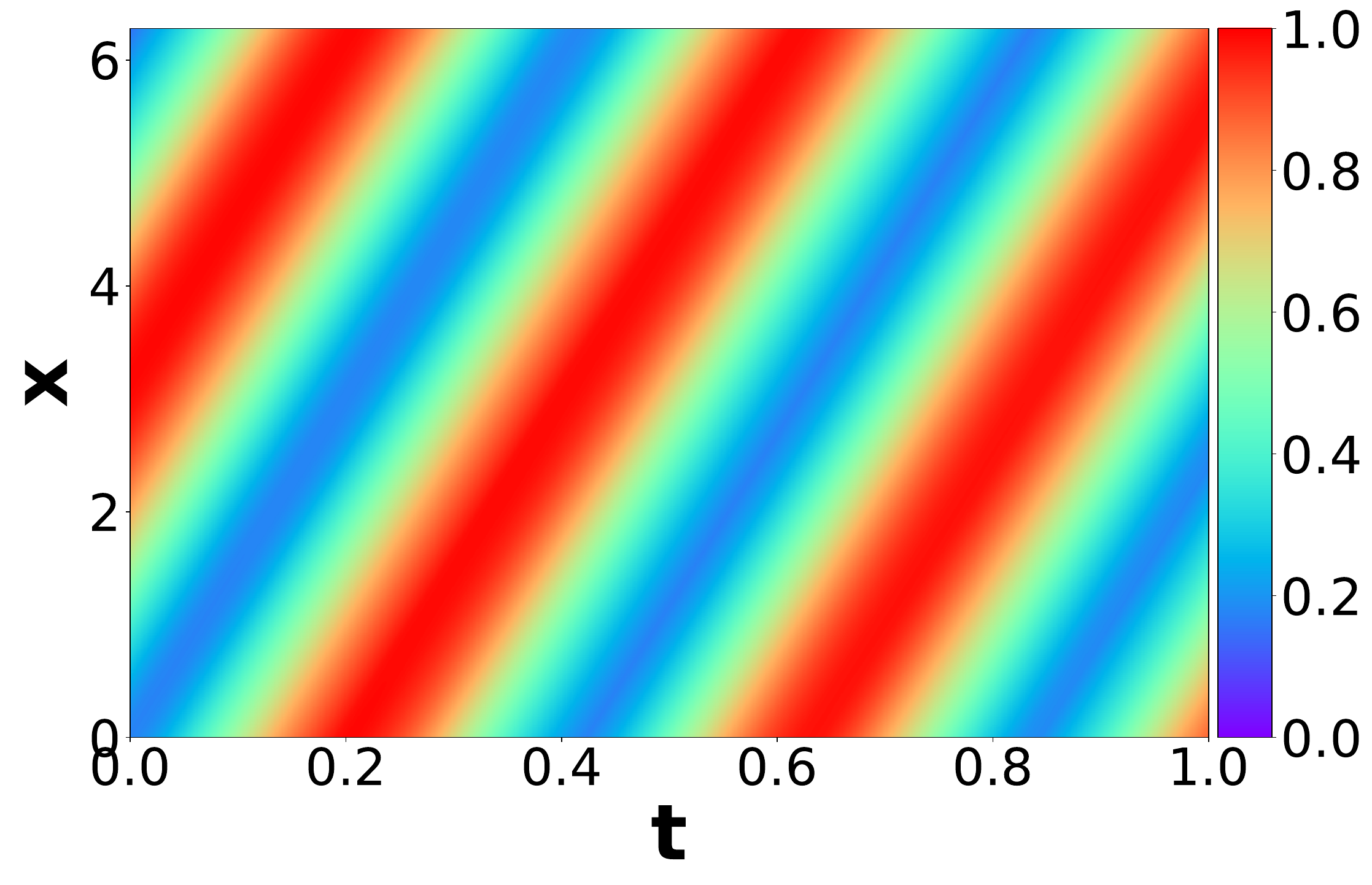}}\hfill
\subfigure[$\beta = 20$]{\includegraphics[width=0.32\columnwidth]{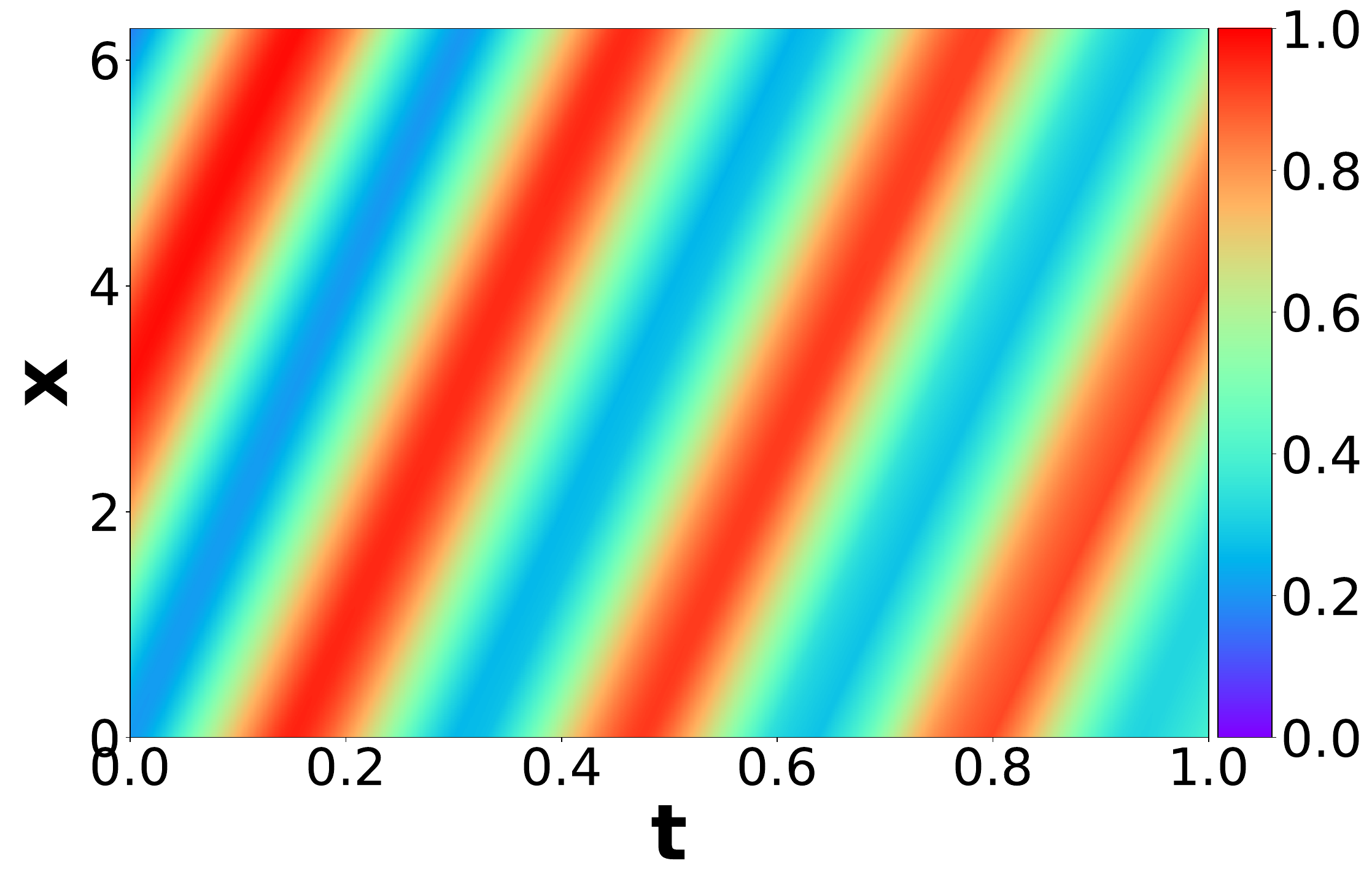}}\\

\subfigure[$\rho = 5$]
{\includegraphics[width=0.32\columnwidth]{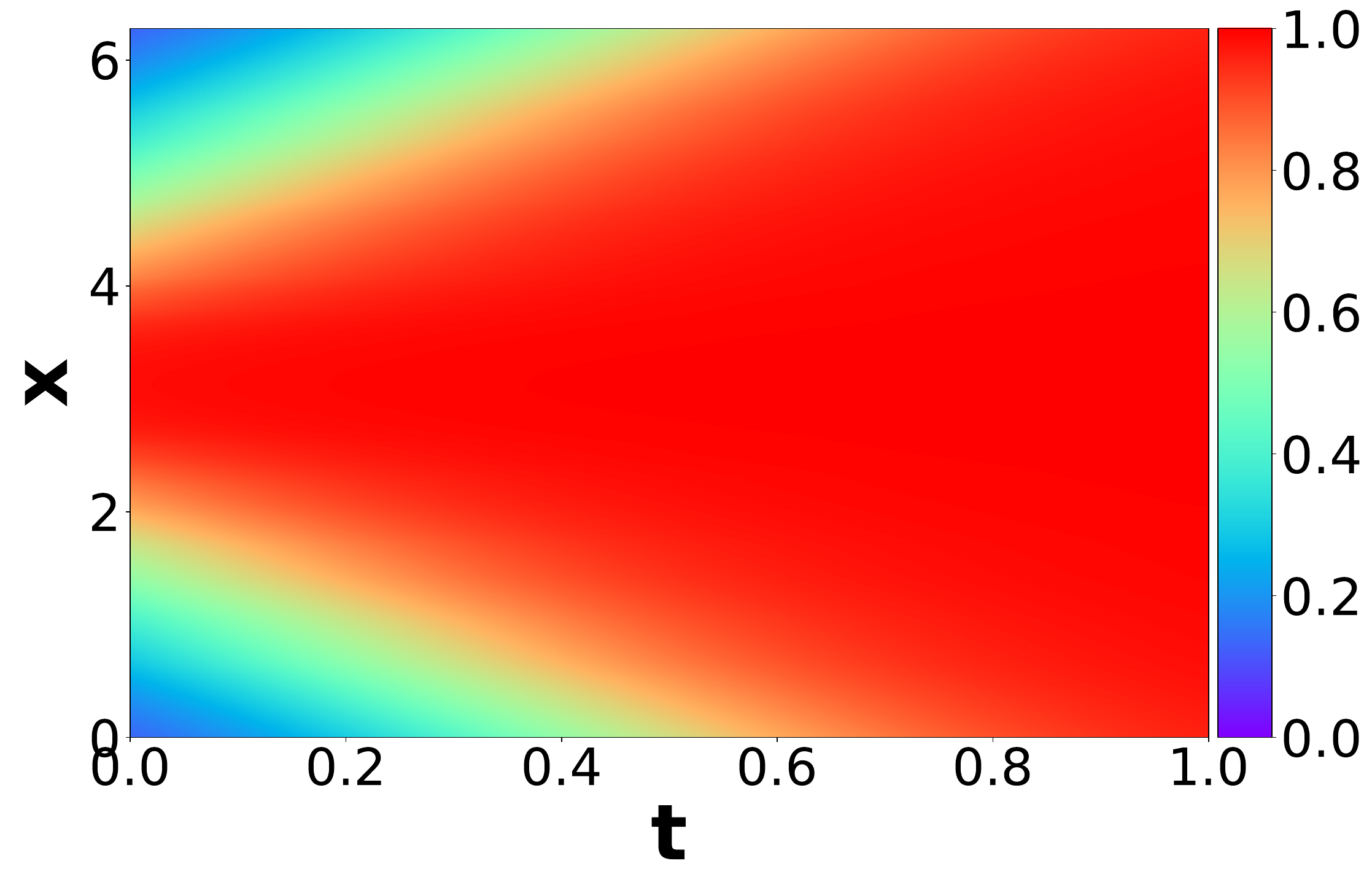}}\hfill
\subfigure[$\rho = 7$]{\includegraphics[width=0.32\columnwidth]{images/p2inn_gauss_pi_2_reaction_10_5_0.pdf}}\hfill
\subfigure[$\rho = 9$]{\includegraphics[width=0.32\columnwidth]{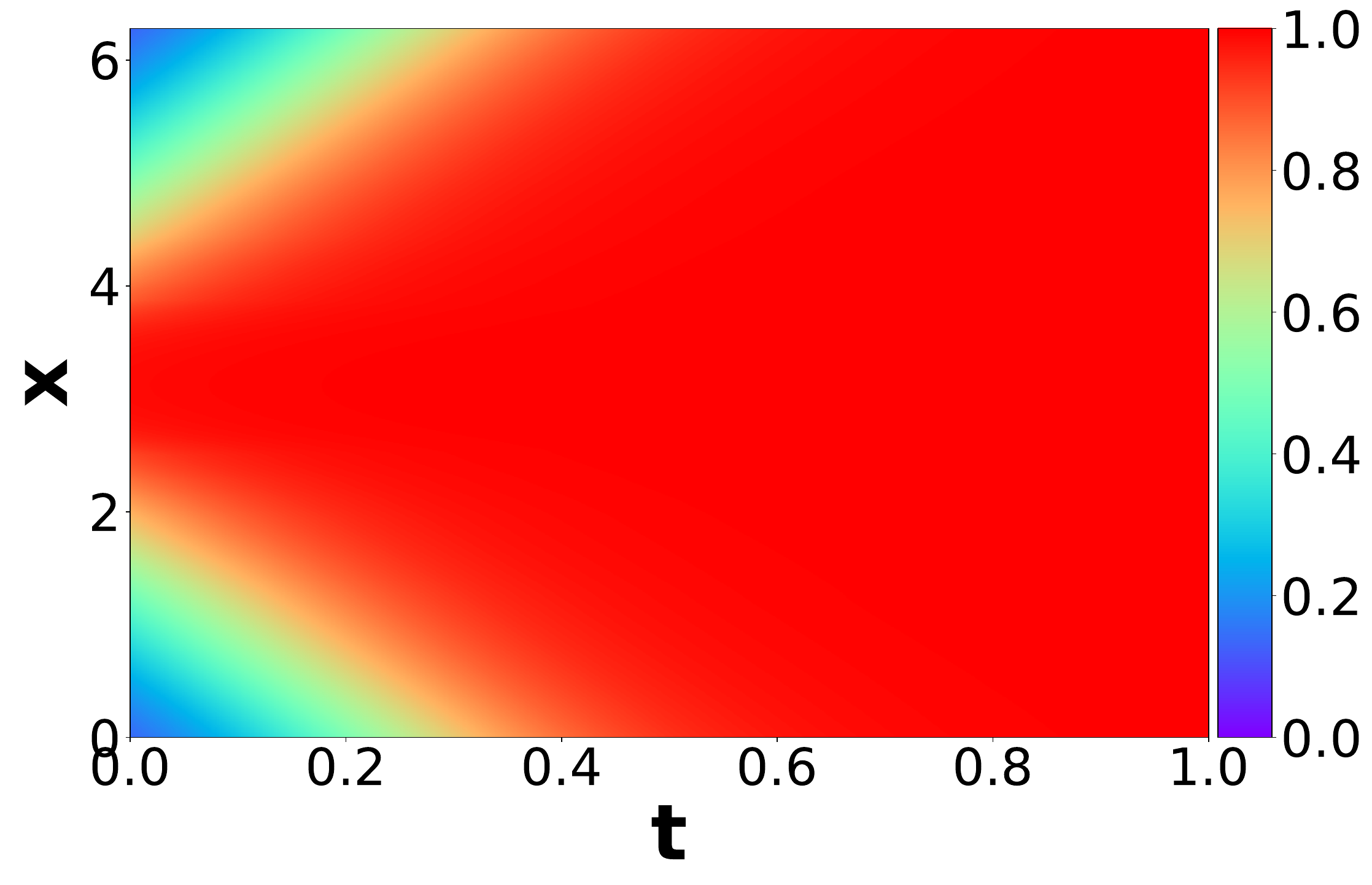}}\\

\caption{Results of P$^2$INN on convection equation and reaction equation without fine-tuning.}\label{fig:main_coeff}
\end{figure}

For the additional study, we show how the results of pre-trained \PPINN{} are affected by varying the PDE parameters. Figures~\ref{fig:main_coeff}(a-c)/(d-f)  are the results of convection/reaction equations. As shown in Figure~\ref{fig:main_coeff}, our \PPINN{} effectively learn the differences among the various coefficient settings.

\clearpage
\subsection{Failure Mode}

\begin{figure}[ht!]
\subfigure[Before fine-tuning]{\includegraphics[width=0.25\columnwidth]{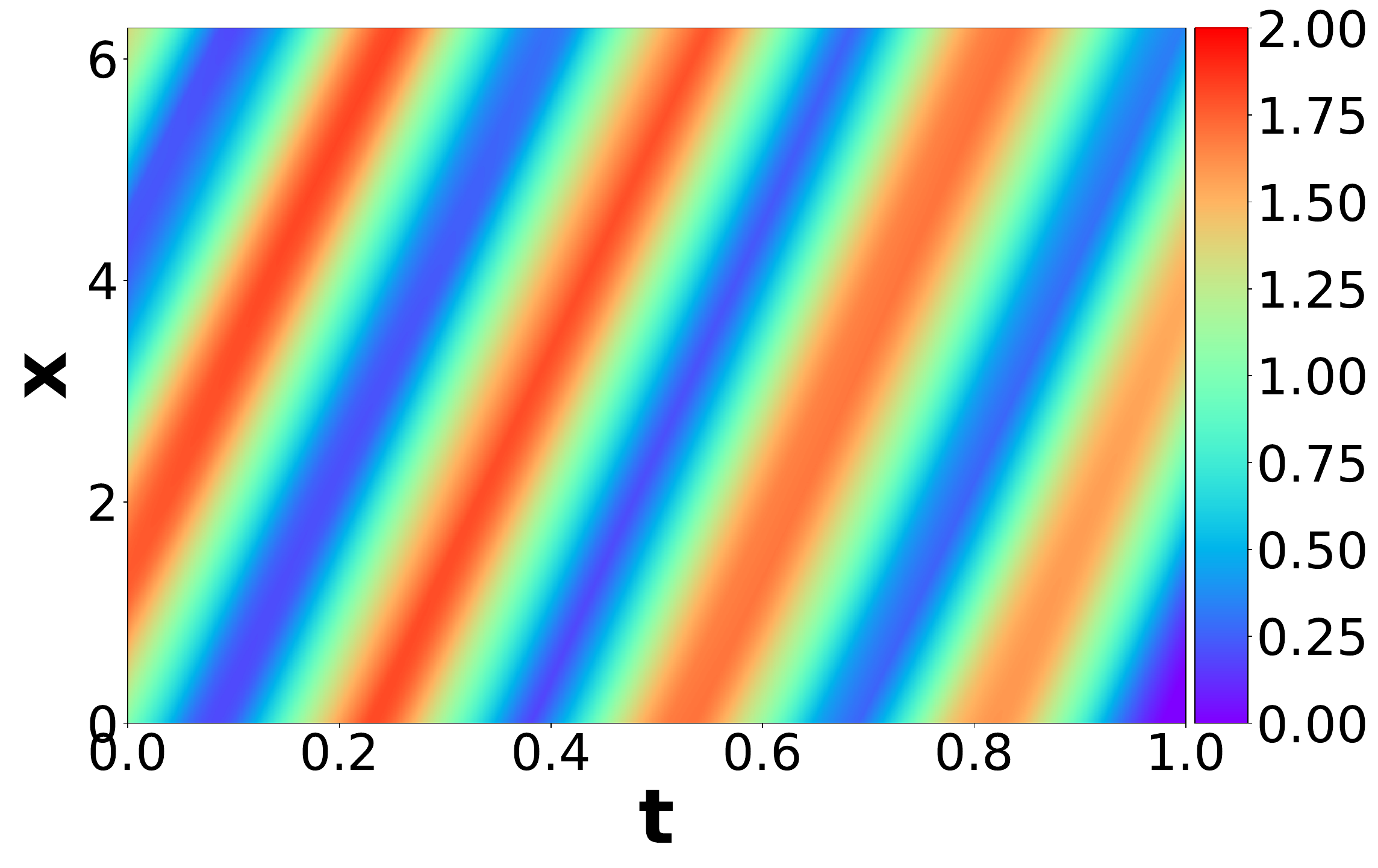}}\hfill
\subfigure[After fine-tuning]{\includegraphics[width=0.25\columnwidth]{images/p2inn_sin_1_conv30_20000.pdf}}\hfill
\subfigure[Before fine-tuning]{\includegraphics[width=0.25\columnwidth]{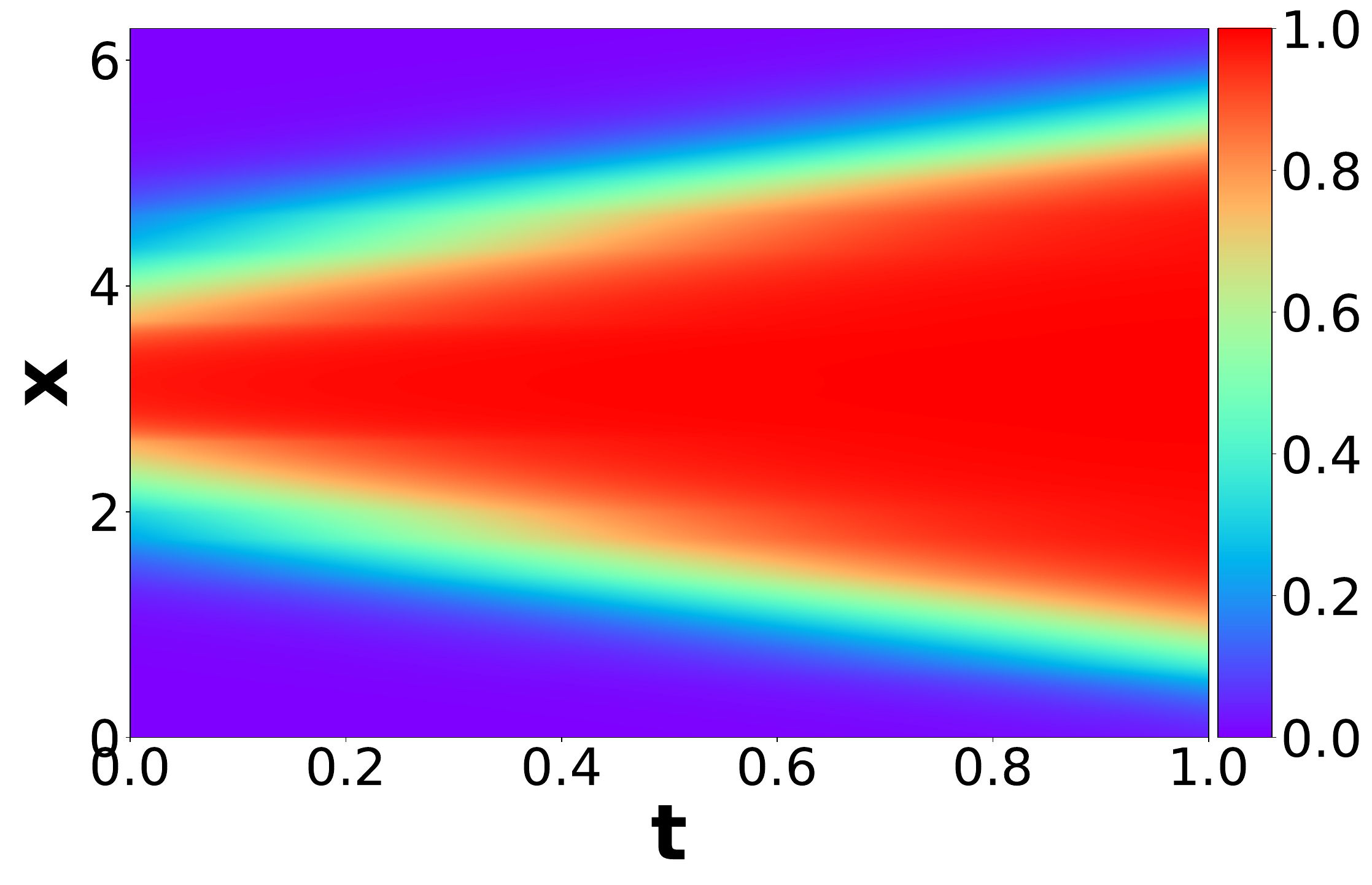}}\hfill
\subfigure[After fine-tuning]{\includegraphics[width=0.25\columnwidth]{images/p2inn_gauss_reac5_20000.pdf}}\\
\caption{Experimental results of P$^2$INN in Section~\ref{sec:failure_mode}. Figures (a) and (b) are the results of convection equation $\beta = 30$, and Figures (c) and (d) are reaction equation $\rho = 5$.} \label{fig:failure_finetune}
\end{figure}

\begin{table}[ht!]
\centering
% \small
\setlength{\tabcolsep}{6pt}
\renewcommand{\arraystretch}{1.0}
\caption{%
Results of \PPINN{} for the failure mode. We use a convection equation with $1+\sin(x)$ as an initial condition and a reaction equation with the Gaussian distribution $N(\pi, (\pi/4)^2)$.
}
\begin{tabular}{crrrr}
\specialrule{1pt}{2pt}{2pt}
\multirow{2}{*}{\textbf{Failure}} & \multicolumn{2}{c}{\textbf{PINN}} & \multicolumn{2}{c}{\textbf{P$^2$INN}}  \\
\cmidrule{2-5}
\textbf{mode} & Abs. err. & Rel. err. & Abs. err. & Rel. err. \\
\specialrule{1pt}{2pt}{2pt} 
$\beta = 30$  & 0.6132 & 0.5734 & 0.0910 & 0.0916 \\ \midrule
$\rho = 5$ & 0.5490 & 0.9844 & 0.0058 & 0.0173\\
\specialrule{1pt}{2pt}{2pt}
\end{tabular}
\label{tbl:failure_finetune}
\end{table}

Figure~\ref{fig:failure} is the result of P$^2$INN for the failure mode, and Figure~\ref{fig:failure_finetune} is a comparison between before and after fine-tuning on the results of P$^2$INN. Figures~\ref{fig:failure_finetune} (a) and (b) are the results on convection equation of $\beta=30$, and Figures~\ref{fig:failure_finetune} (c) and (d) are the results on reaction equation of $\rho=5$. As shown in Table~\ref{tbl:failure_finetune}, P$^2$INN significantly improves the performance compared to PINN.

\subsection{Modulation for P$^2$INNs}\label{a:full_mod}

Here, we provide the full results from Table~\ref{tbl:modulation_results} for six-types of CDR equations employing our SVD modulation.

\begin{table}[ht!]
\centering
\setlength{\tabcolsep}{3.8pt}
\small
\caption{Experimental results of modulations with an initial condition of
    a Gaussian distribution $N(\pi, ({\pi}/2)^2)$}
\begin{tabular}{llrrrrrrr}
\specialrule{1pt}{2pt}{2pt}
\multirow{2.5}{*}{\textbf{PDE type}} & \multirow{2.5}{*}{\textbf{Metric}} & \multirow{2.5}{*}{\textbf{PINN}} & \multirow{2.5}{*}{\textbf{PINN-R}} & \multirow{2.5}{*}{\textbf{PINN-seq2seq}} & \multirow{2.5}{*}{\textbf{P$^2$INN}} & \multicolumn{3}{c}{\textbf{Modulation}} \\ \cmidrule{7-9}
& & & & & & \textbf{All} & \textbf{Shift} & \textbf{SVD} \\
\specialrule{1pt}{2pt}{2pt}
\multirow{2}{*}{\textbf{Convection}} & Abs. err. & 0.0183 & 0.0222 & 0.1281 & 0.0174 & 0.1959 & 0.0139  & \textbf{0.0138} \\
                                   & Rel. err. & 0.0327 & 0.1665 & 0.1987 & 0.0316 & 0.3319 & 0.0248  & \textbf{0.0246} \\
                                   \midrule

\multirow{2}{*}{\textbf{Diffusion}} & Abs. err. & 0.1335 & 0.1665 & 0.1987 & 0.0764 & 0.0726  & 0.0801  & \textbf{0.0689} \\
                                   & Rel. err. & 0.2733 & 0.3462 & 0.4050 & 0.1694 & \textbf{0.1392} & 0.1785 & 0.1518 \\
                                   \midrule
                                   
\multirow{2}{*}{\textbf{Reaction}} & Abs. err. & 0.3341 & 0.3336 & 0.4714 & 0.0126 & 0.0713 & 0.0095 & \textbf{0.0089} \\
                                   & Rel. err. & 0.3907 & 0.3907 & 0.5907 & 0.0229 & 0.1211 & 0.0198 & \textbf{0.0184} \\ \midrule

\multirow{2}{*}{\textbf{Conv.-Diff.}} & Abs. err. & 0.0610 & 0.0654 & 0.0979 & 0.0443 & 0.0555 &  0.0452 & \textbf{0.0422} \\
                                   & Rel. err. & 0.1175 & 0.1289 & 0.1950 & 0.0897 & 0.1074 & 0.0931 & \textbf{0.0832} \\ \midrule

\multirow{2}{*}{\textbf{Reac.-Diff.}} & Abs. err. & 0.1900 & 0.1876 & 0.4201 & 0.0586 & 0.0581 & 0.0603 & \textbf{0.0548} \\
                                   & Rel. err. & 0.2702 & 0.2777 & 0.5346 & 0.1015 & \textbf{0.0886} & 0.1043 & 0.0947 \\ \midrule                                

\multirow{2}{*}{\textbf{Conv.-Diff.-Reac.}} & Abs. err. & 0.1663 & 0.0865 & 0.4943 & 0.0315 & 0.0463 & 0.0321  & \textbf{0.0303} \\
                                   & Rel. err. & 0.2057 & 0.1415 & 0.6104 & 0.0508 & 0.0690 & 0.0521 & \textbf{0.0486} \\
\specialrule{1pt}{2pt}{2pt}
\end{tabular}\label{tbl:modulation_results_appendix}
\end{table}

\section{Experimental Results on 2D Helmholtz Equation}\label{a:helmholtz}

\begin{figure}[ht!]
    \subfigure[Exact ($a=2.5$)]
    {\includegraphics[width=0.19\columnwidth]{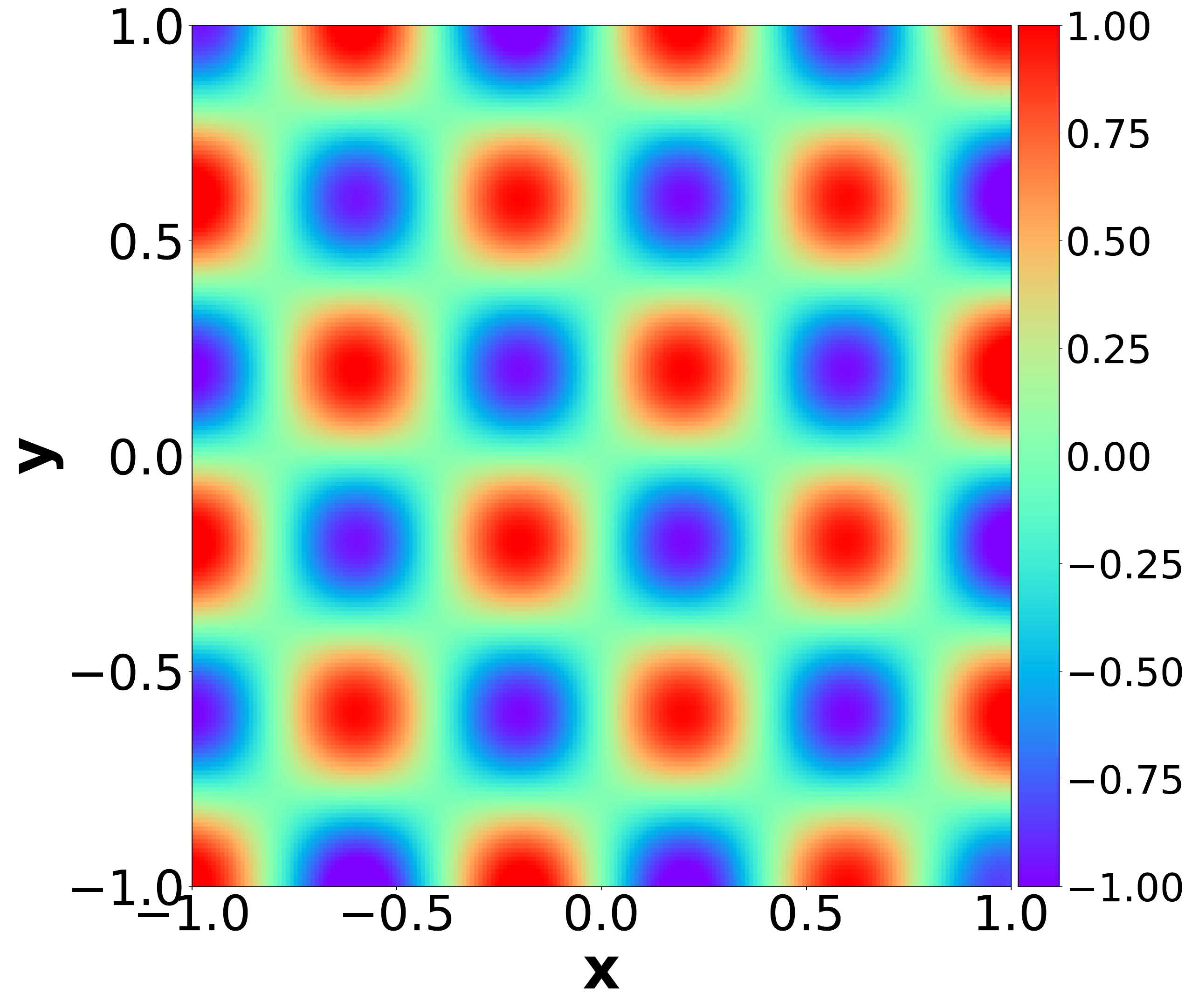}}\hfill
    \subfigure[Exact ($a=2.65$)]
    {\includegraphics[width=0.19\columnwidth]{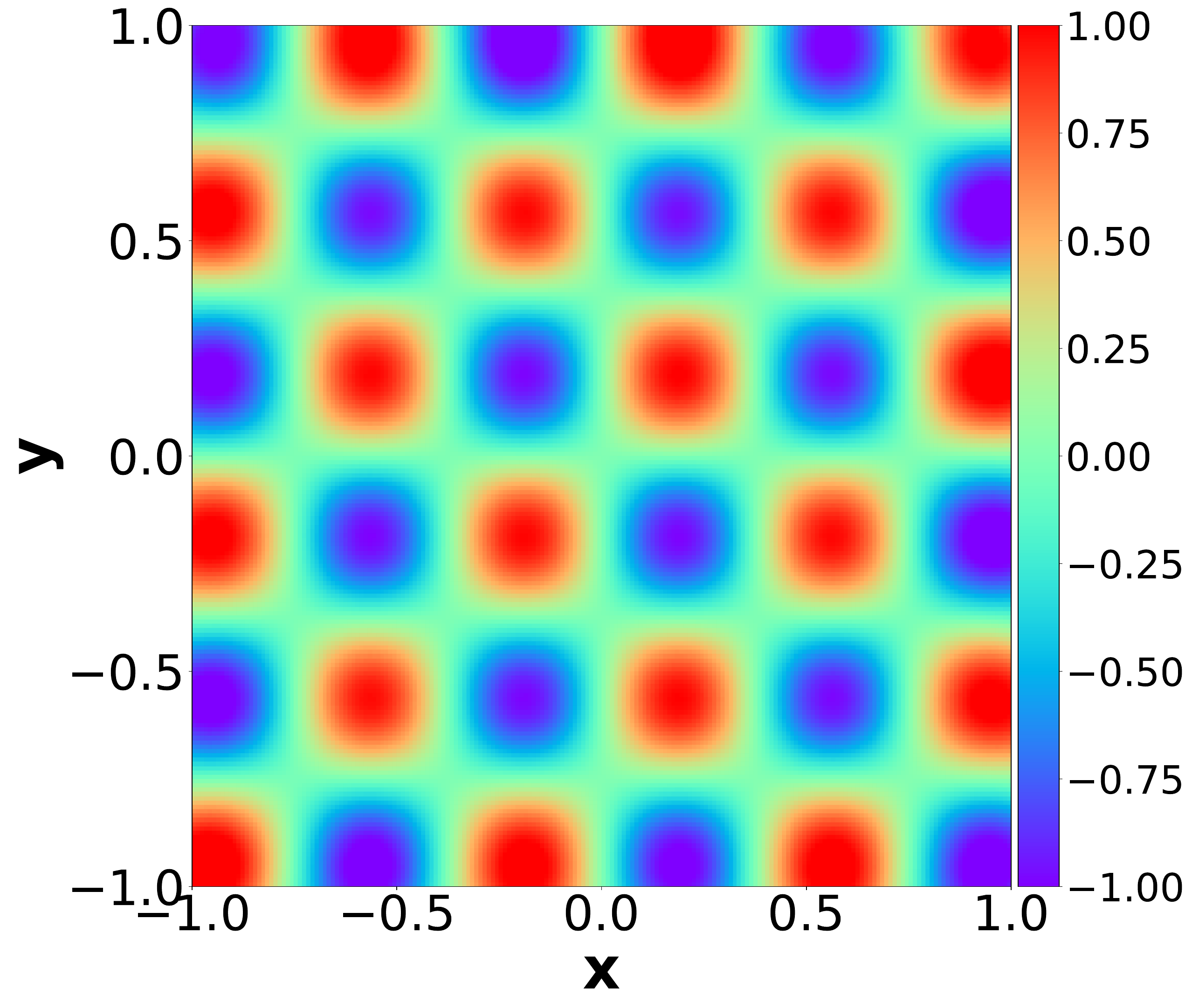}}\hfill
    \subfigure[Exact ($a=2.75$)]
    {\includegraphics[width=0.19\columnwidth]{images/helm_p2inn_2.75.pdf}}\hfill
    \subfigure[Exact ($a=2.85$)]
    {\includegraphics[width=0.19\columnwidth]{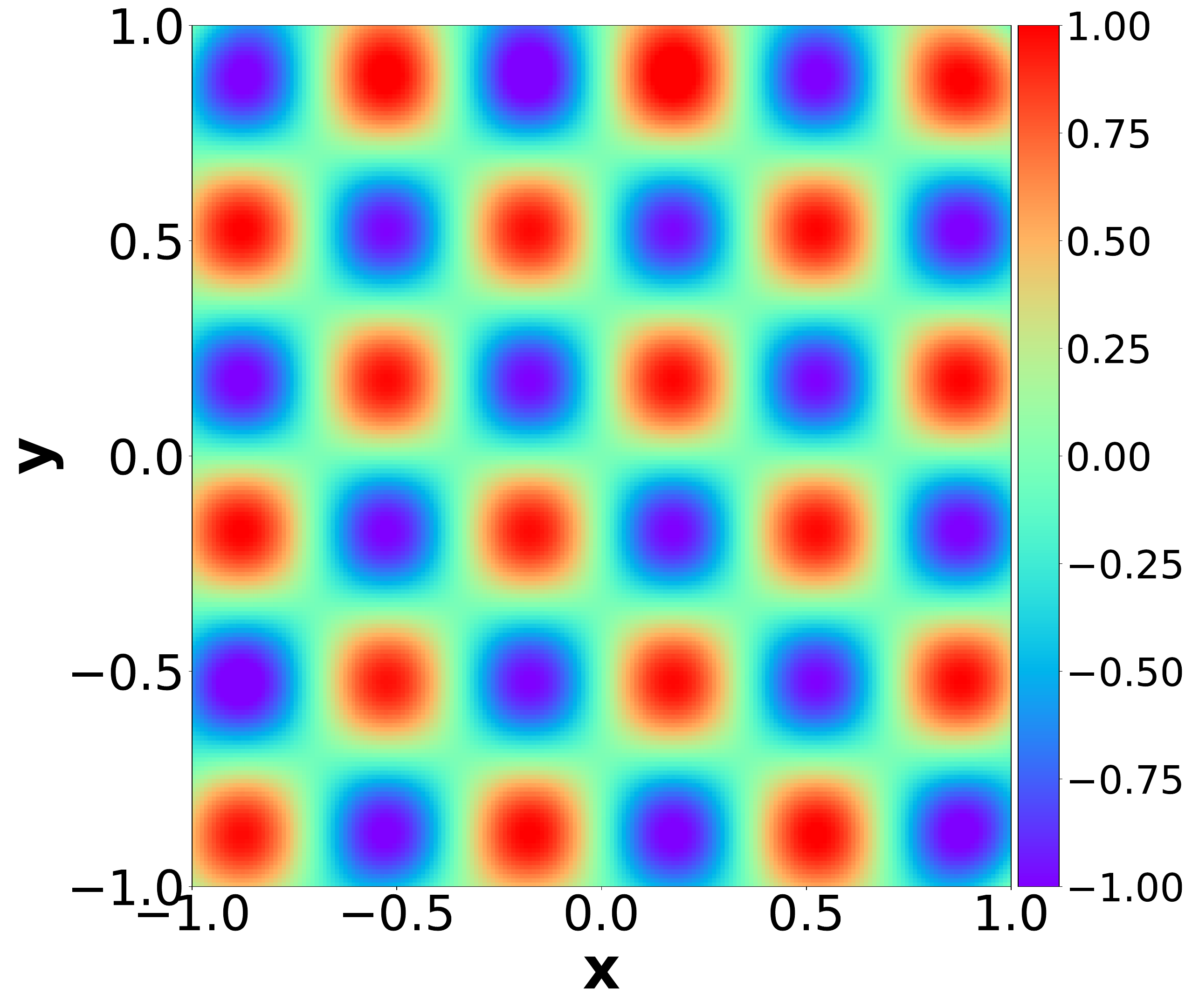}}\hfill
    \subfigure[Exact ($a=3.0$)]
    {\includegraphics[width=0.19\columnwidth]{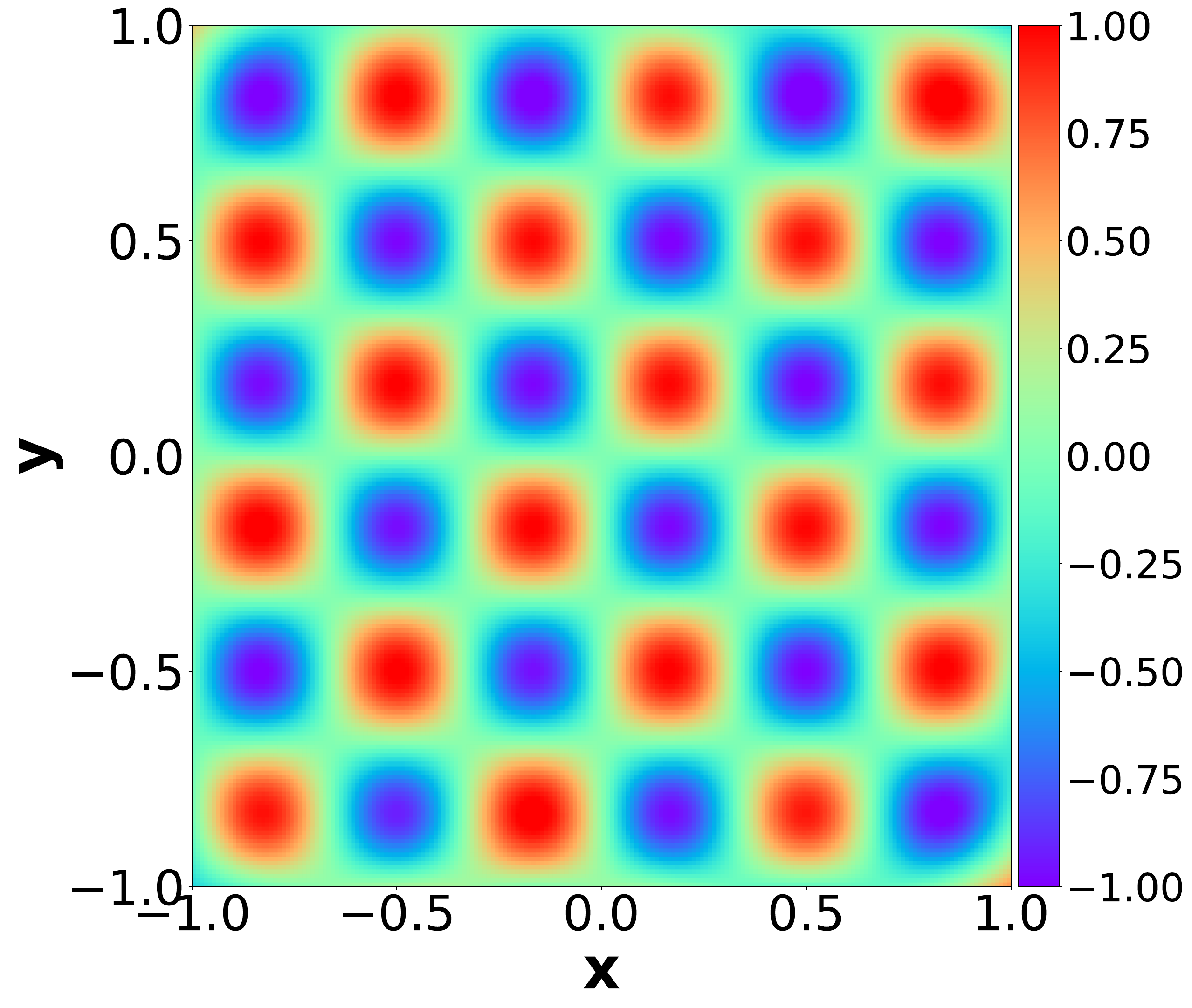}}\\

    \subfigure[PINN ($a=2.5$)]
    {\includegraphics[width=0.19\columnwidth]{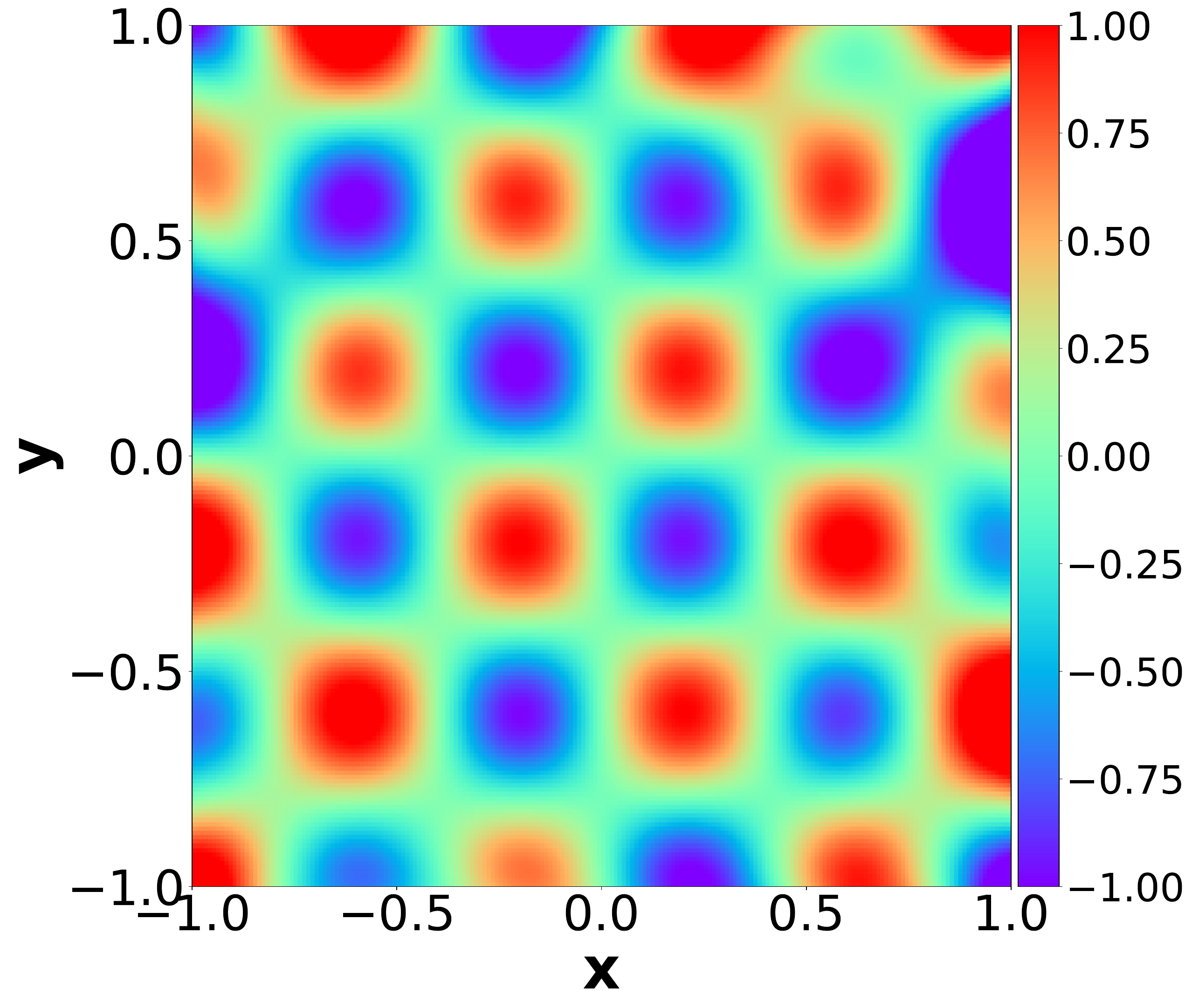}}\hfill
    \subfigure[PINN ($a=2.65$)]
    {\includegraphics[width=0.19\columnwidth]{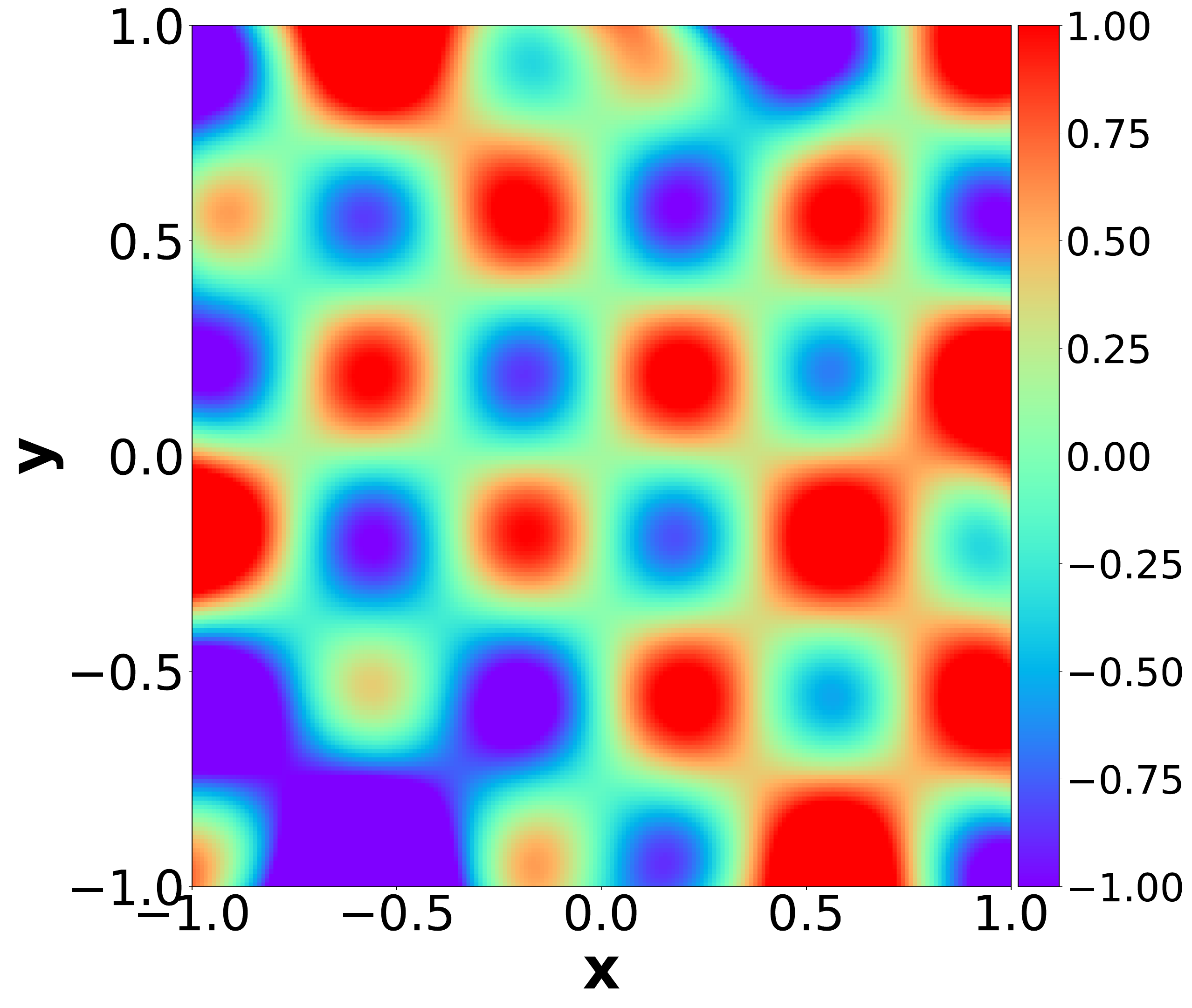}}\hfill
    \subfigure[PINN ($a=2.75$)]
    {\includegraphics[width=0.19\columnwidth]{images/pinn_helm_2.75_2000.pdf}}\hfill
    \subfigure[PINN ($a=2.85$)]
    {\includegraphics[width=0.19\columnwidth]{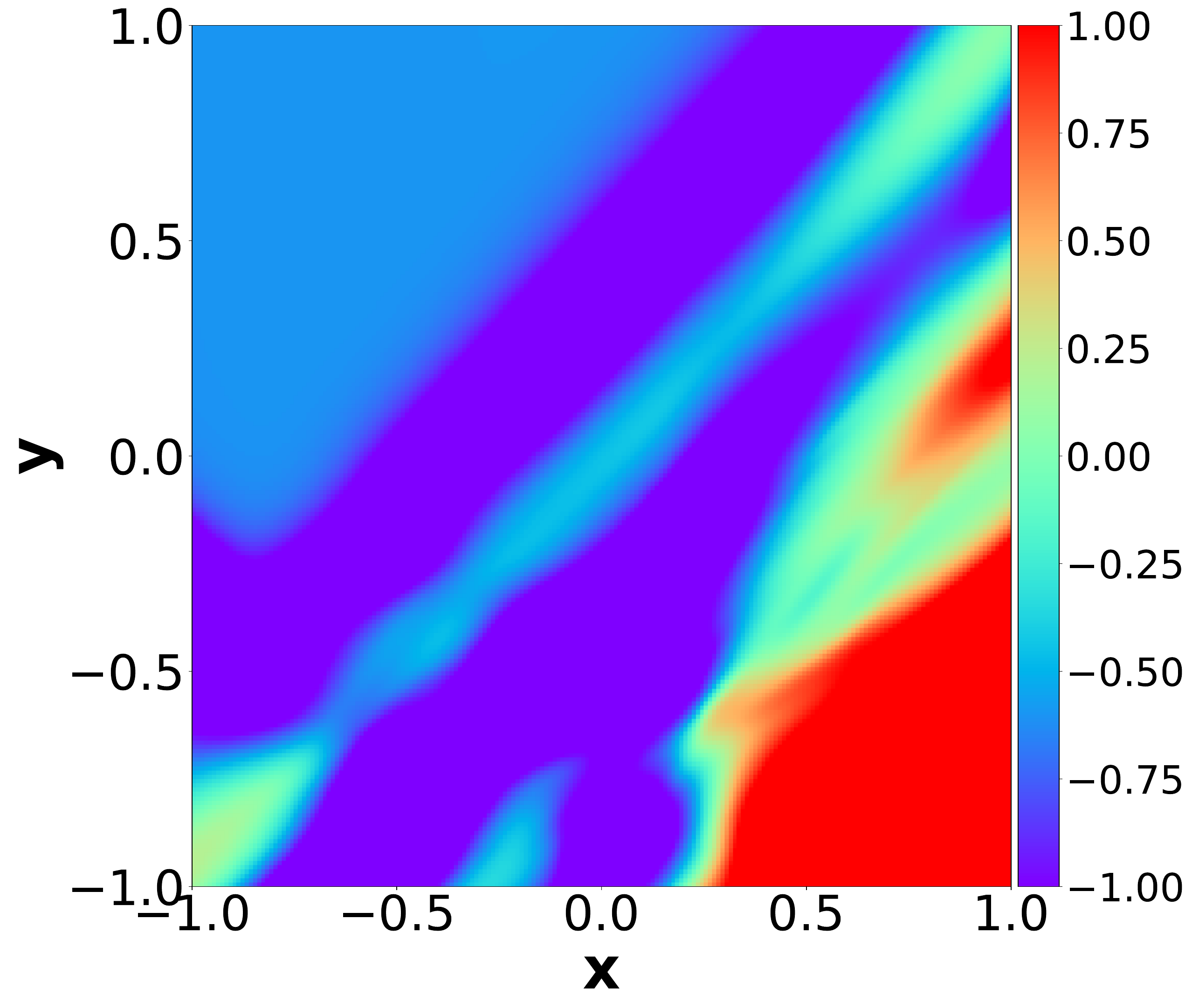}}\hfill
    \subfigure[PINN ($a=3.0$)]
    {\includegraphics[width=0.19\columnwidth]{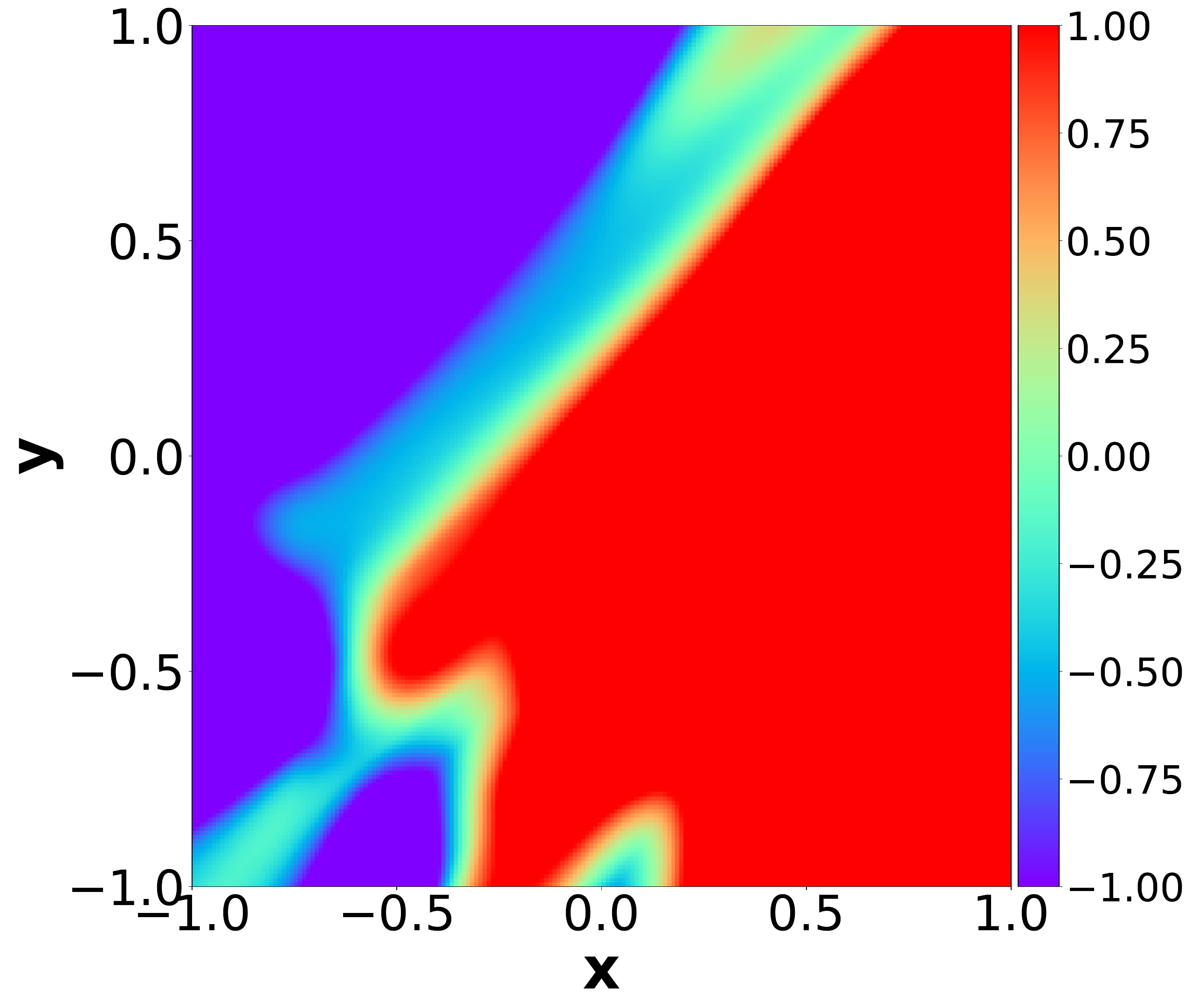}}\\

    \subfigure[PINN-R ($a=2.5$)]
    {\includegraphics[width=0.19\columnwidth]{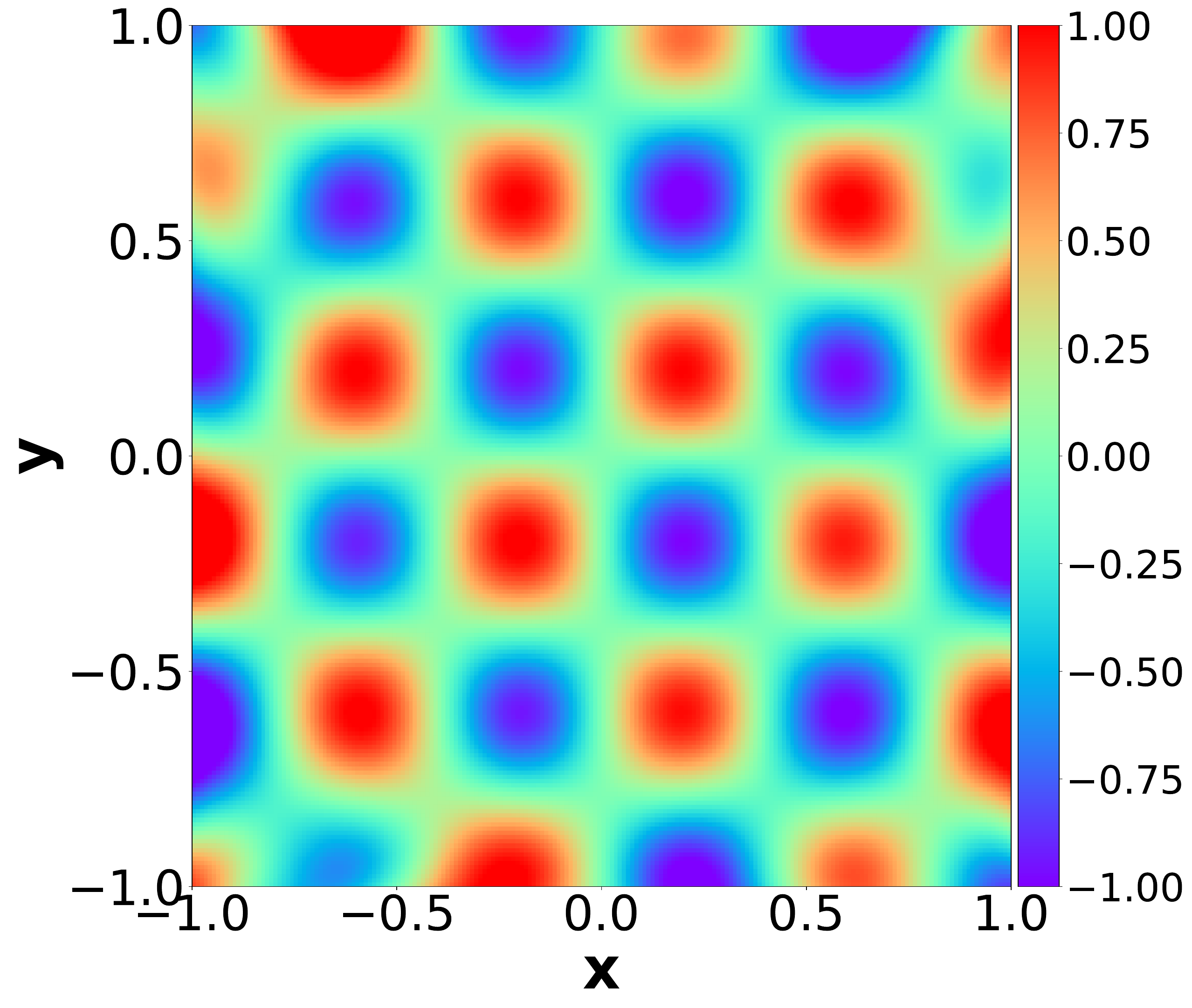}}\hfill
    \subfigure[PINN-R ($a=2.65$)]
    {\includegraphics[width=0.19\columnwidth]{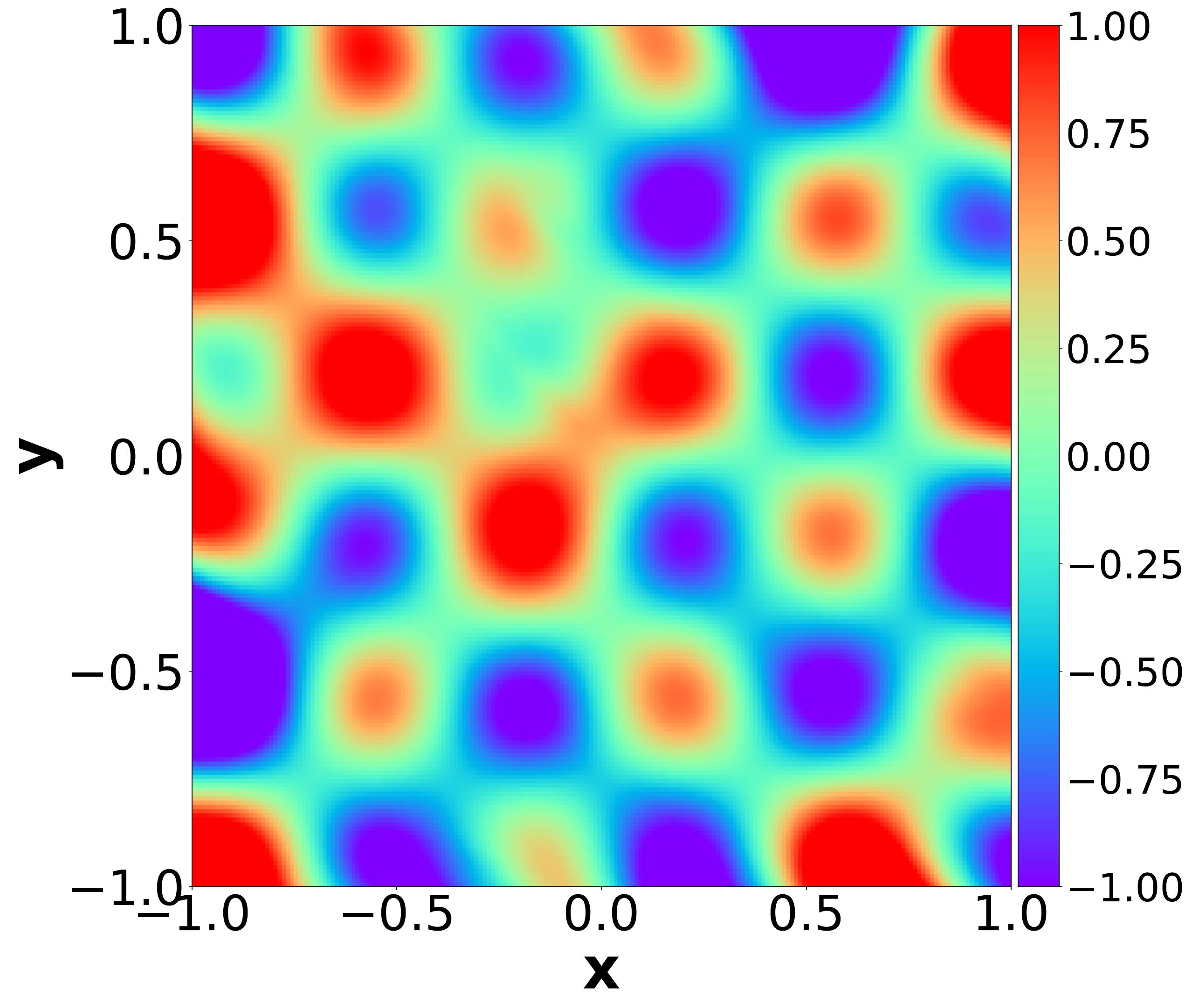}}\hfill
    \subfigure[PINN-R ($a=2.75$)]
    {\includegraphics[width=0.19\columnwidth]{images/pinn_R_helm_2.75_2000.pdf}}\hfill
    \subfigure[PINN-R ($a=2.85$)]
    {\includegraphics[width=0.19\columnwidth]{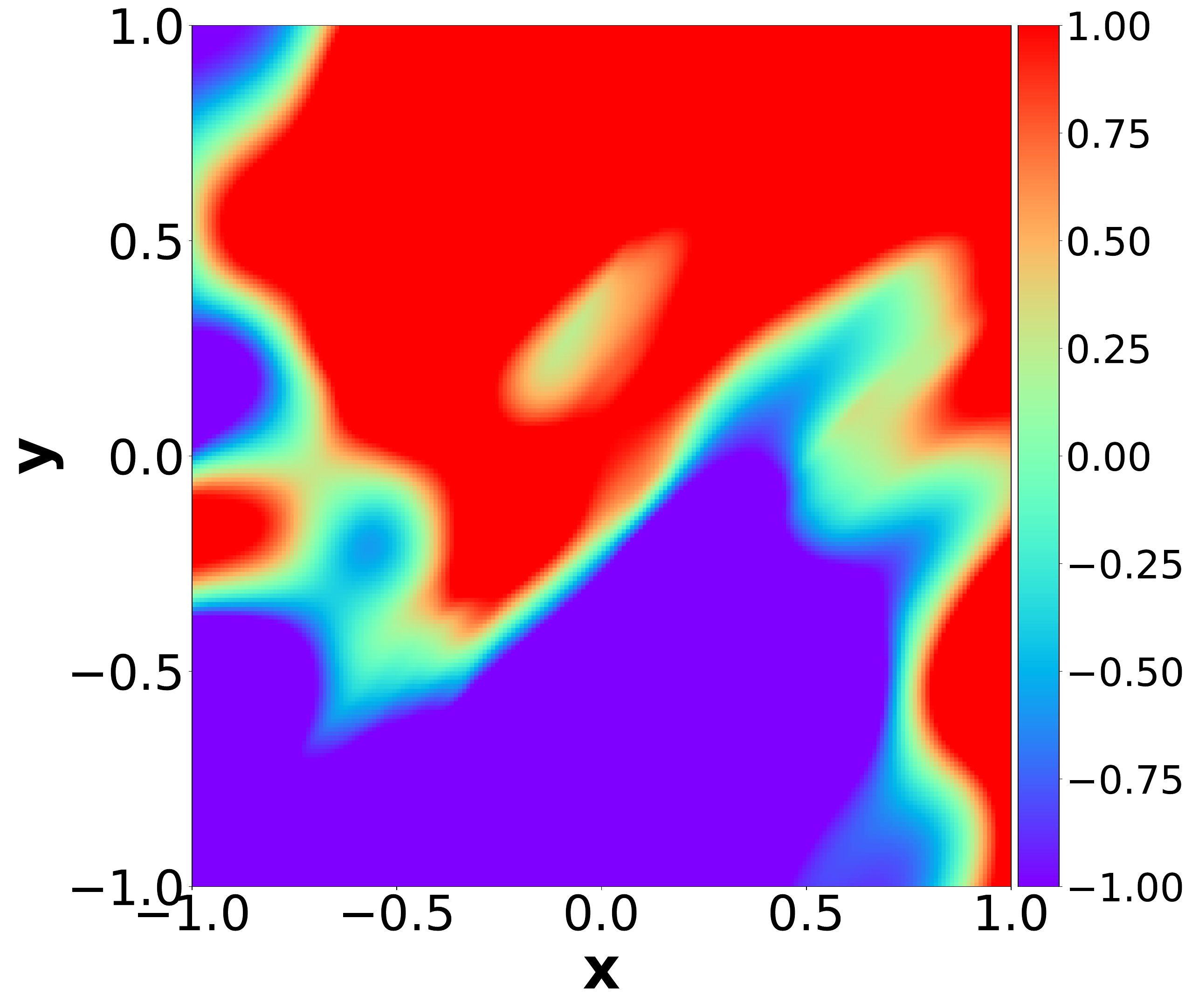}}\hfill
    \subfigure[PINN-R ($a=3.0$)]
    {\includegraphics[width=0.19\columnwidth]{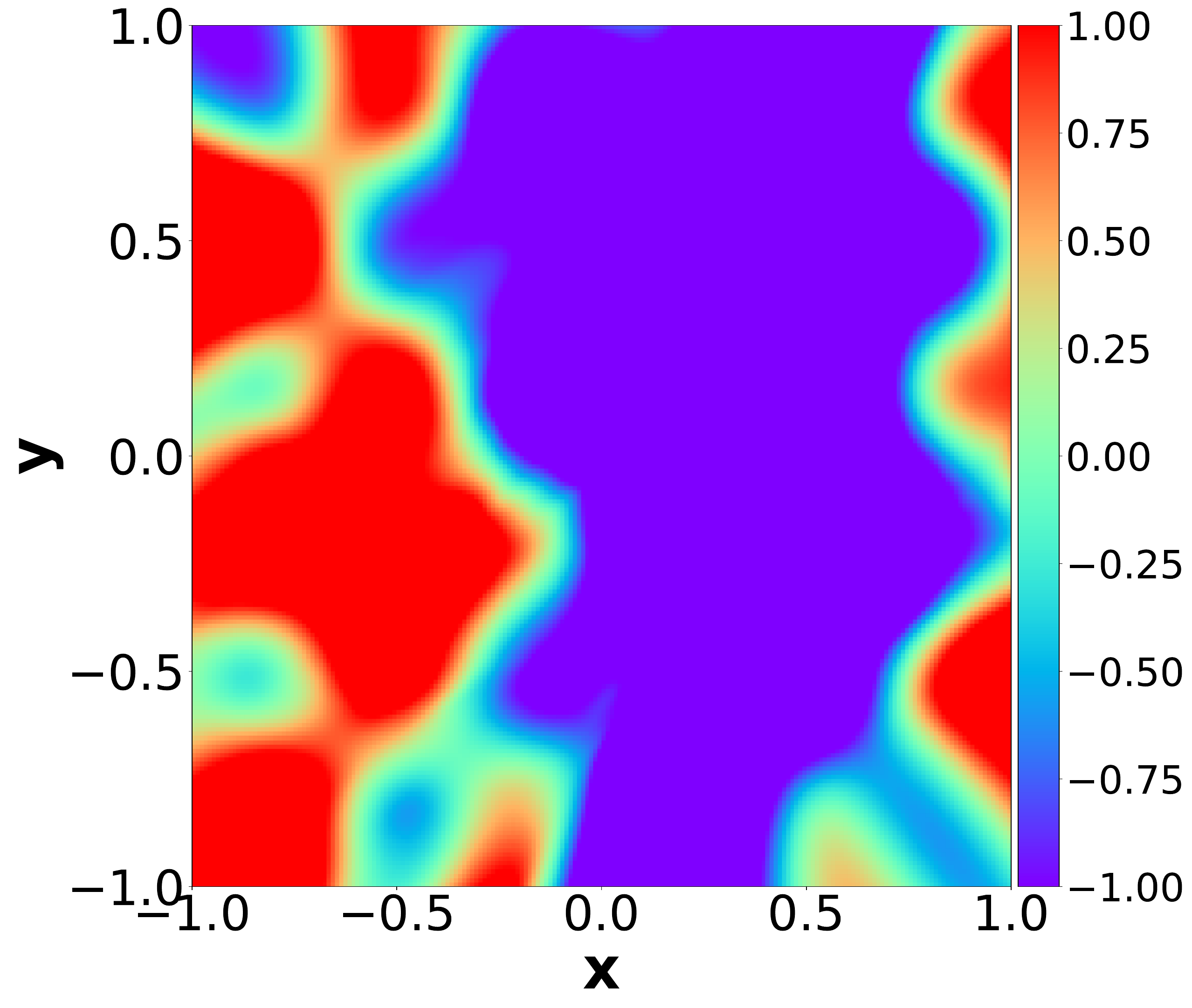}}\\

    \subfigure[P$^2$INN ($a=2.5$)]
    {\includegraphics[width=0.19\columnwidth]{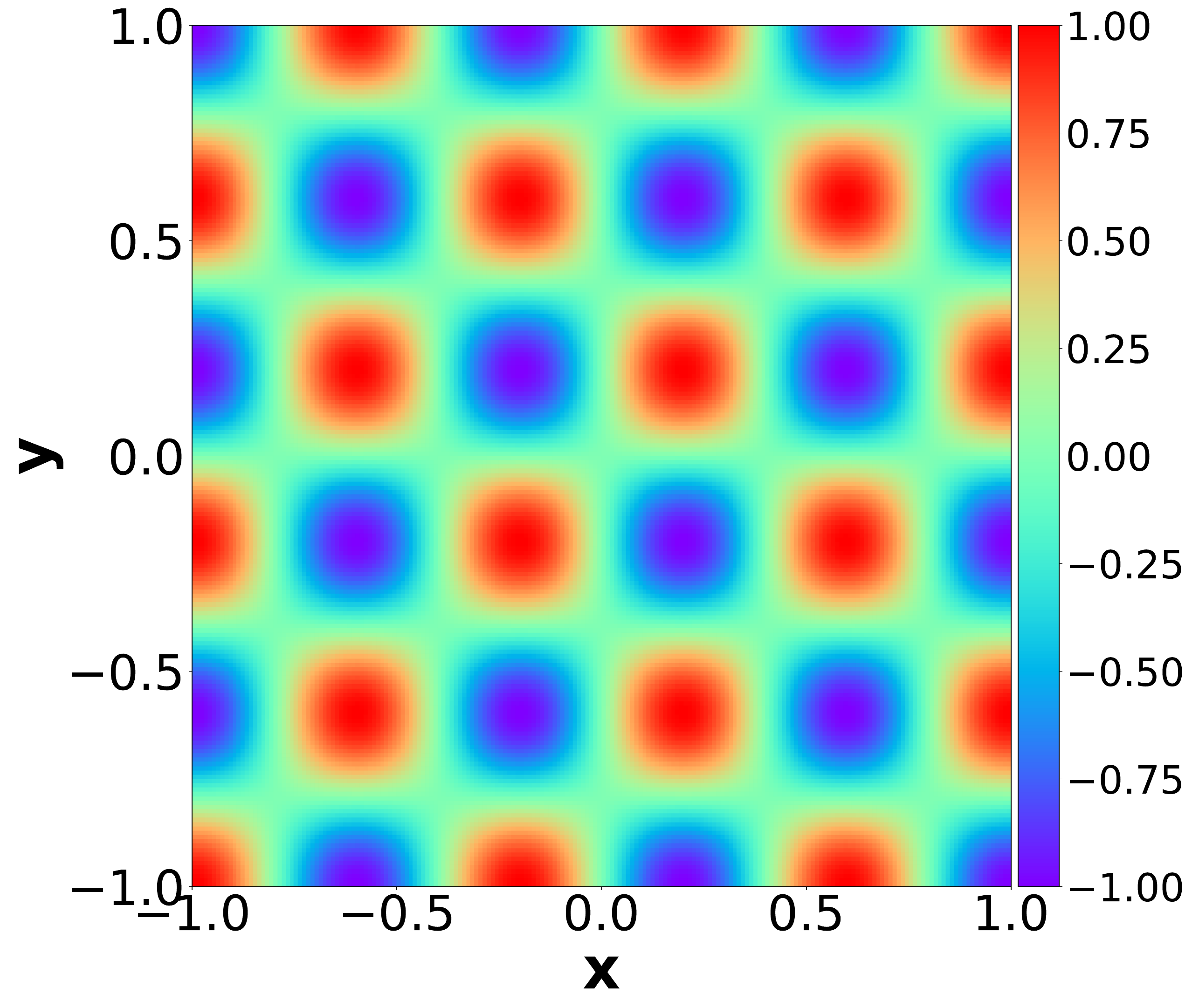}}\hfill
    \subfigure[P$^2$INN ($a=2.65$)]
    {\includegraphics[width=0.19\columnwidth]{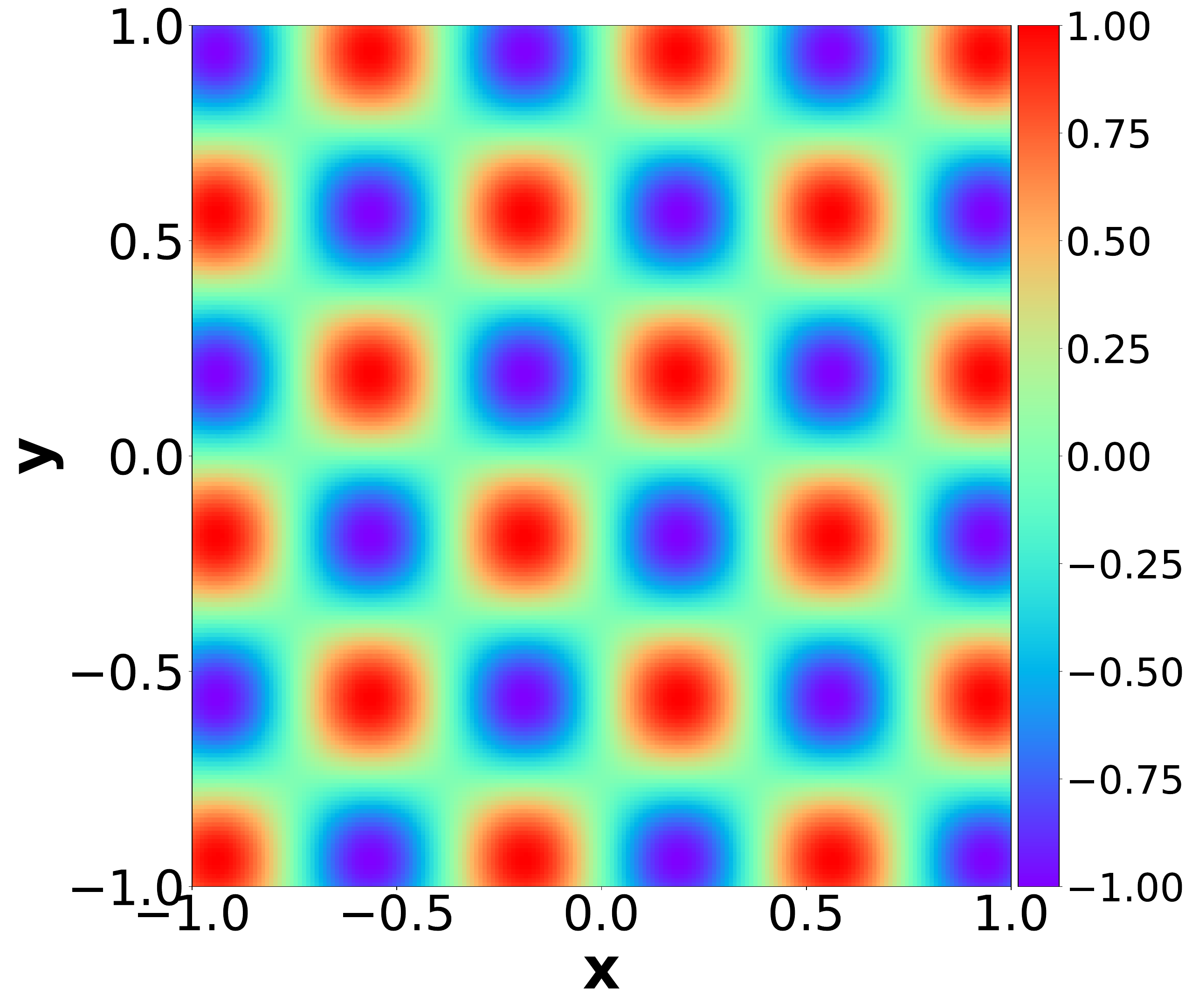}}\hfill
    \subfigure[P$^2$INN ($a=2.75$)]
    {\includegraphics[width=0.19\columnwidth]{images/helm_2.75_2.75.pdf}}\hfill
    \subfigure[P$^2$INN ($a=2.85$)]
    {\includegraphics[width=0.19\columnwidth]{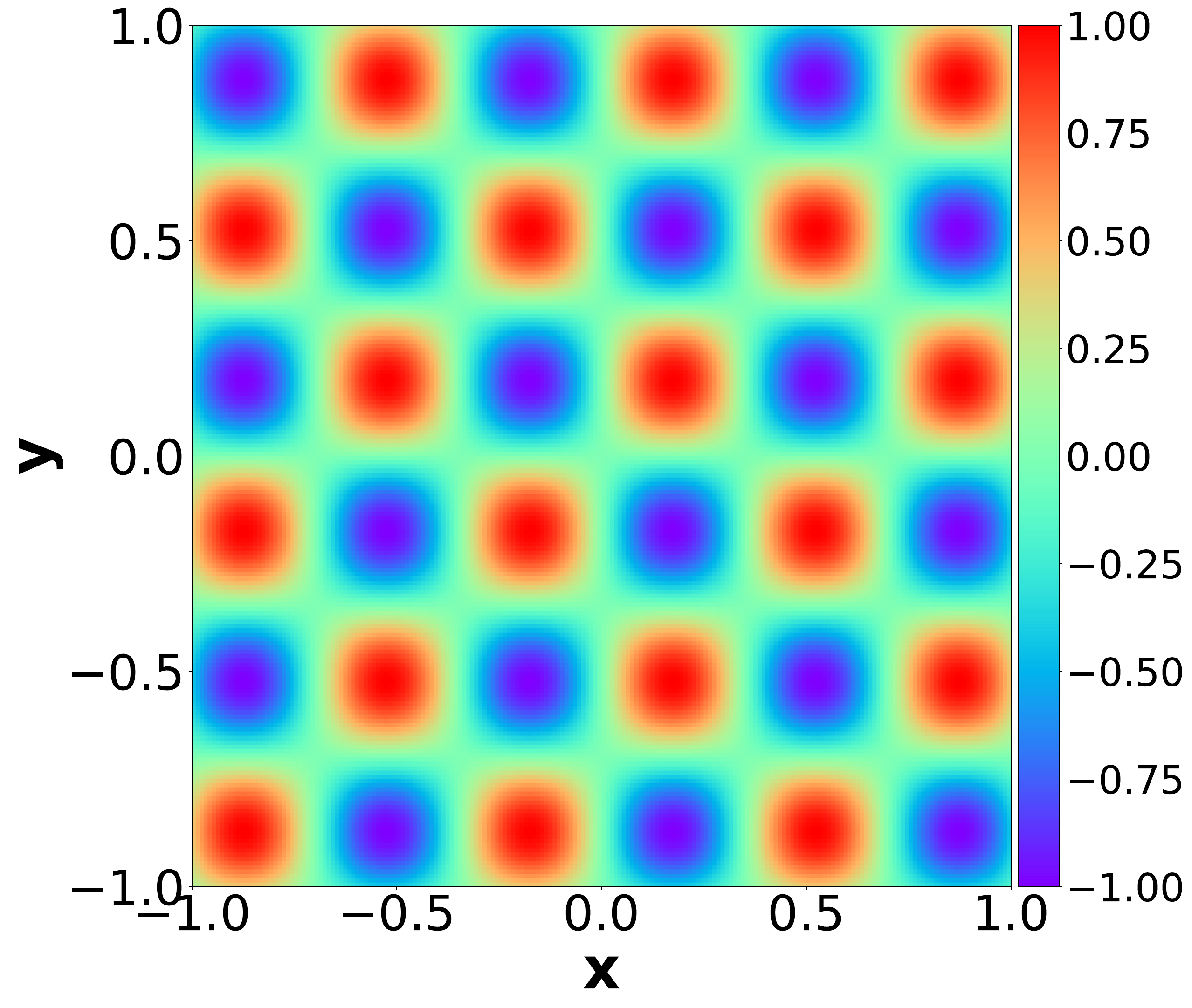}}\hfill
    \subfigure[P$^2$INN ($a=3.0$)]
    {\includegraphics[width=0.19\columnwidth]{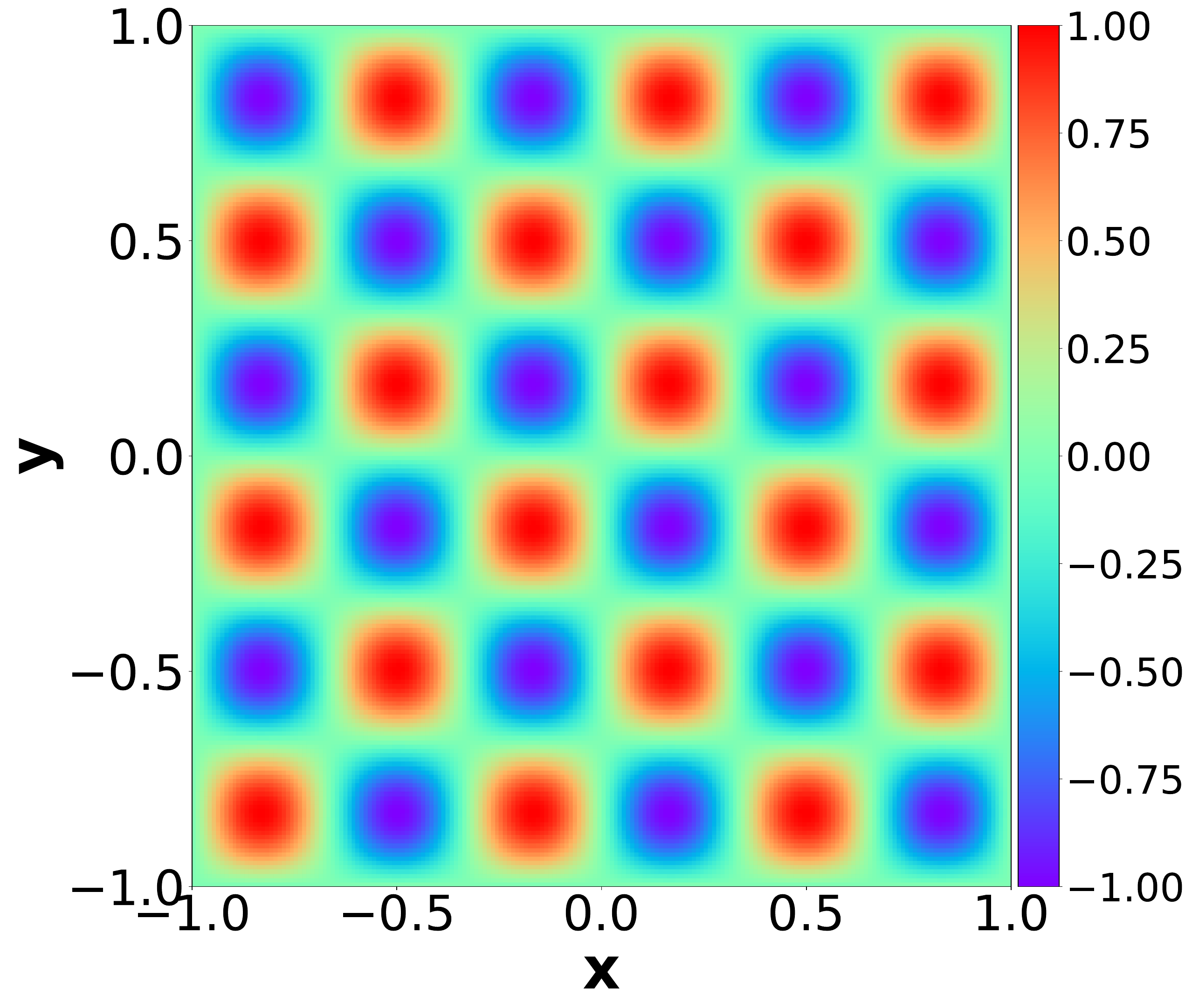}}\\

\caption{[2D-Helmholtz equation] Exact solutions and results of PINN, PINN-R and P$^2$INN for various $a$}\label{fig:helmholtz_vis}
\end{figure}

We undertake an evaluation by training our P$^2$INN model on a 2D Helmholtz equation and subsequently comparing its performance with that of PINNs. In the case of $a=\{2.50, 2.70, 2.80, 3.00\}$, performance is evaluated on the seen PDEs utilized for training, while for $a=\{2.65, 2.75, 2.85\}$, performance is assessed on the unseen PDEs not used during training phase. All test datasets consist of data that is not employed in the training, and the experimental results are reported in Table~\ref{tbl:helmholtz_main} and Figure~\ref{fig:helmholtz_vis}.

\begin{table}[ht!]
\centering
\footnotesize
\renewcommand{\arraystretch}{0.6}
\caption{Comparision with PINN, PINN-R and P$^2$INN on 2D Helmholtz equations}
\begin{tabular}{ccccccccc}
\specialrule{1pt}{2pt}{2pt}
 Model & Metrics & $a=2.50$ & $a=2.65$ & $a=2.70$ & $a=2.75$ & $a=2.80$ & $a=2.85$ & $a=3.00$ \\ \specialrule{1pt}{2pt}{2pt}
\multirow{2}{*}{\textbf{PINN}} & Abs. err. & 0.1484 & 0.9077 & 1.9105 & 1.8942 & 1.5689 & 0.9077 & 2.4981 \\
 & Rel. err. & 0.4817 & 2.0937 & 4.9264 & 4.7584 & 3.3739 & 2.0937 & 6.1532 \\ \midrule
\multirow{2}{*}{\textbf{PINN-R}} & Abs. err. & 0.1107 & 0.2916 & 1.1590 & 1.4000 & 1.1095 & 1.5789 & 1.8800 \\
 & Rel. err. & 0.3830 & 0.7239 & 2.8633 & 3.6641 & 2.6792 & 3.8059 & 4.7755 \\ \midrule
\multirow{2}{*}{\textbf{P$^2$INN}} & Abs. err. & \textbf{0.0240} & \textbf{0.0259} & \textbf{0.0257} & \textbf{0.0263} & \textbf{0.0321} & \textbf{0.0232} & \textbf{0.0315} \\
 & Rel. err. & \textbf{0.0718} & \textbf{0.0767} & \textbf{0.0788} & \textbf{0.0840} & \textbf{0.0975} & \textbf{0.0642} & \textbf{0.0973} \\
\specialrule{1pt}{2pt}{2pt}
\end{tabular}

\label{tbl:helmholtz_main}
\end{table}

% \clearpage

\section{Architectural Details of PINN-P}\label{a:PINN_P_model}
\begin{figure}[ht!]
\centering
\includegraphics[width=0.6\columnwidth]{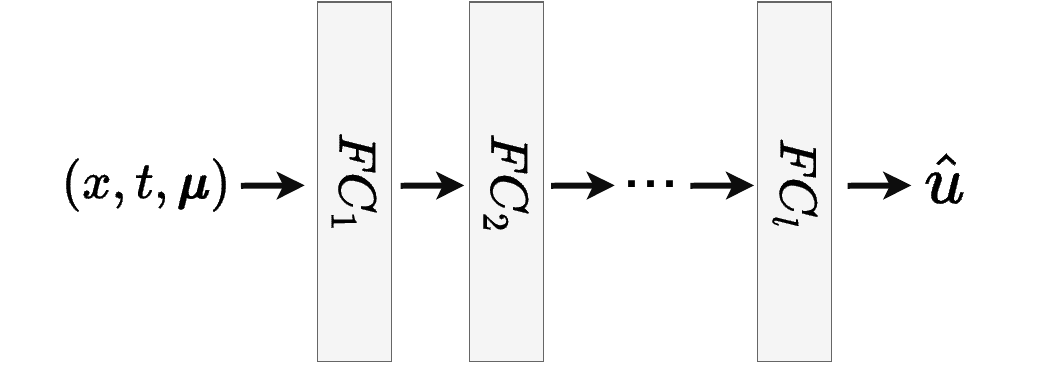}
\caption{PINN-P architecture.}
\label{fig:PINN_Parchi}
\end{figure}

{We propose PINN-P as an ablation model of our P$^2$INN. Unlike P$^2$INN, PINN-P does not have a separate encoder for PDE parameters, so that PDE parameters enter the model with coordinates. As shown in Figure~\ref{fig:PINN_Parchi}, PINN-P consists of $l$-stacked fully-connected layers. For a fair comparison with \PPINN{}, we set size of hidden vector to 150 and $l$ to 6, making the model size similar to \PPINN{}.}

\section{Reproducibility Statement} 
To benefit the community, the code will be posted online. The source code for our proposed method and the dataset used in this paper are attached.

\clearpage

\section{Additional Statistical Analysis on CDR Equations}
In this section, we provide additional statistical analysis on experiments of Table~\ref{tbl:result} with two other metrics, explained variance score (Exp. var.) and max error (Max. err.), and summarize the results in Table~\ref{tbl:gauss_max_exp}.

\begin{table}[h]
\centering
\renewcommand{\arraystretch}{0.75}
\footnotesize
\caption{The max error and explained variance score over all the equations with the initial condition of Gaussian distribution $N(\pi, (\pi/2)^2)$. We do not fine-tune P$^2$INN for each coefficient setting. However, our pre-trained models already outperform others in many cases.}

\begin{tabular}{cccccccc}

\specialrule{1pt}{2pt}{2pt}
\multirow{2}{*}{} & \multirow{2}{*}{\textbf{PDE type}} & \textbf{Coefficient} & \multirow{2}{*}{\textbf{Metric}} & \multirow{2}{*}{\textbf{PINN}} & \multirow{2}{*}{\textbf{PINN-R}}  & \multirow{2}{*}{\textbf{PINN-seq2seq}} & \multirow{2}{*}{\textbf{P$^2$INN}} \\
 &  & \textbf{range}  &  &  &   & &  \\
\specialrule{1pt}{2pt}{2pt}
\multirow{24}{*}{\textbf{Class 1}} & \multirow{8}{*}{\textbf{Convection}} & \multirow{2}{*}{1$\sim$5} 
& Max. err. & 0.0513 & 0.0601 & 0.2751 &  0.0293 \\
 &  &  & Exp. var. & 0.9948 & 0.9919 & 0.6070 & 0.9997 \\ \cmidrule(lr){3-8}
 &  & \multirow{2}{*}{1$\sim$10} & Max. err. & 0.0593 & 0.1681 & 0.3744 & 0.0496 \\
 &  &  & Exp. var. & 0.9952 & 0.8227 & 0.3009 & 0.9985 \\ \cmidrule(lr){3-8}
 &  & \multirow{2}{*}{1$\sim$20} & Max. err. & 0.2431 & 0.3173 & 0.4190 & 0.1430  \\
 &  &  & Exp. var. & 0.6076 & 0.4107 & 0.1500 &  0.9887 \\ \cmidrule(lr){2-8}
 & \multirow{8}{*}{\textbf{Diffusion}} & \multirow{2}{*}{1$\sim$5} & Max. err. & 0.3989 & 0.5190 & 0.5784 &  0.4382  \\
 &  &  & Exp. var. & -0.1383	& -1.0922	& -1.5233 & -0.4871 \\ \cmidrule(lr){3-8}
 &  & \multirow{2}{*}{1$\sim$10} & Max. err. & 0.6194 & 0.7576 & 0.7638 & -0.4871 \\
 &  &  & Exp. var. & -5.4421	& -8.4565	& -7.6938	& -2.0589 \\ \cmidrule(lr){3-8}
 &  & \multirow{2}{*}{1$\sim$20} & Max. err. & 1.3538 & 1.4417 & 0.7564 & 0.4516 \\
 &  &  & Exp. var. & -79.3751 & -85.9107 & -13.9281 & -6.3671 \\ \cmidrule(lr){2-8}
 & \multirow{8}{*}{\textbf{Reaction}} & \multirow{2}{*}{1$\sim$5} & Max. err. & 0.4053 & 0.4093 & 0.7653 & 0.0142 \\ 
 &  &  & Exp. var. & 0.6232 & 0.6123 & -0.1824 &  0.9999 \\ \cmidrule(lr){3-8}
 &  & \multirow{2}{*}{1$\sim$10} & Max. err. & 0.7040 & 0.5116 & 0.8847 & 0.0376  \\
 &  &  & Exp. var. & 0.3008 & 0.5036 & -0.0936 & 0.9995 \\ \cmidrule(lr){3-8}
 &  & \multirow{2}{*}{1$\sim$20} & Max. err. & 0.8529 & 0.7022 & 0.9429 & 0.0846 \\
 &  &  & Exp. var. & 0.1492 & 0.2561 & -0.0478 & 0.9977 \\

\specialrule{1pt}{2pt}{2pt}

\multirow{16}{*}{\textbf{Class 2}} & \multirow{8}{*}{\textbf{Conv.-Diff.}} & \multirow{2}{*}{1$\sim$5} & Max. err. & 0.1412 & 0.1601 & 0.3062 & 0.0399 \\
 &  &  & Exp. var. & 0.7469 & 0.6008 & 0.2710 & 0.0892 \\ \cmidrule(lr){3-8}
 &  & \multirow{2}{*}{1$\sim$10} & Max. err. & 0.2171 & 0.3023 & 0.3634 & 0.3169 \\
 &  &  & Exp. var. & -0.4752 & -1.2314 & -0.3168 & 0.6039 \\ \cmidrule(lr){3-8}
 &  & \multirow{2}{*}{1$\sim$20} & Max. err. & 0.5396 & 0.5468 & 0.3606 &  0.4549\\
 &  &  & Exp. var. & -19.0867 & -17.8158 & -0.4771 &  0.0253\\ \cmidrule(lr){2-8}

 & \multirow{8}{*}{\textbf{Reac.-Diff.}} & \multirow{2}{*}{1$\sim$5} & Max. err. & 0.4901 & 0.5272 & 0.7646 & 0.4335 \\
 &  &  & Exp. var. & 0.1600 & -0.2223 & -0.3109 & 0.5052 \\ \cmidrule(lr){3-8}
 &  & \multirow{2}{*}{1$\sim$10} & Max. err.& 0.8332 & 0.7536 & 0.8982 & 0.7501 \\
 &  &  & Exp. var. & -1.0339 & -1.0530 & -0.6734 & -0.4028 \\ \cmidrule(lr){3-8}
 &  & \multirow{2}{*}{1$\sim$20} & Max. err. & 0.9637 & 0.8345 & 0.9445 &  1.0133\\
 &  &  & Exp. var. & -1.8660 & -1.4076 & -0.5779 & -3.4065\\
 
\specialrule{1pt}{2pt}{2pt}

\multirow{8}{*}{\textbf{Class 3}} & \multirow{8}{*}{\textbf{Conv.-Diff.-Reac.}} & \multirow{2}{*}{1$\sim$5} & Max. err. & 0.2694 & 0.3267 & 0.6831 & 0.2307 \\
 &  &  & Exp. var. & 0.4267 & 0.5533 & -0.2518 & 0.9333 \\ \cmidrule(lr){3-8}
 &  & \multirow{2}{*}{1$\sim$10} & Max. err. & 0.6367 & 0.5262 & 0.8418 & 0.5249 \\
 &  &  & Exp. var.& 0.1239 & 0.3869 & -0.1386 & 0.7111 \\ \cmidrule(lr){3-8}
 &  & \multirow{2}{*}{1$\sim$20} & Max. err. & 0.8192 & 0.6859 & 0.9213 & 0.7230 \\
 &  &  & Exp. var. & 0.0587 & 0.2529 & -0.0720 & 0.6803 \\
\specialrule{1pt}{2pt}{2pt}

\end{tabular}
\label{tbl:gauss_max_exp}
\end{table}

\clearpage
\section{Experiments on Another Initial Condition}
In this section, we present the experimental results with the initial condition of 1+$\sin(x)$, with other conditions remain unchanged. In Table~\ref{tbl:sin_abs_rel}, we summarize the results in terms of $L_2$ absolute and relative errors, and in Table~\ref{tbl:sin_exp_var}, we use max error and explained variance score for evaluation. Even with the initial condition of 1+$\sin(x)$, our \PPINN{} show adequate performance, especially in a wide coefficient range, i.e., 1$\sim$20. On top of that, particularly in the case of reaction and conv.-diff.-reac. equations, \PPINN{} outperform the baselines by a huge gap.

\begin{table}[ht!]
\centering
\renewcommand{\arraystretch}{0.75}
\footnotesize

\caption{The relative and absolute $L_2$ errors over all the equations with the initial condition of 1+$\sin(x)$. We do not fine-tune P$^2$INN for each coefficient setting. However, our pre-trained models already outperform others in many cases.}

\begin{tabular}{cccccccc}

\specialrule{1pt}{2pt}{2pt}
\multirow{2}{*}{} & \multirow{2}{*}{\textbf{PDE type}} & \textbf{Coefficient} & \multirow{2}{*}{\textbf{Metric}} & \multirow{2}{*}{\textbf{PINN}} & \multirow{2}{*}{\textbf{PINN-R}}  & \multirow{2}{*}{\textbf{PINN-seq2seq}} & \multirow{2}{*}{\textbf{P$^2$INN}} \\
 &  & \textbf{range}  &  &  &   & &  \\
 
\specialrule{1pt}{2pt}{2pt}
\multirow{24}{*}{\textbf{Class 1}} & \multirow{8}{*}{\textbf{Convection}} & \multirow{2}{*}{1$\sim$5} & Abs. err. &0.0135 & 0.0073 & 0.2213 & 0.0045  \\
 &  &  & Rel. err. &0.0147 & 0.0076 & 0.2159 & 0.0044  \\ \cmidrule(lr){3-8}
 &  & \multirow{2}{*}{1$\sim$10} & Abs. err. & 0.0117 & 0.0233 & 0.4101 & 0.0095  \\
 &  &  & Rel. err. &  0.0127 & 0.0244 & 0.3821 & 0.0092 \\ \cmidrule(lr){3-8}
 &  & \multirow{2}{*}{1$\sim$20} & Abs. err. & 0.1283 & 0.2558 & 0.5355 & 0.0826   \\
 &  &  & Rel. err. & 0.1295 & 0.2503 & 0.5092 & 0.0827   \\ \cmidrule(lr){2-8}
 & \multirow{8}{*}{\textbf{Diffusion}} & \multirow{2}{*}{1$\sim$5} &Abs. err. & 0.0496 & 0.0688 & 0.2682 & 0.4027    \\
 &  &  & Rel. err. & 0.0714 & 0.0956 & 0.3048 & 0.4608    \\ \cmidrule(lr){3-8}
 &  & \multirow{2}{*}{1$\sim$10} & Abs. err. & 0.0835 & 0.1106 & 0.3047 & 0.4863    \\
 &  &  & Rel. err. & 0.1078 & 0.1416 & 0.3356 & 0.5487   \\ \cmidrule(lr){3-8}
 &  & \multirow{2}{*}{1$\sim$20} & Abs. err. & 0.1206 & 0.2045 & 0.3015 & 0.5254   \\
 &  &  & Rel. err. & 0.1500 & 0.2333 & 0.3308 & 0.5963   \\ \cmidrule(lr){2-8}
 & \multirow{8}{*}{\textbf{Reaction}} & \multirow{2}{*}{1$\sim$5} & Abs. err. &0.2349 & 0.2396 & 0.5100 & 0.0254  \\ 
 &  &  & Rel. err. & 0.3241 & 0.3316 & 0.6117 & 0.0736   \\ \cmidrule(lr){3-8}
 &  & \multirow{2}{*}{1$\sim$10} & Abs. err. & 0.5688 & 0.3418 & 0.7054 & 0.0617   \\
 &  &  & Rel. err. &0.6561 & 0.4494 & 0.7994 & 0.1199   \\ \cmidrule(lr){3-8}
 &  & \multirow{2}{*}{1$\sim$20} & Abs. err. & 0.7615 & 0.4290 & 0.8294 & 0.1487  \\
 &  &  & Rel. err. & 0.8274 & 0.5268 & 0.8987 & 0.2901   \\

\specialrule{1pt}{2pt}{2pt}

\multirow{16}{*}{\textbf{Class 2}} & \multirow{8}{*}{\textbf{Conv.-Diff.}} & \multirow{2}{*}{1$\sim$5} & Abs. err. & 0.0213 & 0.0182 & 0.2382 & 0.1997   \\
 &  &  & Rel. err. & 0.0261 & 0.0233 & 0.2605 & 0.2296  \\ \cmidrule(lr){3-8}
 &  & \multirow{2}{*}{1$\sim$10} & Abs. err.& 0.0294 & 0.0345 & 0.2316 & 0.1297  \\
 &  &  & Rel. err. & 0.0350 & 0.0410 & 0.2574 & 0.1514 \\ \cmidrule(lr){3-8}
 &  & \multirow{2}{*}{1$\sim$20} & Abs. err. & 0.0824 & 0.0960 & 0.2504 & 0.1173  \\
 &  &  & Rel. err. & 0.0928 & 0.1094 & 0.2741 & 0.1339   \\ \cmidrule(lr){2-8}

 & \multirow{8}{*}{\textbf{Reac.-Diff.}} & \multirow{2}{*}{1$\sim$5} & Abs. err. & 0.0386 & 0.0368 & 0.3038 & 0.1150  \\
 &  &  & Rel. err. & 0.0632 & 0.0581 & 0.3562 & 0.1566  \\ \cmidrule(lr){3-8}
 &  & \multirow{2}{*}{1$\sim$10} & Abs. err. & 0.2305 & 0.1632 & 0.5427 & 0.0816   \\
 &  &  & Rel. err.& 0.2681 & 0.2054 & 0.5807 & 0.1190   \\ \cmidrule(lr){3-8}
 &  & \multirow{2}{*}{1$\sim$20} & Abs. err. & 0.5454 & 0.3107 & 0.6892 & 0.0413  \\
 &  &  & Rel. err. & 0.5710 & 0.3497 & 0.7131 & 0.0727   \\

\specialrule{1pt}{2pt}{2pt}

 \multirow{8}{*}{\textbf{Class 3}} & \multirow{8}{*}{\textbf{Conv.-Diff.-Reac.}} & \multirow{2}{*}{1$\sim$5} & Abs. err. & 0.0593 & 0.0932 & 0.3626 & 0.0551   \\
 &  &  & Rel. err. & 0.0899 & 0.1465 & 0.4554 & 0.0735   \\ \cmidrule(lr){3-8}
 &  & \multirow{2}{*}{1$\sim$10} & Abs. err. & 0.4953 & 0.3407 & 0.6500 & 0.0337   \\
 &  &  & Rel. err. & 0.5341 & 0.4006 & 0.7196 & 0.0538   \\ \cmidrule(lr){3-8}
 &  & \multirow{2}{*}{1$\sim$20} & Abs. err. & 0.7329 & 0.4283 & 0.8101 & 0.0614   \\
 &  &  & Rel. err. & 0.7663 & 0.4911 & 0.8588 & 0.1240  \\ 
 
\specialrule{1pt}{2pt}{2pt}

\end{tabular}
\label{tbl:sin_abs_rel}
\end{table}

\clearpage

\begin{table}[ht!]
\centering
\renewcommand{\arraystretch}{0.75}
\footnotesize

\caption{The max error and explained variance score over all the equations with the initial condition of 1+$\sin(x)$. We do not fine-tune P$^2$INN for each coefficient setting. However, our pre-trained models already outperform others in many cases.}

\begin{tabular}{cccccccc}

\specialrule{1pt}{2pt}{2pt}
\multirow{2}{*}{} & \multirow{2}{*}{\textbf{PDE type}} & \textbf{Coefficient} & \multirow{2}{*}{\textbf{Metric}} & \multirow{2}{*}{\textbf{PINN}} & \multirow{2}{*}{\textbf{PINN-R}}  & \multirow{2}{*}{\textbf{PINN-seq2seq}} & \multirow{2}{*}{\textbf{P$^2$INN}} \\
 &  & \textbf{range}  &  &  &   & &  \\
 
\specialrule{1pt}{2pt}{2pt}
\multirow{24}{*}{\textbf{Class 1}} & \multirow{8}{*}{\textbf{Convection}} & \multirow{2}{*}{1$\sim$5} & Max. err. & 0.0531 & 0.0316 & 0.6235 & 0.0141 \\
 &  &  & Exp. var. & 0.9993 & 0.9998 & 0.8369 & 1.0000  \\ \cmidrule(lr){3-8}
 &  & \multirow{2}{*}{1$\sim$10} & Max. err. & 0.0479 & 0.0919 & 0.8839 & 0.0358  \\
 &  &  & Exp. var. & 0.9994 & 0.9947 & 0.4700 & 0.9998  \\ \cmidrule(lr){3-8}
 &  & \multirow{2}{*}{1$\sim$20} & Max. err. & 0.3495 & 0.5931 & 1.1652 & 0.2295  \\
 &  &  & Exp. var. & 0.8662 & 0.6638 & 0.2362 & 0.9681 \\ \cmidrule(lr){2-8}
 & \multirow{8}{*}{\textbf{Diffusion}} & \multirow{2}{*}{1$\sim$5} & Max. err. & 0.2706 & 0.3166 & 0.6683 & 0.9959  \\
 &  &  & Exp. var. & 0.9322 & 0.8433 & 0.3848 & -1.3721  \\ \cmidrule(lr){3-8}
 &  & \multirow{2}{*}{1$\sim$10} & Max. err. & 0.3190 & 0.3693 & 0.6645 & 1.0063  \\
 &  &  & Exp. var. & 0.7238 & 0.5667 & 0.0207 & -4.0842  \\ \cmidrule(lr){3-8}
 &  & \multirow{2}{*}{1$\sim$20} & Max. err. & 0.4056 & 0.4829 & 0.6612 & 0.9708  \\
 &  &  & Exp. var. & 0.0209 & -0.4390 & -0.5409 & -21.5206  \\ \cmidrule(lr){2-8}
 & \multirow{8}{*}{\textbf{Reaction}} & \multirow{2}{*}{1$\sim$5} & Max. err. & 0.8754 & 0.8815 & 1.1852 & 0.8770  \\ 
 &  &  & Exp. var. & 0.6996 & 0.6760 & 0.1010 & 0.9692  \\ \cmidrule(lr){3-8}
 &  & \multirow{2}{*}{1$\sim$10} & Max. err. & 1.2884 & 1.0865 & 1.4784 & 1.0001 \\
 &  &  & Exp. var. & 0.3725 & 0.3932 & 0.0618 & 0.8961  \\ \cmidrule(lr){3-8}
 &  & \multirow{2}{*}{1$\sim$20} & Max. err. & 1.4931 & 1.2158 & 1.5880 & 1.0000 \\
 &  &  & Exp. var. & 0.1881 & 0.2003 & 0.0331 & 0.0000  \\

\specialrule{1pt}{2pt}{2pt}

\multirow{16}{*}{\textbf{Class 2}} & \multirow{8}{*}{\textbf{Conv.-Diff.}} & \multirow{2}{*}{1$\sim$5} & Max. err. & 0.0811 & 0.0702 & 0.6079 & 0.6515  \\
 &  &  & Exp. var. & 0.9914 & 0.9888 & 0.5136 & 0.4635  \\ \cmidrule(lr){3-8}
 &  & \multirow{2}{*}{1$\sim$10} & Max. err. & 0.1002 & 0.1154 & 0.6613 & 0.5590 \\
 &  &  & Exp. var. & 0.9641 & 0.9532 & 0.2876 & 0.6620  \\ \cmidrule(lr){3-8}
 &  & \multirow{2}{*}{1$\sim$20} & Max. err. & 0.2583 & 0.3162 & 0.7221 & 0.3527  \\
 &  &  & Exp. var. & 0.7337 & 0.6718 & -0.1857 & 0.1614  \\ \cmidrule(lr){2-8}

 & \multirow{8}{*}{\textbf{Reac.-Diff.}} & \multirow{2}{*}{1$\sim$5} & Max. err. & 0.2870 & 0.2659 & 0.9199 & 0.5404  \\
 &  &  & Exp. var. & 0.8536 & 0.8962 & 0.2670 & 0.5430  \\ \cmidrule(lr){3-8}
 &  & \multirow{2}{*}{1$\sim$10} & Max. err.& 0.7232 & 0.6471 & 1.2166 & 0.5853 \\
 &  &  & Exp. var. & 0.2866 & 0.1213 & 0.1293 & 0.4415  \\ \cmidrule(lr){3-8}
 &  & \multirow{2}{*}{1$\sim$20} & Max. err. & 1.1176 & 0.9268 & 1.3338 & 0.3204 \\
 &  &  & Exp. var. & 0.1430 & -0.0407 & -0.0576 & 0.3938 \\
 
\specialrule{1pt}{2pt}{2pt}

\multirow{8}{*}{\textbf{Class 3}} & \multirow{8}{*}{\textbf{Conv.-Diff.-Reac.}} & \multirow{2}{*}{1$\sim$5} & Max. err. & 0.3264 & 0.5064 & 1.1676 & 0.3694  \\
 &  &  & Exp. var. & 0.9039 & 0.6877 & -0.0153 & 0.8884  \\ \cmidrule(lr){3-8}
 &  & \multirow{2}{*}{1$\sim$10} & Max. err. & 1.0318 & 0.9348 & 1.4829 & 0.4511  \\
 &  &  & Exp. var.& 0.4165 & 0.3708 & -0.0199 & 0.8822  \\ \cmidrule(lr){3-8}
 &  & \multirow{2}{*}{1$\sim$20} & Max. err. & 1.3732 & 1.1398 & 1.5986 & 0.9006  \\
 &  &  & Exp. var. & 0.2066 & 0.1825 & -0.0091 & 0.0134  \\
\specialrule{1pt}{2pt}{2pt}

\end{tabular}
\label{tbl:sin_exp_var}
\end{table}

\clearpage

\section{Standard Deviation of the Evaluation Metrics from Table~\ref{tbl:result}}

In Table~\ref{tbl:result_std}, we report the standard deviation of evaluation metrics, Abs. err. and Rel. err., from Table~\ref{tbl:result}.  We conduct the experiments in Table~\ref{tbl:result} with three different random seeds.

\begin{table}[ht!]
\centering
\setlength{\tabcolsep}{5.2pt}
\renewcommand{\arraystretch}{0.8}
\footnotesize
\caption{Standard deviation of evaluation metrics from Table~\ref{tbl:result}.}

\begin{tabular}{cccccccc}

\specialrule{1pt}{2pt}{2pt}
\multirow{2}{*}{} & \multirow{2}{*}{\textbf{PDE type}} & \textbf{Coefficient} & \multirow{2}{*}{\textbf{Metric}} & \multirow{2}{*}{\textbf{PINN}} & \multirow{2}{*}{\textbf{PINN-R}}  & \multirow{2}{*}{\textbf{PINN-seq2seq}} & \multirow{2}{*}{\textbf{P$^2$INN}} \\
 &  & \textbf{range}  &  &  &   & &  \\

\specialrule{1pt}{2pt}{2pt}
\multirow{24}{*}{\textbf{Class 1}} & \multirow{8}{*}{\textbf{Convection}} & \multirow{2}{*}{1$\sim$5} 
                                    & Abs. err. & 0.0012  & 0.0058 &	 0.0130& 	 0.0005    \\
 &  &                               & Rel. err. & 0.0022	 & 0.0095 &	0.0221	& 0.0008	   \\ \cmidrule(lr){3-8}
 &  & \multirow{2}{*}{1$\sim$10}    & Abs. err. & 0.0022	 & 0.0177 &	0.0065	& 0.0039	   \\
 &  &                               & Rel. err. & 0.0051	 & 0.0305 &	0.0109	& 0.0072	   \\ \cmidrule(lr){3-8}
 &  & \multirow{2}{*}{1$\sim$20}    & Abs. err. & 0.0021	 & 0.0066 &	0.0032	& 0.0064	   \\
 &  &                               & Rel. err. & 0.0012	 & 0.0125 &	0.0055	& 0.0100	   \\ \cmidrule(lr){2-8}

 & \multirow{8}{*}{\textbf{Diffusion}} & \multirow{2}{*}{1$\sim$5}                                    
                                    & Abs. err. &  0.0023	&0.0156&	0.0066	& 0.0010   \\
 &  &                               & Rel. err. &  0.0068	&0.0321&	0.0142	& 0.0013   \\ \cmidrule(lr){3-8}
 &  & \multirow{2}{*}{1$\sim$10}    & Abs. err. &  0.0178	&0.0282&	0.0116	& 0.0001   \\
 &  &                               & Rel. err. &  0.0326	&0.0583&	0.0225	& 0.0001   \\ \cmidrule(lr){3-8}
 &  & \multirow{2}{*}{1$\sim$20}    & Abs. err. &  0.0093	&0.0115&	0.0183	& 0.0001   \\
 &  &                               & Rel. err. &  0.0205	&0.0244&	0.0358	& 0.0002   \\ \cmidrule(lr){2-8}

 & \multirow{8}{*}{\textbf{Reaction}} & \multirow{2}{*}{1$\sim$5} 
                                    & Abs. err. &  0.0018	& 0.1605 &	0.0159 & 	0.0025	 \\
 &  &                               & Rel. err. &  0.0023	& 0.1853 &	0.0087 & 	0.0030	 \\ \cmidrule(lr){3-8}
 &  & \multirow{2}{*}{1$\sim$10}    & Abs. err. &  0.0011	& 0.0915 &	0.0080 & 	0.0017	 \\
 &  &                               & Rel. err. &  0.0014	& 0.1009 &	0.0044 & 	0.0019	 \\ \cmidrule(lr){3-8}
 &  & \multirow{2}{*}{1$\sim$20}    & Abs. err. &  0.0005	& 0.0199 &	0.0040 & 	0.0649	 \\
 &  &                               & Rel. err. &  0.0007	& 0.0258 &	0.0022 & 	0.1445	 \\ 

\specialrule{1pt}{2pt}{2pt}

\multirow{16}{*}{\textbf{Class 2}} & \multirow{8}{*}{\textbf{Conv.-Diff.}} & \multirow{2}{*}{1$\sim$5} 
                                    & Abs. err. &  0.0018 & 	0.0021 &	0.0023 &	0.0097   \\
 &  &                               & Rel. err. &  0.0026 & 	0.0045 &	0.0034 &	0.0176   \\ \cmidrule(lr){3-8}
 &  & \multirow{2}{*}{1$\sim$10}    & Abs. err. &  0.0033 & 	0.0105 &	0.0023 &	0.0035   \\
 &  &                               & Rel. err. &  0.0064 & 	0.0194 &	0.0043 &	0.0037   \\ \cmidrule(lr){3-8}
 &  & \multirow{2}{*}{1$\sim$20}    & Abs. err. &  0.0012 & 	0.0066 &	0.0007 &	0.0055   \\
 &  &                               & Rel. err. &  0.0018 & 	0.0117 &	0.0013 &	0.0146   \\ \cmidrule(lr){2-8}

 & \multirow{8}{*}{\textbf{Reac.-Diff.}} & \multirow{2}{*}{1$\sim$5} 
                                    & Abs. err. & 0.0100 &	0.0322 &	0.0126 &	0.0245   \\
 &  &                               & Rel. err. & 0.0135 &	0.0352 &	0.0105 &	0.0420   \\ \cmidrule(lr){3-8}
 &  & \multirow{2}{*}{1$\sim$10}    & Abs. err. & 0.0045 &	0.0174 &	0.0043 &	0.0507   \\
 &  &                               & Rel. err. & 0.0025 &	0.0182 &	0.0046 &	0.0745   \\ \cmidrule(lr){3-8}
 &  & \multirow{2}{*}{1$\sim$20}    & Abs. err. & 0.0007 &	0.0070 &	0.0018 &	0.2010   \\
 &  &                               & Rel. err. & 0.0011 &	0.0070 &	0.0030 &	0.2114   \\ 
 
\specialrule{1pt}{2pt}{2pt}

\multirow{8}{*}{\textbf{Class 3}} & \multirow{8}{*}{\textbf{Conv.-Diff.-Reac.}} & \multirow{2}{*}{1$\sim$5} 
                                    & Abs. err. & 0.0072 & 0.0015  & 0.0042 & 0.0054  \\
 &  &                               & Rel. err. & 0.0071 & 0.0024  & 0.0021 & 0.0062  \\ \cmidrule(lr){3-8}
 &  & \multirow{2}{*}{1$\sim$10}    & Abs. err. & 0.0102 & 0.0022  & 0.0055 & 0.0021  \\
 &  &                               & Rel. err. & 0.0118 & 0.0047  & 0.0062 & 0.0037  \\ \cmidrule(lr){3-8}
 &  & \multirow{2}{*}{1$\sim$20}    & Abs. err. & 0.0034 & 0.0082  & 0.0008 & 0.0024  \\
 &  &                               & Rel. err. & 0.0047 & 0.0077  & 0.0011 & 0.0041  \\ 
\specialrule{1pt}{2pt}{2pt}

\end{tabular}

\label{tbl:result_std}
\end{table}

\clearpage
\section{Ablation Studies on PINN-P and LargePINN}\label{sec:abl_p2inn}

\begin{table}[ht!]
\centering
\footnotesize
\renewcommand{\arraystretch}{0.7}
\caption{Number of model parameters.}
\begin{tabular}{ccccccc}
\specialrule{1pt}{2pt}{2pt}
 & \textbf{PINN} & \textbf{PINN-R} & \textbf{PINN-seq2seq} & \textbf{LargePINN} & \textbf{PINN-P} & \textbf{P$^2$INN} \\ \cmidrule{1-7}
\#params & 10,401 & 10,401 & 10,401 & 82,941 & 91,651 & 76,851 \\
\specialrule{1pt}{2pt}{2pt}
\end{tabular}
\label{tbl:param}
\end{table}

\begin{table}[ht!]
\centering
\footnotesize
\caption{The relative and absolute $L_2$ errors over all the equations. Our \PPINN{} surpass LargePINN and PINN-P in all but one cases, even without fine-tuning.}
\begin{tabular}{ccccccc}
\specialrule{1pt}{2pt}{2pt}
 & \textbf{PDE type} & \textbf{Metric} & \textbf{PINN} & \textbf{LargePINN} & \textbf{PINN-P} & \textbf{P2INN} \\
\specialrule{1pt}{2pt}{2pt}
\multirow{6}{*}{\textbf{Class 1}} & \multirow{2}{*}{\textbf{Convection}} & Abs. err. & 0.1140 & 0.1191 & 0.0209 & \textbf{0.0198} \\
 &  & Rel. err. & 0.1978 & 0.2084 & \textbf{0.0410} & 0.0464 \\ \cmidrule(lr){2-7}
 & \multirow{2}{*}{\textbf{Diffusion}} & Abs. err. & 0.6782 & 0.5868 & 0.3800 & \textbf{0.1916} \\
 &  & Rel. err. & 1.2825 & 1.0994 & 0.7912 & \textbf{0.3745} \\ \cmidrule(lr){2-7}
 & \multirow{2}{*}{\textbf{Reaction}} & Abs. err. & 0.7902 & 0.7910 & 0.8975 & \textbf{0.0042} \\
 &  & Rel. err. & 0.8460 & 0.8469 & 0.9908 & \textbf{0.0092} \\
 \specialrule{1pt}{2pt}{2pt}
\multirow{4}{*}{\textbf{Class 2}} & \multirow{2}{*}{\textbf{Conv.-Diff.}} & Abs. err. & 0.2735 & 0.1626 & 0.1253 & \textbf{0.0622} \\
 &  & Rel. err. & 0.5106 & 0.3189 & 0.3009 & \textbf{0.1495} \\ \cmidrule(lr){2-7}
 & \multirow{2}{*}{\textbf{Reac.-Diff.}} & Abs. err. & 0.7167 & 0.7399 & 0.1756 & \textbf{0.0898} \\
 &  & Rel. err. & 0.7998 & 0.8186 & 0.2632 & \textbf{0.1411} \\
 \specialrule{1pt}{2pt}{2pt}
\multirow{2}{*}{\textbf{Class 3}} & \multirow{2}{*}{\textbf{Conv.-Diff.-Reac.}} & Abs. err. & 0.7450 & 0.7415 & 0.8590 & \textbf{0.0353} \\
 &  & Rel. err. & 0.7960 & 0.7915 & 0.9532 & \textbf{0.0812} \\
\specialrule{1pt}{2pt}{2pt}
\end{tabular}
\label{tbl:add_main_table}
\end{table}

For more comprehensive evaluation, we conduct additional ablation studies following the experimental settings of Table~\ref{tbl:result} with the coefficient range of $1\sim20$ using PINN-P (cf. Appendix~\ref{sec:baseline_appendix}) and LargePINN, which is PINN with bigger network size. As shown in Table~\ref{tbl:param}, since the model size of our proposed P$^2$INN is larger than original PINN, we conduct experiments using a LargePINN model. The LargePINN has the same MLP architecture as the original PINN but with increased hidden dimensions, resulting in a model size of 82,941.

The experimental results of LargePINN, PINN-P, and P$^2$INN are summarized in Table~\ref{tbl:add_main_table}. In all scenarios, as indicated by Table~\ref{tbl:add_main_table}, the LargePINN model consistently performs inferiorly compared to \PPINN{}, and \PPINN{} outperforms PINN-P in all cases except one. That is, while the baselines struggles when learning the equations encompassing wide coefficient ranges, i.e., $1\sim20$. For instance, considering Conv.-Reac.-Diff. equation, the $L_2$ absolute error exhibited by P$^2$INN is 0.0353 whereas LargePINN and PINN-P have errors of 0.7415 and 0.8590, respectively. Note that this collective outcome underscores that P$^2$INN’s separation of PDE parameters and spatial/temporal coordinates during the learning process significantly enhances both generalization capabilities and scalability. 

\clearpage

\section{Ablation Studies on Varying Data Points}

We conduct a more challenging experiment by extremely reducing the training data points. Assuming the training of reaction equations for $\rho\in[1, 10]$, a total of 10,000 collocation points are utilized during the PINN’s training process - with 10 separate models processing 1,000 collocation points each. On the other hand, for \PPINN{}, 10 equations are jointly learned by a single model, also with a total of 10,000 collocation points. We name those two models PINN(10,000) and P$^2$INN(10,000) respectively, and compare these two models with new ablation model, P$^2$INN(1,000). To be specific, for P$^2$INN(1,000), a single model learned from a subset of 100 data points taken from each of the 10 equations, resulting in a total of 1,000 collocation points. The outcome of this experiment is summarized in the table presented below.

\begin{table}[ht!]
\centering
\renewcommand{\arraystretch}{0.4}
\footnotesize
\caption{Experimental results on Reaction equation with varying $\rho$}
\begin{tabular}{lcccccccccc}
\specialrule{1pt}{2pt}{2pt}
& $\pmb{\rho=1}$ & $\pmb{\rho=2}$ & $\pmb{\rho=3}$ & $\pmb{\rho=4}$ & $\pmb{\rho=5}$ & $\pmb{\rho=6}$ & $\pmb{\rho=7}$ & $\pmb{\rho=8}$ & $\pmb{\rho=9}$ & $\pmb{\rho=10}$ \\
\specialrule{1pt}{2pt}{2pt}
\textbf{PINN(10,000)} & 0.0045 & 0.0041 & 0.0047 & 0.7974 & 0.8598 & 0.8820 & 0.9021 & 0.9159 & 0.9263 & 0.9350 \\ \cmidrule{1-11}
\textbf{P$^2$INN(1,000)} & 0.0078 & 0.0049 & 0.0036 & 0.0036 & 0.0029 & 0.0020 & 0.0020 & 0.0014 & 0.0014 & 0.0016 \\ \cmidrule{1-11}
\textbf{P$^2$INN(10,000)} & 0.0034 & 0.0029 & 0.0017 & 0.0022 & 0.0014 & 0.0012 & 0.0008 & 0.0008 & 0.0008 & 0.0009 \\
\specialrule{1pt}{2pt}{2pt}
\end{tabular}
\label{tbl:add_data_point_num}
\end{table}

According to the Table~\ref{tbl:add_data_point_num}, \PPINN{} with only 1,000 data points, i.e., P$^2$INN(1,000), achieves successful outcomes for the failure mode of PINN involving reaction equations with values over 4, outperforming PINN by a big margin. Notably, its results are comparable to \PPINN{} with the original setting, i.e., P$^2$INN(10,000).

\end{document}